\documentclass[sigconf, nonacm]{acmart}

\newcommand{\eat}[1]{}

\usepackage{latexsym}
\usepackage{amsfonts}
\usepackage{amsmath}
\usepackage{xcolor}
\usepackage{colortbl}
\usepackage{epsfig}
\usepackage{xspace}
\usepackage{graphicx}
\usepackage{subfigure}
\usepackage{paralist}
\usepackage{enumerate}
\usepackage[color,matrix,arrow,all]{xy}
\usepackage{comment}
\usepackage{booktabs}
\usepackage{balance}
\usepackage{stmaryrd}
\usepackage{pifont}
\usepackage{hhline}
\usepackage{listings}
\usepackage{array}
\usepackage{float}
\usepackage[flushleft]{threeparttable}

\usepackage{mathrsfs}
\usepackage{makecell}
\usepackage{xparse}
\usepackage{wrapfig}

\usepackage{epsfig}
\usepackage{multirow}
\usepackage{url}

\usepackage{multirow}
\usepackage{natbib}
\usepackage{graphicx}

\usepackage{listings}
\usepackage{framed}
\usepackage{xcolor}
\usepackage{color}
\usepackage{geometry}
\usepackage{changepage}
\colorlet{shadecolor}{gray!20}

\definecolor{shadecolor}{RGB}{220,220,220}

\definecolor{inputcolor}{RGB}{255,139,35}
\definecolor{outputcolor}{RGB}{120,212,252}
\definecolor{embedcolor}{RGB}{254,127,156}
\definecolor{maskcolor}{RGB}{122,128,255}
\definecolor{ecolor}{RGB}{58,149,54}

\definecolor{highcolor}{RGB}{255,153,153}
\definecolor{midcolor}{RGB}{255,204,204}
\definecolor{lowcolor}{RGB}{204,229,255}

\newtheorem{example}{Example}

\usepackage{tikz}
\usetikzlibrary{shapes,snakes}
\usetikzlibrary{calc}

\usepackage{algorithm} 
\usepackage{algorithmic}  
\usepackage[algo2e]{algorithm2e} 

\usepackage[export]{adjustbox}

\definecolor{green}{RGB}{0,128,0}

\definecolor{yellow}{RGB}{255,200,18}

\newcommand{\add}[1]{\textcolor{black}{{#1}}}

\newcommand{\stab}{\vspace{1.2ex}\noindent}
\newcommand{\sstab}{\rule{0pt}{8pt}\\[-2.2ex]}

\newcommand{\bi}{\begin{itemize}}
\newcommand{\ei}{\end{itemize}}

\newcommand{\be}{\begin{enumerate}}
\newcommand{\ee}{\end{enumerate}}
\newcommand{\beqn}{\begin{eqnarray*}}
\newcommand{\eeqn}{\end{eqnarray*}}

\newcommand{\stitle}[1]{\stab\noindent{\bf #1}}

\newcommand{\ie}{i.e.,\xspace}
\newcommand{\eg}{e.g.,\xspace}

\newcommand{\eop}{\hspace*{\fill}\mbox{$\Box$}}     

\newcommand{\sys}{TaxoGlimpse\xspace}

\makeatletter
    \newcommand\figcaption{\def\@captype{figure}\caption}
    \newcommand\tabcaption{\def\@captype{table}\caption}
\makeatother

\tikzstyle{mybox} = [draw=black, fill=black!5, thick,
   rectangle, rounded corners, inner sep=0pt, inner ysep=6pt]
\tikzstyle{fancytitle} =[fill=black, text=white]

\NewDocumentCommand{\nan}{ mO{} }{\textcolor{blue}{\textsuperscript{\textit{Nan}}\textsf{\textbf{\small[#1]}}}}

\NewDocumentCommand{\yushi}{ mO{} }{\textcolor{green}{\textsuperscript{\textit{Yushi}}\textsf{\textbf{\small[#1]}}}}

\usepackage{tabularx}
\usepackage{multirow}
\usepackage{algorithm}
\usepackage{algorithmic}
\usepackage{subfigure}
\usepackage{adjustbox}
\usepackage{tcolorbox} 
\tcbuselibrary{breakable}
\usepackage{booktabs}
\usepackage{tabularx}

\usepackage[normalem]{ulem}

\newcommand\vldbavailabilityurl{https://github.com/ysunbp/TaxoGlimpse}

\newcommand{\rev}[1]{\textcolor{black}{#1}}

\begin{document}

\pagestyle{plain}


\title{Are Large Language Models a Good Replacement of Taxonomies? [Experiment, Analysis \& Benchmark]}

\author{Yushi Sun}
\orcid{0000-0003-3853-6364}
\affiliation{%
  \institution{HKUST}
}
\email{ysunbp@cse.ust.hk}

\author{Hao Xin}
\orcid{0000-0002-0523-6675}
\affiliation{%
  \institution{HKUST}
}
\email{hxinaa@cse.ust.hk}

\author{Kai Sun}
\orcid{0000-0001-8262-4906}
\affiliation{%
  \institution{Meta Reality Labs}
}
\email{sunkaicn@meta.com}

\author{Yifan Ethan Xu}
\affiliation{%
  \institution{Meta Reality Labs}
}
\email{ethanxu@meta.com}

\author{Xiao Yang}
\affiliation{%
  \institution{Meta Reality Labs}
}
\email{xiaoyangfb@meta.com}

\author{Xin Luna Dong}
\affiliation{%
  \institution{Meta Reality Labs}
}
\email{lunadong@meta.com}

\author{Nan Tang}
\orcid{0000-0003-2832-0295}
\affiliation{%
  \institution{HKUST (GZ) / HKUST}
}
\email{nantang@hkust-gz.edu.cn}

\author{Lei Chen}
\orcid{0000-0002-8257-5806}
\affiliation{%
  \institution{HKUST(GZ) / HKUST}
}
\email{leichen@cse.ust.hk}

\begin{abstract}
Large language models (LLMs) demonstrate an impressive ability to internalize knowledge and answer natural language questions. Although previous studies validate that LLMs perform well on general knowledge while presenting poor performance on long-tail nuanced knowledge, the community is still doubtful about whether the traditional knowledge graphs should be replaced by LLMs. In this paper, we ask {\em if the schema of knowledge graph (\ie taxonomy) is made obsolete by LLMs}. Intuitively, LLMs should perform well on common taxonomies and at taxonomy levels that are common to people. Unfortunately, there lacks a comprehensive benchmark that evaluates the LLMs over a wide range of taxonomies from common to specialized domains and at levels from root to leaf so that we can draw a confident conclusion. To narrow the research gap, we constructed a novel taxonomy hierarchical structure discovery benchmark named \sys\footnote{All datasets collection was done by HKUST. The source code, data, and/or other artifacts have been made available at \url{\vldbavailabilityurl}.} to evaluate the performance of LLMs over taxonomies. \sys covers \rev{ten representative taxonomies} from common to specialized domains with in-depth experiments of different levels of entities in this taxonomy from root to leaf. Our comprehensive experiments of \add{eighteen state-of-the-art LLMs} under three prompting settings validate that LLMs can still not well capture the knowledge of specialized taxonomies and leaf-level entities. 

\end{abstract}

\maketitle





\section{Introduction}
\label{sec:introduction}

Recently, we have witnessed the rapid advancements of large language models (LLMs) such as GPTs~\cite{achiam2023gpt} and Llamas~\cite{touvron2023llama}. These LLMs have demonstrated impressive abilities in a wide range of applications such as question answering~\cite{tan2023evaluation}, information retrieval~\cite{zhu2023large}, news summarization~\cite{zhang2023benchmarking}, entity relation extraction~\cite{wadhwa2023revisiting}, and data preparation~\cite{sun2023reca}, among many others, disrupting and redefining the development of these areas. However, several previous studies also pointed out that LLMs are significantly less knowledgeable in domain-specific long-tail knowledge details~\cite{sun2023head, bang2023multitask}, sparking a growing debate about whether traditional knowledge graphs will be replaced by LLMs in real applications~\cite{sun2023head, choi2023common, tan2023evaluation}.

As the schema of knowledge graphs, taxonomies provide a structured way to organize and categorize knowledge, which is indeed a kind of ``knowledge about knowledge'' (or meta-knowledge), serving as an important asset in different applications such as knowledge/information management~\cite{sen2019knowledge}, data integration~\cite{papadimitriou2012taci}, knowledge extraction~\cite{karamanolakis2020txtract}, and domain-specific recommendation~\cite{huang2019taxonomy}. The general form of taxonomies involves a hierarchical structure that organizes entities and concepts into categories based on their characteristics or relationships. Typically, taxonomies follow a tree-like structure, where each category is represented as a node, and the relationships between categories are depicted as hypernymy (Is-A) links. The top-level category represents the broadest classification, while the lower-level categories become more specific. 

Traditionally, taxonomies are used to assist entity searching (e.g., "best health tracker" as a shopping query), category display, and knowledge reasoning, which rely on the basic ``Is-A'' relation constructed between parent and child entities.
%
Recently, LLMs have shown the ability to learn world knowledge, which opens up an opportunity to also store the ``Is-A'' taxonomy structure in LLMs' parameters. This raises an important question: Are traditional hierarchical structures in taxonomies made obsolete by LLMs~\cite{sun2023head, neuhaus2023ontologies, babaei2023llms4ol}? 

\begin{figure}[t!]
  \centering
  \includegraphics[width=1.0\linewidth]{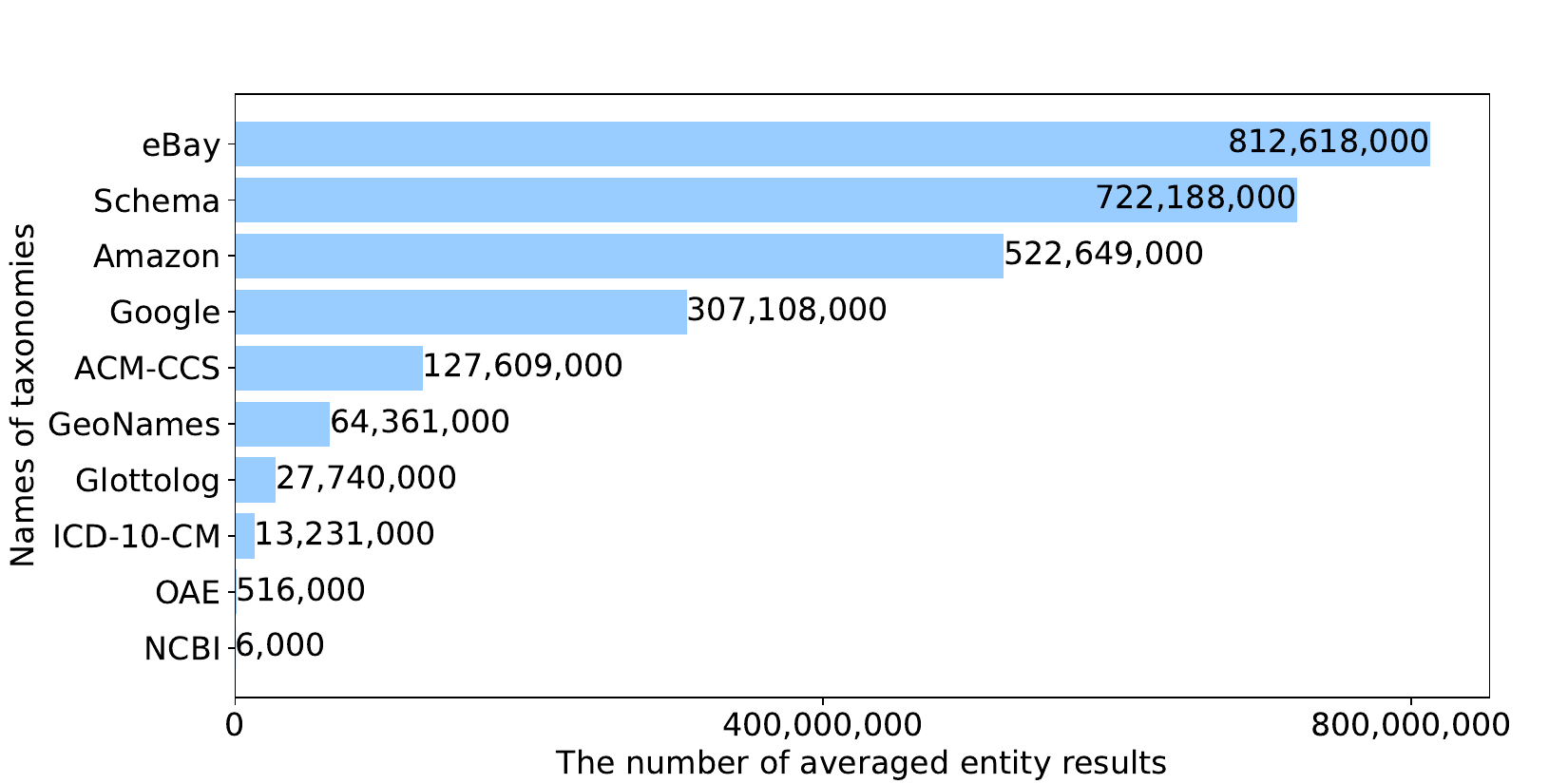}
  \vspace{-2.5em}
  \caption{\rev{The popularity of different taxonomies.}}
  \label{fig:popularity}
  \vspace{-1em}
\end{figure}

\vspace{-0.6em}
\begin{example} 
\label{exam:mot}
$[${\bf Taxonomies.}$]$
Figure~\ref{fig:example} shows taxonomy snippets from common (at the top) to specialized (at the bottom) domains ranked by the popularity of taxonomies as illustrated in Figure~\ref{fig:popularity}. From left to right, we present the entities from root to leaf levels. 

\sstab
$[${\bf From Common to Specialized Domains.}$]$
To get an understanding of LLMs' knowledgeability in determining the  ``Is-A'' relations in taxonomies, we prompt GPT-4~\cite{achiam2023gpt} model with: `` Is <child entity> a type of <parent entity>?'' and record the overall annotation accuracy of GPT-4~\cite{achiam2023gpt} on each taxonomy. Specifically, GPT-4 achieves $85.7\%$ accuracy on Google taxonomy, while achieving $70.8\%$ and $62.6\%$ accuracies on ACM-CCS and Glottolog taxonomies respectively, which is consistent with the intuition that GPT-4 performs better on common domains while exhibiting poorer performance on specialized domains.

\begin{figure*}[t!]
  \centering
  \includegraphics[width=1.0\linewidth]{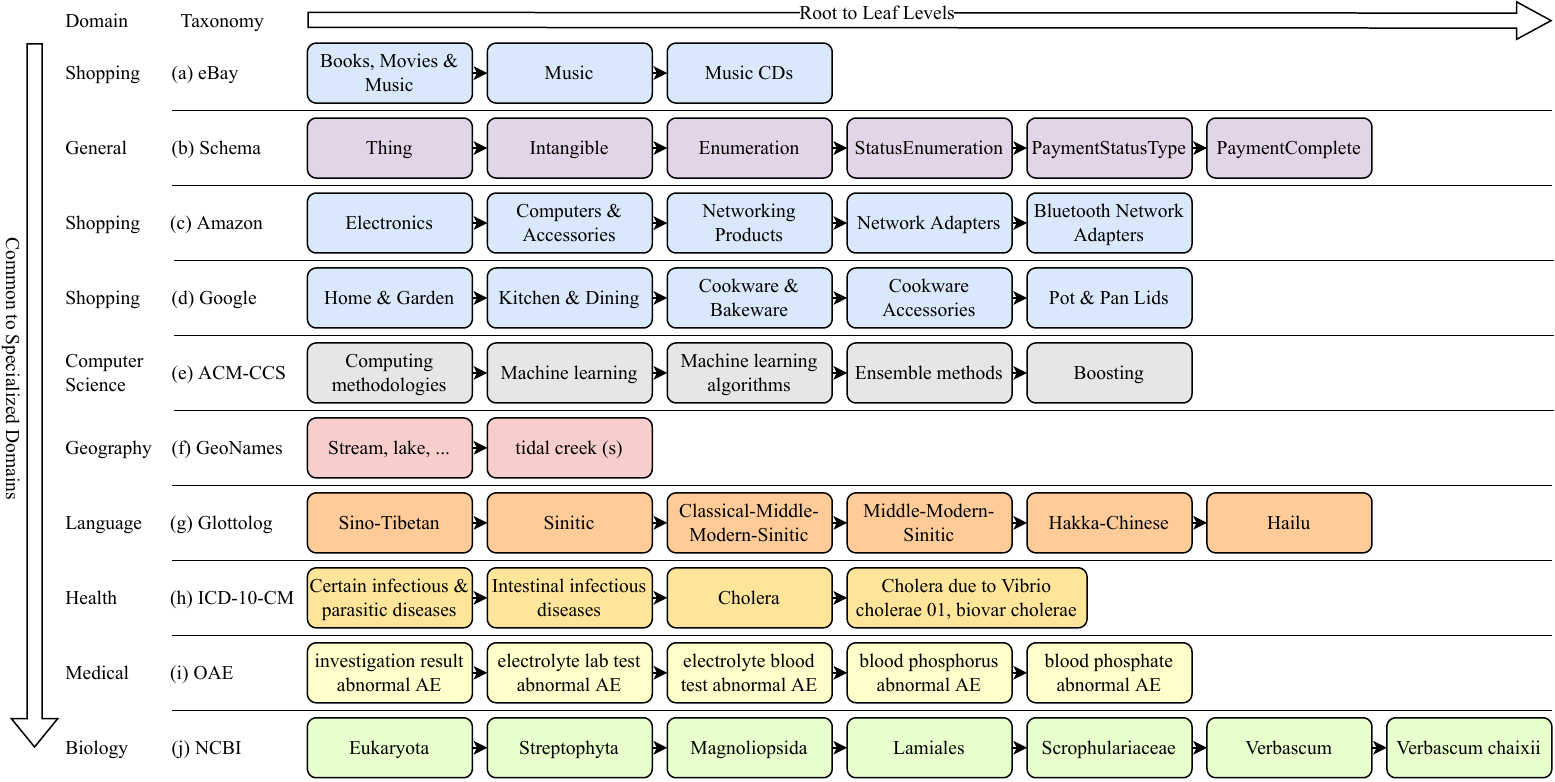}
  \vspace{-2em}
  \caption{Exemplar chain of entities snippets of \add{ten taxonomies}. From top to bottom, we list the taxonomy snippets from common domains to specialized domains. From left to right, we present entities from the root to leaf levels.}
  \label{fig:example}
  \vspace{-1em}
\end{figure*}

\sstab
$[${\bf From Root to Leaf Levels.}$]$
\rev{We further query each level of the exemplar chain of Glottolog taxonomy. 
The queries are again provided in a child-to-parent manner: e.g., Is Sinitic language a type of Sino-Tibetan language? We observe that GPT-4 gave incorrect answers at the Hailu to Hakka-Chinese and the Hakka-Chinese to Middle-Modern-Sinitic levels, while correctly answering the rest,} which means it tends to be more knowledgeable near the root levels while becoming less reliable near the leaf levels of Glottolog. \eop
\end{example}

Example~\ref{exam:mot} shows that LLMs' knowledgeability in taxonomies varies based on multiple factors such as the popularity of the taxonomies or the depth at which a question is posed. Motivated by the observations, we conduct a systematic study to address a crucial question: {\bf Can LLMs Effectively Replace Taxonomies?}

The importance of the study is three-fold:
(1) {\bf Industrial users} can understand if constructing and maintaining traditional taxonomies is worth investing in;
(2) {\bf LLM developers} can learn about the pros and cons of their models in taxonomies and improve accordingly to help users better perform taxonomy-related tasks with LLMs; and
(3) {\bf Database researchers} can innovate on the novel forms of taxonomy structures, and explore interesting and meaningful research problems/application domains that may boost the reasoning of LLMs.


\stitle{Challenges.}
%
Addressing the question faces three challenges:





\sstab
({\bf C1}) {\bf The Absence of a Comprehensive Benchmark.} To our knowledge, there is no comprehensive benchmark that can effectively answer the key research question. Such a benchmark should encompass a wide variety of taxonomies taking into account diverse characteristics such as popularity, domains, and complexity.

\sstab
({\bf C2}) {\bf Formulating an Evaluation Strategy. } Taxonomies possess a unique hierarchical structure, which presents a challenge when designing probing strategies to thoroughly evaluate the knowledgeability of LLMs from root to leaf levels.
    
\sstab
({\bf C3}) {\bf Diversity of LLMs.} The field of LLMs is rapidly evolving, with a multitude of base models available and an even greater number of configurations such as model sizes and prompting methods. It is critical to establish a systematic approach for a thorough study. 


%




\stitle{Contributions.}
We make the following notable contributions.

\stitle{(1) \sys: A New Benchmark.}
In response to challenges (C1) and (C2), we constructed a benchmark, namely \sys, which covers taxonomies from \rev{eight} different domains ranging from common to specialized domains. The taxonomies selected for each domain are representative, and have a range of different numbers of entities, levels, and trees.
Moreover, we designed the questions on each level of different taxonomies, enabling users to have an in-depth glimpse into the knowledge of LLMs from root to leaf of the taxonomies.


\stitle{(2) A Comprehensive Evaluation.}
To resolve challenge (C3), we evaluate \add{eighteen} state-of-the-art LLMs, including GPTs, \rev{Claude-3, Llama-2 (7B, 13B, 70B), Llama-3 (8B, 70B), Mistral (7B), Mixtral (8*7B),} and so on, to have decent coverage of state-of-the-art LLMs and their variants.
In addition, we consider common prompt engineering settings, such as zero-shot, few-shot, and Chain-of-Thoughts, to systematically evaluate the knowledge of LLMs.


\stitle{(3) Key Research Questions.}
By designing the new benchmark and conducting comprehensive experiments, we answer the following questions.


\begin{itemize}
    \item [({\bf i})] {\em How reliable are LLMs for determining hierarchical structures in different taxonomies?} 
    LLMs perform well on common taxonomies (\eg Shopping); however, their performance downgrades significantly on specialized taxonomies (\eg Biology).
    
    \item [({\bf ii})] {\em Do LLMs perform equally well among different levels of taxonomies?}
    LLMs roughly achieve progressively worse performance from root to leaf in most taxonomies.

    \item [({\bf iii})] {\em Do normal methods that improve LLMs increase the reliability?} 
    The increase in sizes and the adoption of domain-agnostic fine-tuning of LLMs may not lead to an increase in performance. \rev{The domain-specific instruction tuning leads to stable and significant performance improvements.}

    \item [({\bf iv})] {\em Do different prompting settings influence the performance?} 
    The performance changes of best LLMs brought by few-shot and Chain-of-Thoughts prompting settings are minimal.
\end{itemize}

\vspace{-2ex}
\stitle{(4) Future Opportunities.}
%
For practitioners, we recommend a) continuing manual taxonomy construction for specialized domains such as Language, and near the root levels of taxonomies for the purposes of displaying; b) leveraging LLMs for taxonomy-based search and reasoning in common domains such as shopping, to save manual work in constructing lower level of taxonomies.
For researchers, we suggest exploring the development of taxonomies in an LLM-tree-structure-combined form, where entities reside implicitly as LLM embeddings or exist explicitly in link forms.

Moreover, we publish our code and datasets on GitHub~\cite{taxo-exp} to attract more research in this direction. 

\section{Benchmark Construction and Question Design}
\label{sec:benchmark}

\subsection{Benchmark Construction}
\label{sec:data-collection}

We selected taxonomies from \rev{eight} domains to cover a wide range of taxonomic knowledge. When selecting the taxonomies, we considered the following criteria: 

\sstab 
(1) The taxonomies are publicly accessible; 

\sstab 
(2) The taxonomies cover different domains with different popularity to ensure a comprehensive view of LLMs' performance from common to specialized domains; 

\sstab 
(3) The characteristics (number of entities, number of levels, number of trees) of taxonomies are diverse to ensure representativity; and 

\sstab 
(4) The taxonomies are representative in each domain and widely used by their respective communities. 

With these criteria in mind, we selected taxonomies from \rev{eight} domains: \rev{Shopping (Google Product Category~\cite{Google-shopping-tax}, Amazon Product Category~\cite{Amazon-shopping-tax}, and eBay Categories~\cite{ebay}), General (Schema.org~\cite{guha2016schema}), Computer Science Research (ACM Computing Classification System~\cite{ccs-concept}), Geography (GeoNames~\cite{geonames}), Language (Glottolog~\cite{Glottolog-release}), Health (ICD-10-CM~\cite{ICD-tax}), Medical (OAE~\cite{he2014oae}), and Biology (The NCBI Taxonomy Database~\cite{ncbi-data}). }

Figure~\ref{fig:popularity} depicts the popularity of the selected taxonomies, \rev{measured by the average number of results returned by google.com by searching (exact match) the names of 100 randomly sampled concepts from each taxonomy.} \rev{Specifically, eBay, Schema.org, Amazon, and Google taxonomies are the representatives of the common taxonomies that cover common entities known to ordinary people. ACM-CCS, GeoNames, Glottolog, ICD-10-CM, OAE, and NCBI are the specialized taxonomies that are domain-specific and likely to be accessed by domain experts.} Apart from the popularity of taxonomies, we also considered the characteristics of different taxonomies. As shown in Table~\ref{tab:taxonomies}, \rev{the characteristics we selected give the audience the basic information about the scale of the taxonomies (number of entities), the depth of the taxonomies (number of levels), the width of the taxonomies (number of trees), the rough shape of the taxonomies (number of nodes and classes in each level). The selected taxonomies have a range of cover of different numbers of entities (from 500 to 2M), number of levels (from 2 to 7), and number of tree roots (from 3 to 245), representing well the diverse distribution of the morphology of taxonomies in different application domains and scenarios.} We now discuss the data collection details on each domain.

\begin{table*}[t!]
  \caption{\rev{Statistics of taxonomies.}}
  \vspace{-1em}
  \label{tab:taxonomies}
  {\color{black}
  {\small
  \begin{tabular}{l|l|l|l|l|l}
    \Xhline{0.8pt}
    \multicolumn{1}{c|}{\textbf{Domain}}&\multicolumn{1}{c|}{\textbf{Taxonomy}}&\multicolumn{1}{c|}{\textbf{\# of entities}}&\multicolumn{1}{c|}{\textbf{\# of levels}}&\multicolumn{1}{c|}{\textbf{\# of trees}}&\multicolumn{1}{c}{\textbf{\# of nodes and classes in each level}}\\
    \cline{0-5}
    \multirow{3}{*}{Shopping} & eBay & 595 & 3 & 13 & 13-110-472\\
    & Amazon & 43814 & 5 &41 & 41-507-3910-13579-25777\\
    & Google & 5595 & 5 & 21 & 21-192-1349-2203-1830 \\
    \cline{0-5}
    General & Schema & 1346 & 6 & 3 & 3-17-215-403-436-272 \\
        \cline{0-5}
        Computer Science & ACM-CCS & 2113 & 5 & 13 & 13-84-543-1087-386\\
        \cline{0-5}
    Geography & GeoNames & 689 & 2 & 9 & 9-680\\
    \cline{0-5}
    Language & Glottolog & 11969 & 6 & 245 & 245-712-1048-1205-1366-7393\\
    \cline{0-5}
    Health & ICD-10-CM & 4523 & 4 & 22 &22-155-963-3383\\
    \cline{0-5}
    Medical & OAE & 9547 & 5 & 181 & 181-1854-3817-2587-1108\\
    \cline{0-5}
    Biology & NCBI & 2190125 & 7 & 53 & 53-309-514-1859-10215-107615-2069560\\
        \Xhline{0.8pt}
  \end{tabular}}
  } 
  \vspace{-1em}
\end{table*}

\stitle{Shopping Taxonomies.}
%
\rev{We selected Google Product Category~\cite{Google-shopping-tax}, Amazon Product Category~\cite{Amazon-shopping-tax}, and eBay Categories~\cite{ebay}, which are the representative taxonomies in the shopping domains}: Google Product Category is used by Google Shopping, which is the most widely used for product price comparison in the United States according to~\cite{Google-shopping-stat}. Amazon Product Category is from Amazon.com, which is the most visited e-commerce shopping website in the United States~\cite{Amazon-shopping-stat}. \rev{The eBay Categories is from ebay.com, which is another popular online shopping platform.}

Despite that all these taxonomies target the shopping domain, they have significant differences in size and organization of categorization. As shown in Table~\ref{tab:taxonomies}, the Amazon Product Category is larger in the number of entities and the size of top-level classifications. As a result, the Amazon Product Category provides a finer-grained classification of products. 
By evaluating LLMs on \rev{the three shopping taxonomies}, we can gain a comprehensive view of LLMs' performance in the shopping domain. 

\rev{We collected the Google Product Category (US version) from the official link provided by Google~\cite{Google-shopping-tax} and crawled the Amazon Product Category and eBay Categories from~\cite{Amazon-shopping-tax} and~\cite{ebay}, respectively.} In order to gain a holistic view of LLMs performance in different levels of the taxonomies, we pre-processed and divided the entities into five levels for the Google and Amazon taxonomies: root level, level 1, level 2, level 3, and level 4 or lower for the Google and Amazon taxonomies \rev{and three levels for the eBay taxonomy. We present the detailed statistics of the three taxonomies in Table~\ref{tab:taxonomies} and exemplar snippets of the three taxonomies in Figure~\ref{fig:example}(a),(c),(d).} 

{\color{black}
\stitle{General Taxonomies.}
We adopted the Schema.org taxonomy~\cite{guha2016schema} as a representative for the general domain, which is a community effort to develop schemas for the structured data from the internet. Schema.org is a general domain taxonomy covering a wide range of concepts on the internet and serves as the basis for other general-purpose knowledge bases such as YAGO~\cite{suchanek2007yago}. 

As shown in Table~\ref{tab:taxonomies}, the Schema.org taxonomy contains six levels with a total number of 1346 entities, covering coarse concepts such as Thing to fine-grained concepts such as PaymentComplete. We used the newest release v26.0 of Schema.org from the official link~\cite{schemaorg}.
}

\stitle{Computer Science Research Taxonomies.}
For the computer science research domain, we selected the ACM Computing Classification System (ACM CCS)~\cite{ccs-concept}, which is the standard classification system for papers in the computer science field. The CCS concept taxonomy is widely used by researchers to accurately categorize their work so that other researchers can easily overview the main topics and quickly refer to related papers. 

As shown in Table~\ref{tab:taxonomies}, we considered five levels in the ACM CCS concept taxonomy. We provide an example for the entities in ACM CCS in Figure~\ref{fig:example}(e). We adopted the ACM CCS concept taxonomy version 2012 through ACM's website~\cite{ccs-concept-code}.

{\color{black}
\stitle{Geography Taxonomies.}
We selected the GeoNames taxonomy~\cite{geonames} for the geography domain. The GeoNames taxonomy is representative of this domain covering a two-level classification of the common geographical concepts. 

As shown in Table~\ref{tab:taxonomies}, the GeoNames taxonomy contains two levels with 689 concepts. We downloaded the GeoNames taxonomy from the official data release website~\cite{geonames}.
}

\stitle{Language Taxonomies.}
We chose Glottolog taxonomy~\cite{nordhoff2011glottolog, hammarstrom2023glottolog, swj-glottocodes} to represent the language domain, which is widely used by linguists. Glottolog offers an extensive inventory of languages, language families, and dialects found across the globe, that linguists need to be able to identify~\cite{caines2016glottolog}.  

As shown in Table~\ref{tab:taxonomies}, we considered six levels in the Glottolog taxonomy with a total number of 11,969 languoid entities. The six levels of Glottolog cover a taxonomic structure from language family (e.g., Sino-Tibetan in Figure~\ref{fig:example}(g)) to a specific language (e.g., Hakka-Chinese) or dialect (e.g., Hailu). We adopted the release v4.8 of Glottolog from~\cite{Glottolog-release}.

\stitle{Health Taxonomies.}
%
We selected the ICD-10-CM taxonomy~\cite{ICD-tax} for evaluation. ICD-10-CM is a representative candidate for the health domain designed by the Centers for Disease Control and Prevention of the US.

As shown in Table~\ref{tab:taxonomies}, ICD-10-CM taxonomy contains four levels. The root to level 3 concepts correspond to the rough classification of diseases based on body system or condition, the detailed classification, common disease group, and disease entities with different causes. Level 3 entities can be considered as the instances in the ICD-10-CM taxonomy. We present a snippet for ICD-10-CM in Figure~\ref{fig:example}(h) for better understanding. The ICD-10-CM taxonomy is accessed through the simple\_icd\_10\_CM 2.0.1 package~\cite{ICD-python}.

{\color{black}
\stitle{Medical Taxonomies.}
We selected the OAE taxonomy (Ontology of Adverse Events)~\cite{he2014oae}, which is a taxonomy specialized for adverse events. The OAE taxonomy has been developed to standardize the annotation of adverse events, integrate various adverse event data, and support computer-assisted reasoning.

As shown in Table~\ref{tab:taxonomies}, we considered five levels with a total number of 9547 entities in the OAE taxonomy from the coarse- to fine-grained classifications of the adverse events. We adopted the newest release v1.2.47 of OAE from~\cite{OAE}.} 

\stitle{Biology Taxonomies.}
We selected the NCBI Taxonomy Database~\cite{schoch2020ncbi, sayers2019genbank} as a representative in the biology domain. The NCBI taxonomy serves as the primary repository for standard nomenclature and classification within the International Nucleotide Sequence Database Collaboration (INSDC) and encompasses several prominent databases, including GenBank, ENA (EMBL), and DDBJ~\cite{federhen2012ncbi}. 

Following the instructions provided by~\cite{schoch2020ncbi}, we considered seven levels in the taxonomy, aligning with the biological taxonomy order: 1) superkingdom/kingdom/high-level clade, 2) phylum, 3) class, 4) order, 5) family, 6) genus, and 7) species. An example for the seven levels is presented in Figure~\ref{fig:example}(j). We downloaded the version (Sep 2023) of the NCBI taxonomy at the time when our experiments started from the official website~\cite{ncbi-data}.



\subsection{Question Design}
\label{sec:question-design}


We first discuss the question templates we designed for each taxonomy, followed by the question generation process for each respective question type.

\stitle{Question Templates.}
In order to understand LLMs' ability to discover hierarchical relationships in taxonomies and to take into account the characteristics of different taxonomies, we designed the simple-formed True/False templates \rev{and Multiple-Choice Question (MCQ) templates for each domain in Tables~\ref{tab:question} and~\ref{tab:question-mcq}.}

We observed similar results when using slight paraphrasing of the templates \rev{(the slight paraphrasing for the True/False questions replaces the words ``a type of'' with ``a kind of'' and ``a sort of''; while for the MCQ questions, we replace the word ``appropriate'' with ``suitable'' and ``proper'')}, so will report results on these templates only \rev{and present the full experimental results on all the template variants in our GitHub repository~\cite{taxo-exp}}.

\add{Tables~\ref{tab:question} and~\ref{tab:question-mcq} present the detailed templates we used for evaluating the LLMs on True/False and MCQ question types respectively.}

\begin{table}[t!]
  \vspace{-2em}
  \caption{\rev{Question templates (True/False).}}
  \label{tab:question}
  \vspace{-1em}
  {\small
  {\color{black}\begin{tabular}{m{0.075\textwidth}<{\raggedright}|m{0.355\textwidth}<{\raggedright}}
    \Xhline{0.8pt}
    \textbf{Domains} & \textbf{Question Templates} \\
    \hline
    Shopping & Are <child-type> products a type of <parent-type> products? answer with (Yes/No/I don't know)\\
    \hline
    General & Is <child-type> entity type a type of <parent-type> entity type? answer with (Yes/No/I don't know)\\
    \hline
    Computer Science & Is <child-type> computer science research concept a type of <parent-type> computer science research concept? answer with (Yes/No/I don't know)\\
    \hline
    Geography & Is <child-type> geographical concept a type of <parent-type> geographical concept? answer with (Yes/No/I don't know)\\
    \hline
    Language & Is <child-type> language a type of <parent-type> language? answer with (Yes/No/I don't know)\\
    \hline
    Health / Biology & Is <child-type> a type of <parent-type>? answer with (Yes/No/I don't know)\\
    \hline
    Medical & Is <child-type> Adverse Events concept a type of <parent-type> Adverse Events concept? answer with (Yes/No/I don't know)\\
    \Xhline{0.8pt}
  \end{tabular}
  } 
  } 
  \vspace{-2em}
\end{table}

\stitle{Question Generation.}
%
The questions were generated concerning the levels of child entities. For each taxonomy, we randomly sample entities from each level of the taxonomy except the root level. The sample sizes were determined based on the number of entities in each level, with a confidence level of 95\% and a margin of error of 5\% as suggested by~\cite{Qualtrics}. As shown in Table~\ref{tab:question}, besides <child-type>, we also need to obtain <parent-type> to form valid True/False questions. \rev{As for the MCQs in Table~\ref{tab:question-mcq}, we need to obtain four options to form each valid question.} For ease of demonstration of relationship inside a taxonomy, we considered the following notations: $e_n$ denotes an entity in level $n$, ($n=0,1,2,...$); $E_n$ denotes the set of all entities in level $n$; $e_n.p$ denotes the entity $e_n$'s intermediate parent entity \rev{(direct parent entity)}; $e_n.s$ denotes the set of sibling entities of $e_n$. To comprehensively understand LLMs' performance on taxonomic data, we consider the following generation modes:

\begin{itemize}
    \item {\bf positive:} Directly get the intermediate parent entity $e_n.p$ of the sampled child entity $e_n$.
    \item {\bf negative-easy:} Randomly sample a negative parent entity from the set $E_{n-1}-\{e_n.p\}$.
    \item {\bf negative-hard:} Randomly sample a negative parent entity from the set $(e_n.p).s$ (uncles of the child entity).
    \item \rev{{\bf MCQ:} Randomly sample three negative options from the set $(e_n.p).s$, and preserve the parent entity $e_n.p$ as the ground truth option.}
\end{itemize}

The reason why we generated negative-hard and negative-easy questions is to provide hard negatives and easy negatives. Intuitively, the negative-hard questions tend to be more difficult since the negative samples are siblings of the ground truth parent entity (i.e., uncles), which means these entities are more similar to the ground truth in comparison with randomly sampled negative samples, serving as the hard negatives for the LLMs. \rev{The evaluation was conducted in three sets of data for each level of taxonomies: positive + negative-easy, positive + negative-hard, and MCQ, which were denoted as easy, hard, and MCQ datasets respectively. Detailed statistics of the easy, hard, and MCQ datasets at each level of different taxonomies are presented in Table~\ref{tab:data-stat}.}


\section{Experimental settings}
\label{sec:settings}

In this section, we introduce the LLMs considered in our experiments, the implementation details we adopted, and the metrics.

\subsection{Large Language Models}
\label{sec:llms}

\begin{table}[t!]
  \vspace{-2em}
  \caption{\rev{Question templates (MCQ).}}
  \label{tab:question-mcq}
  \vspace{-1em}
  {\small
  {\color{black}\begin{tabular}{m{0.075\textwidth}<{\raggedright}|m{0.355\textwidth}<{\raggedright}}
    \Xhline{0.8pt}
    \textbf{Domains} & \textbf{Question Templates} \\
    \hline
    Shopping & What is the most appropriate supertype of <child-type> product? A) B) C) D)\\
    \hline
    General & What is the most appropriate supertype of <child-type> entity type? A) B) C) D)\\
    \hline
    Computer Science & What is the most appropriate supertype of <child-type> research concept? A) B) C) D)\\
    \hline
    Geography & What is the most appropriate supertype of <child-type> geographical concept? A) B) C) D)\\
    \hline
    Language & What is the most appropriate supertype of <child-type> language? A) B) C) D)\\
    \hline
    Health / Biology & What is the most appropriate supertype of <child-type>? A) B) C) D)\\
    \hline
    Medical & What is the most appropriate supertype of <child-type> Adverse Events concept? A) B) C) D)\\
    \Xhline{0.8pt}
  \end{tabular}
  } 
  } 
  \vspace{-1em}
\end{table}

\begin{table*}[t!]
  \vspace{-2em}
  \caption{\rev{Statistics of datasets.}}
  \vspace{-1em}
  \label{tab:data-stat}
  {\color{black}\begin{tabular}{@{}ll|l|l|l|l|l|l|l|l|l|l}
    \Xhline{0.8pt}
     & & \textbf{eBay} & \textbf{Amazon} & \textbf{Google} & \textbf{Schema} & \textbf{ACM-CCS} & \textbf{GeoNames} & \textbf{Glottolog} & \textbf{ICD-10-CM} & \textbf{OAE} & \textbf{NCBI} \\
    \hline
    \multirow{3}{*}{Level 1-root} & Easy &   176 & 438 & 258 & 34 & 138 & 492 & 500 & 222 & 638 & 344 \\
    \cline{2-12}
    ~ & Hard &   176 & 438 & 258 & 34 & 138 & 492 & 500 & 222 & 638 & 344 \\
    \cline{2-12}
    ~ & MCQ &  88 & 219 & 129 & 17 & 69 & 246 & 250 & 111 & 319 & 172 \\
    \hline
    \multirow{3}{*}{Level 2-1}                          & Easy &   430 & 700 & 600 & 276 & 450 & n/a & 564 & 550 & 700 & 440 \\ \cline{2-12}
    ~ & Hard &   430 & 700 & 597 & 276 & 450 & n/a & 564 & 550 & 700 & 439 \\ 
    \cline{2-12}
    ~ & MCQ & 215 & 350 & 300 & 138 & 225 & n/a & 282 & 275 & 350 & 220 \\
    \hline
    \multirow{3}{*}{Level 3-2}                          & Easy &   n/a & 748 & 656 & 394 & 568 & n/a & 584 & 690 & 670 & 638 \\
 \cline{2-12}
    ~ & Hard &  n/a & 748 & 653 & 394 & 567 & n/a & 584 & 690 & 670 & 636 \\
    \cline{2-12}
    ~ & MCQ & n/a & 374 & 328 & 197 & 284 & n/a & 192 & 345 & 335 & 319 \\
    \hline
    \multirow{3}{*}{Level 4-3}                          & Easy &   n/a & 758 & 636 & 410 & 386 & n/a & 600 & n/a & 572 & 742 \\
\cline{2-12}
    ~ & Hard &   n/a & 758 & 626 & 410 & 370 & n/a & 600 & n/a & 572 & 741 \\
    \cline{2-12}
    ~ & MCQ & n/a & 379 & 318 & 205 & 193 & n/a & 300 & n/a & 286 & 371 \\
    \hline
    \multirow{3}{*}{Level 5-4}                          & Easy &  n/a & n/a & n/a & 320 & n/a & n/a & 732 & n/a & n/a & 766 \\
\cline{2-12}
    ~ & Hard &  n/a & n/a & n/a & 320 & n/a & n/a & 732 & n/a & n/a & 766 \\
    \cline{2-12}
    ~ & MCQ &  n/a & n/a & n/a & 160 & n/a & n/a & 366 & n/a & n/a & 383 \\
    \hline
    \multirow{3}{*}{Level 6-5}                          & Easy & n/a & n/a & n/a & n/a & n/a & n/a & n/a & n/a & n/a & 770 \\
\cline{2-12}
    ~ & Hard & n/a & n/a & n/a & n/a & n/a & n/a & n/a & n/a & n/a & 770 \\
    \cline{2-12}
    ~ & MCQ & n/a & n/a & n/a & n/a & n/a & n/a & n/a & n/a & n/a & 385 \\
    \hline
    \multirow{3}{*}{{\bf Total}}                          & Easy & 606 & 2644 & 2150 & 1434 & 1542 & 492 & 2980 & 1462 & 2580 & 3700 \\
 \cline{2-12}
    ~ & Hard & 606 & 2644 & 2134 & 1434 & 1525 & 492 & 2980 & 1462 & 2580 & 3696 \\
    \cline{2-12}
    ~ & MCQ & 303 & 1322 & 1075 & 717 & 771 & 246 & 1490 & 731 & 1290 & 1850 \\
    \Xhline{0.8pt}
  \end{tabular}}
  \vspace{-1em}
\end{table*}

We now introduce the LLM series considered in our experiments. In order to comprehensively evaluate the performance of state-of-the-art LLMs, we selected \rev{nine} popular LLM series with \rev{eighteen} models to conduct the experiments.
\begin{itemize}
    \item {\bf GPTs}~\cite{achiam2023gpt}: The Generative Pre-trained Transformers series, are advanced language models by OpenAI. We selected GPT-3.5 and GPT-4 as two representatives for evaluation. The models are close-sourced and accessed through API only.
    \rev{\item {\bf Claude-3}~\cite{Claude-v3-documentation}: Claude-3 is the newest release of the Claude family models by Anthropic, which is close-sourced and claims to set new benchmarks for multiple cognitive tasks. We experimented with the best variant Claude-3-Opus.}
    
    \item {\bf Llama-2s}~\cite{touvron2023llama}: The Llama-2 series is a set of open-sourced large language models released by Meta. We adopted Llama-2 7B, 13B, and 70B models with chat settings, which are suitable for the question-answering application scenario.
    \rev{\item {\bf Llama-3s}~\cite{llama3modelcard}: The Llama-3 series is a novel set of open-sourced large language models released by Meta in April 2024. We adopted Llama-3 8B and 70B models with instruct settings.}

    \item {\bf Flan-T5s}~\cite{chung2022scaling}: The Flan-t5s is an encoder-decoder LLM series developed by Google. We selected the best models from the series: Flan-t5-3B and Flan-t5-11B.
    
    \item {\bf Falcons}~\cite{almazrouei2023falcon}: Developed by TIIUAE, the Falcon series is claimed to achieve comparable performance with Llama-2s in question answering tasks~\cite{almazrouei2023falcon}. We chose Falcon-Instruct models with 7B and 40B parameters for our experiments, which are optimized for chat format.
    
    \item {\bf Vicunas}~\cite{chiang2023vicuna}: The Vicuna series~\cite{chiang2023vicuna} are the large language models developed based on the weights through domain-agnostic instruction fine-tuning. We include these models to investigate if domain-agnostic fine-tuning improves performance. We adopted Vicunas 7B, 13B, and 33B.
\rev{\item {\bf Mistrals}~\cite{jiang2023mistral,jiang2024mixtral}: Designed by Mistral AI, the Mistral and Mixtral models claimed to outperform the Llama-2 13B on several reasoning benchmarks. We adopted the Mistral-7B-Instruct and Mixtral-8*7B-Instruct models.}
\rev{\item {\bf LLMs4OL}~\cite{babaei2023llms4ol}: LLMs4OL is the state-of-the-art approach that utilizes instruction tuning~\cite{chung2024scaling} based on the Flan-T5-3B model to perform ontology learning tasks. Different from other LLMs, the LLMs4OL is the only domain-specific finetuned approach selected for comparison.}
\end{itemize}

\subsection{Implementation Details}
\label{sec:implementation-details}
We interacted with GPTs through Azure OpenAI API and the OpenAI official API. The employed GPT-3.5 version is 2023-05-15 and the GPT-4 version is 2023-11-06-preview. For other LLMs, we used 8 GeForce RTX 3090 GPUs and 4 NVIDIA A100 GPUs for the deployment. For the Vicuna series, we adopted the latest models Vicuna-7B-v1.5, Vicuna-13B-v1.5, and Vicuna-33B-v1.3. \rev{The original implementation of the LLMs4OL model is limited in general, geography, and medical domains. We adapted and fine-tuned the model to other taxonomy domains by preserving the official implementation details~\cite{llms4ol-code} suggested by LLMsOL as much as possible.} To obtain stable outputs from LLMs, we employed the most deterministic hyper-parameter settings (e.g., temperature=0). All LLMs were experimented with the same question set for all the experiments.


\subsection{Metrics}
\label{sec:metrics-selection}

    To cater to the needs for evaluating the quality of the answers provided by LLMs, similar to previous works~\cite{sun2023head, huang2023c}, we selected the following metrics for evaluation: {\bf accuracy} ($A$), and {\bf miss rate} ($M$), which measures the number of questions that LLMs give correct answers and `` I don't know'' over the number of all questions respectively. We consider an LLM as a good model if it achieves high accuracy with a low miss rate. 

\section{Experimental results}
\label{sec:experiment-results}


In this section, we analyze the experimental results of LLMs following multiple different research questions focusing on the overall performance, performance with respect to levels of taxonomies, the relationship between performance and model sizes, domain-agnostic fine-tuning\rev{, and domain-specific fine-tuning} and influence from the prompting settings. Additionally, we include an additional instance typing experiment to further evaluate LLMs on a taxonomy-related task.

\vspace{-1em}
\subsection{How reliable are LLMs for discovering hierarchical structures in different taxonomies?}
\label{sec:exp-common-to-specialized}


\begin{table*}[t!]
  \vspace{-2em}
  \caption{\rev{Overall results on hard datasets.}}
  \vspace{-1em}
  \label{tab:exp-main-hard}
  {\color{black}\begin{tabular}{l|l|c|c|c|c|c|c|c|c|c|c}
    \Xhline{0.8pt}
    \multicolumn{1}{c|}{}&\multicolumn{1}{c|}{}&\multicolumn{1}{c|}{\textbf{eBay}}&\multicolumn{1}{c|}{\textbf{Amazon}}&\multicolumn{1}{c|}{\textbf{Google}}&\multicolumn{1}{c|}{\textbf{Schema}}&\multicolumn{1}{c|}{\textbf{ACM-CCS}}&\multicolumn{1}{c|}{\textbf{GeoNames}}&\multicolumn{1}{c|}{\textbf{Glottolog}}&\multicolumn{1}{c|}{\textbf{ICD-10-CM}}&\multicolumn{1}{c|}{\textbf{OAE}}&\multicolumn{1}{c}{\textbf{NCBI}}\\
    \hline
    \multirow{2}{*}{GPT-3.5} & $A$ &0.891&0.724&0.814&0.591&0.617&0.598&0.510&0.838&0.767&0.495  \\
    & $M$ &  0.021&0.138&0.042&0.324&0.150&0.057&0.298&0.063&0.144&0.301\\
    \multirow{2}{*}{GPT-4} & $A$ & 0.921&0.806&0.857&0.734&0.708&0.652&0.626&0.917&0.822&0.653  \\
    & $M$ &  0.003&0.051&0.011&0.193&0.017&0.002&0.154&0.001&0.035&0.132
\\
\hline
    \multirow{2}{*}{Claude-3} & $A$ &  0.901&0.668&0.781&0.315&0.624&0.679&0.244&0.871&0.766&0.449
 \\
    & $M$ &  0.033&0.231&0.090&0.663&0.111&0.138&0.714&0.041&0.161&0.456
\\
    \hline
    \multirow{2}{*}{Llama-2-7B} & $A$ & 0.201&0.052&0.092&0.000&0.032&0.006&0.001&0.114&0.004&0.000
 \\
    & $M$ & 0.789&0.946&0.903&1.000&0.963&0.994&0.999&0.871&0.996&1.000
 \\
    \multirow{2}{*}{Llama-2-13B} & $A$ & 0.898&0.766&0.822&0.712&0.658&0.543&0.192&0.811&0.757&0.457
\\
    & $M$ & 0.000&0.002&0.003&0.010&0.011&0.006&0.681&0.027&0.078&0.252
  \\
    \multirow{2}{*}{Llama-2-70B} & $A$ & 0.899&0.806&0.836&0.616&0.687&0.553&0.305&0.826&0.747&0.535
\\
    & $M$ & 0.000&0.000&0.000&0.000&0.003&0.000&0.467&0.023&0.017&0.130
  \\
    \hline
    \multirow{2}{*}{Llama-3-8B} & $A$ & 0.880&0.789&0.774&0.788&0.664&0.663&0.608&0.878&0.851&0.691
 \\
    & $M$ & 0.000&0.000&0.000&0.000&0.000&0.000&0.078&0.000&0.000&0.011
 \\
    \multirow{2}{*}{Llama-3-70B} & $A$ & 0.904&0.770&0.824&0.419&0.705&0.693&0.388&0.881&0.800&0.551
\\
    & $M$ & 0.000&0.050&0.011&0.531&0.048&0.073&0.474&0.003&0.088&0.231

 \\
    \hline
    \multirow{2}{*}{Flan-T5-3B} & $A$ &0.899&0.781&0.835&0.743&0.672&0.539&0.584&0.767&0.838&0.593
\\
    & $M$ & 0.000&0.000&0.000&0.000&0.000&0.000&0.000&0.000&0.000&0.000
 \\
    \multirow{2}{*}{Flan-T5-11B} & $A$ & 0.919&0.793&0.864&0.786&0.698&0.520&0.589&0.842&0.856&0.633
\\
    & $M$ & 0.000&0.000&0.000&0.000&0.000&0.000&0.000&0.000&0.000&0.000
 \\
    \hline
    \multirow{2}{*}{Falcon-7B} & $A$ & 0.597&0.547&0.556&0.501&0.550&0.537&0.503&0.636&0.497&0.587
 \\
    & $M$ & 0.000&0.000&0.000&0.000&0.000&0.000&0.000&0.000&0.000&0.000
 \\
    \multirow{2}{*}{Falcon-40B} & $A$ & 0.434&0.253&0.348&0.013&0.043&0.108&0.021&0.454&0.007&0.013
 \\
    & $M$ & 0.515&0.711&0.591&0.987&0.950&0.858&0.975&0.489&0.991&0.986
  \\
        \hline
    \multirow{2}{*}{Vicuna-7B} & $A$ & 0.827&0.725&0.728&0.699&0.599&0.705&0.637&0.757&0.813&0.609
  \\
    & $M$ & 0.000&0.000&0.000&0.000&0.000&0.000&0.000&0.000&0.000&0.002
  \\
    \multirow{2}{*}{Vicuna-13B} & $A$ & 0.690&0.625&0.601&0.581&0.527&0.492&0.301&0.666&0.424&0.350
\\
    & $M$ & 0.007&0.037&0.022&0.093&0.023&0.114&0.557&0.077&0.450&0.460
 \\
    \multirow{2}{*}{Vicuna-33B} & $A$ & 0.759&0.713&0.682&0.728&0.591&0.728&0.496&0.772&0.820&0.522
 \\
    & $M$ & 0.000&0.000&0.000&0.020&0.000&0.000&0.277&0.001&0.002&0.187
 \\
    \hline
    \multirow{2}{*}{Mistral} & $A$ & 0.583&0.474&0.532&0.214&0.433&0.240&0.146&0.478&0.405&0.176
 \\
    & $M$ & 0.262&0.361&0.230&0.750&0.308&0.691&0.818&0.410&0.528&0.772
 \\
    \multirow{2}{*}{Mixtral} & $A$ & 0.805&0.739&0.738&0.707&0.618&0.604&0.394&0.840&0.789&0.482
 \\
    & $M$ & 0.000&0.000&0.000&0.017&0.111&0.041&0.450&0.018&0.052&0.313
 \\
    \hline
    \multirow{2}{*}{LLMs4OL} & $A$ & 0.904&0.849&0.860&0.912&0.753&0.677&0.711&0.891&0.906&0.725
 \\
    & $M$ & 0.000&0.000&0.000&0.000&0.000&0.000&0.000&0.000&0.000&0.000
\\

    \Xhline{0.8pt}
  \end{tabular}}
  \vspace{-1em}
\end{table*}    

We present the performance of LLMs on Hard, Easy, and MCQ datasets in Tables~\ref{tab:exp-main-hard},~\ref{tab:exp-main-easy}, and~\ref{tab:exp-mcq}.

\stitle{Accuracy.}
We observe an overall decreasing trend in accuracy for LLMs from the common to specialized taxonomies on the three datasets, demonstrating a drop in LLM's reliability when we go from common to specialized taxonomies. \rev{Exceptions are the OAE and ICD-10-CM taxonomies, where the LLMs achieve good performance. LLMs perform well on the OAE taxonomy might be due to the high similarity in terms of names between the parent and child concepts as shown in Figure~\ref{fig:example}. While the ICD-10-CM data might be covered by the training data of the LLMs since it is also widely used in many common and non-medical domains (e.g., insurance billing processes)~\cite{ICD-public}. On the NCBI, Glottolog and GeoNames hard datasets, the accuracy of the best LLM is only around $70\%$.} We attribute these phenomena to the fact that the domain knowledge of common taxonomies tends to be covered by the pre-training data of LLMs, while the knowledge of specialized taxonomies such as NCBI, Glottolog, and GeoNames is scarce on the internet and thus is less likely to be included in the pre-training data. Accurately determining hierarchical structures on specialized domains still requires support from the traditional taxonomy learning approaches.

\stitle{Miss Rate.}
When analyzing the miss rates of different LLMs, we observe that Flan-T5-3B, Flan-T5-11B, and LLMs4OL have zero miss rates, in other words, they always provide their best guesses, while Llama-2-7B and Falcon-40B tend to be conservative: always provide ``I don't know'' as responses. We further observe rises in miss rates of GPT-3.5, GPT-4, Vicuna-13B, and Vicuna-33B on the Glottolog and NCBI taxonomies, which are the difficult specialized taxonomies that most LLMs perform poorly on. This is desirable since these models have learned to be cautious in taxonomies where they do not have sufficient domain knowledge.

\stitle{Different question types.}
\rev{Comparing the experimental results between the Easy, Hard, and MCQ datasets, we observe that providing MCQ options significantly reduces the miss rates of the LLMs. For instance, the average miss rates of the Llama-3-70B model reduce from $0.151$ on the Hard datasets to $0.005$ on the MCQ datasets. The average accuracy of Llama-3-70B in turn rises from $0.694$ to $0.791$.
}

\stitle{\colorbox{black!10}{Finding 1:}}
The state-of-the-art LLMs are reliable in more common domains such as Shopping and General; while lacking sufficient domain knowledge in more specialized domains such as Computer Science Research, Biology, Language, and Geography.


\subsection{Do LLMs perform equally well among different levels of taxonomies?}
\label{sec:exp-root-to-leaf}

\begin{figure*}[t!]
\vspace{-2em}
\centering
\subfigure[shopping-eBay]{
\label{subfig:ebay-hard-zero-acc}
  \includegraphics[width=0.24\linewidth]{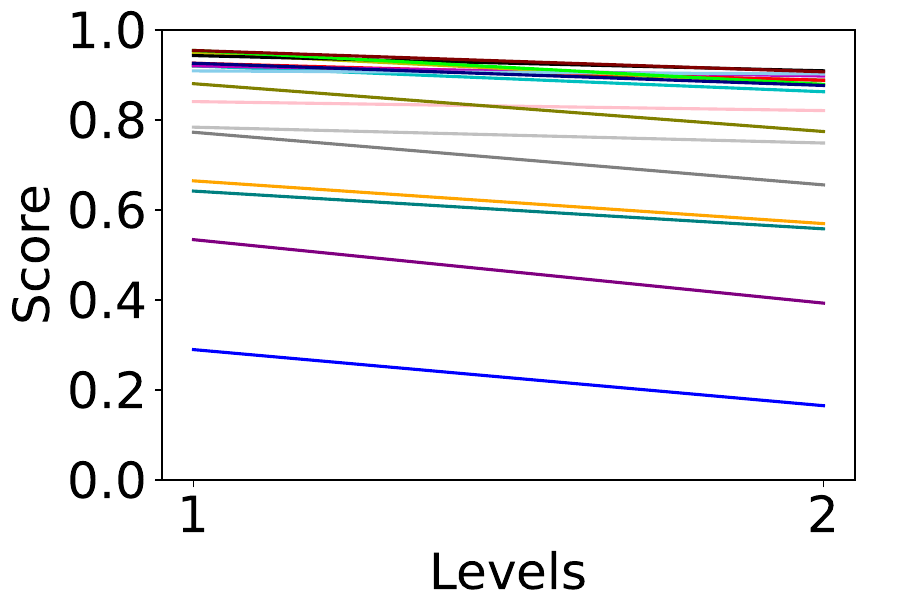}}
  \subfigure[shopping-Amazon]{
\label{subfig:amazon-hard-zero-acc}
  \includegraphics[width=0.24\linewidth]{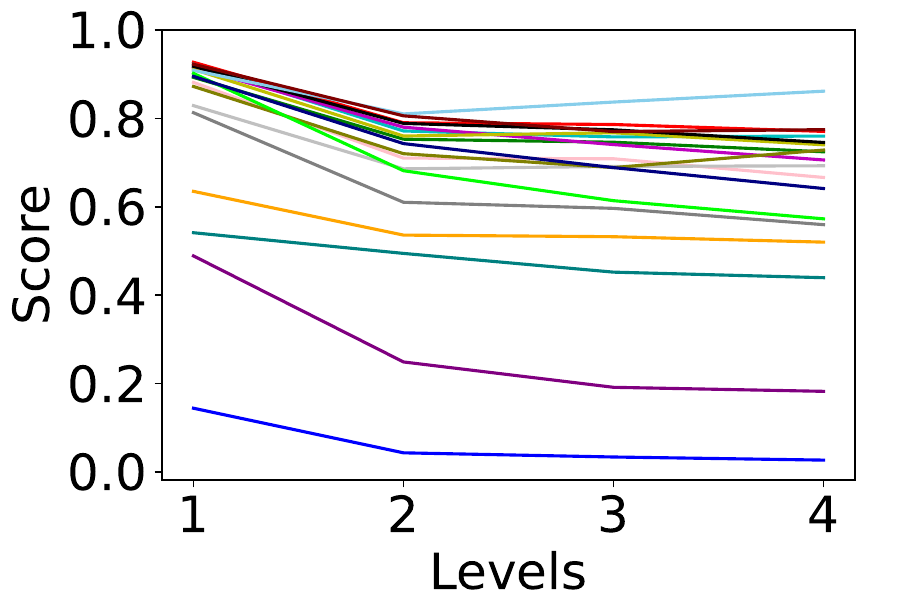}}
\subfigure[shopping-Google]{
\label{subfig:google-hard-zero-acc}
  \includegraphics[width=0.24\linewidth]{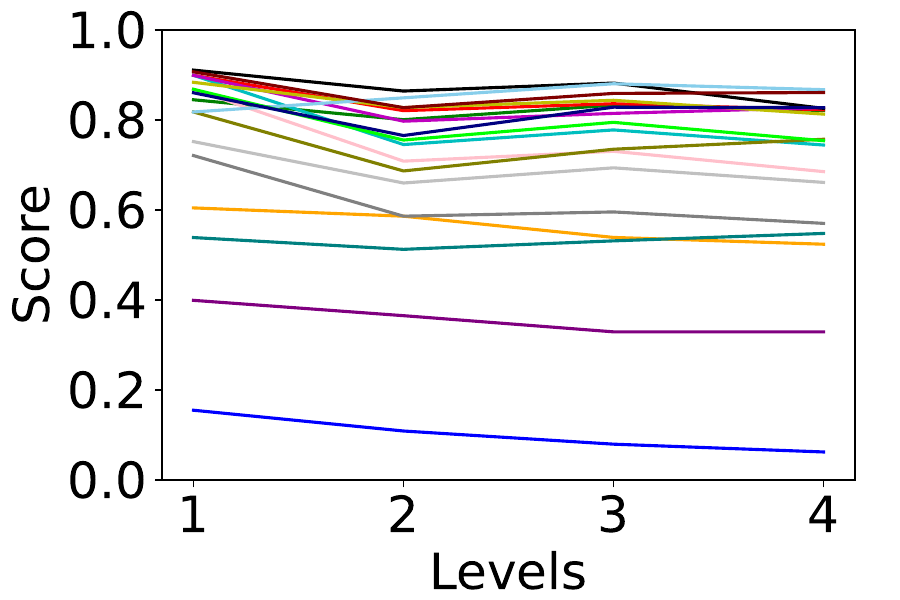}}
  \\
  \subfigure[general-Schema]{
\label{subfig:general-hard-zero-acc}
  \includegraphics[width=0.24\linewidth]{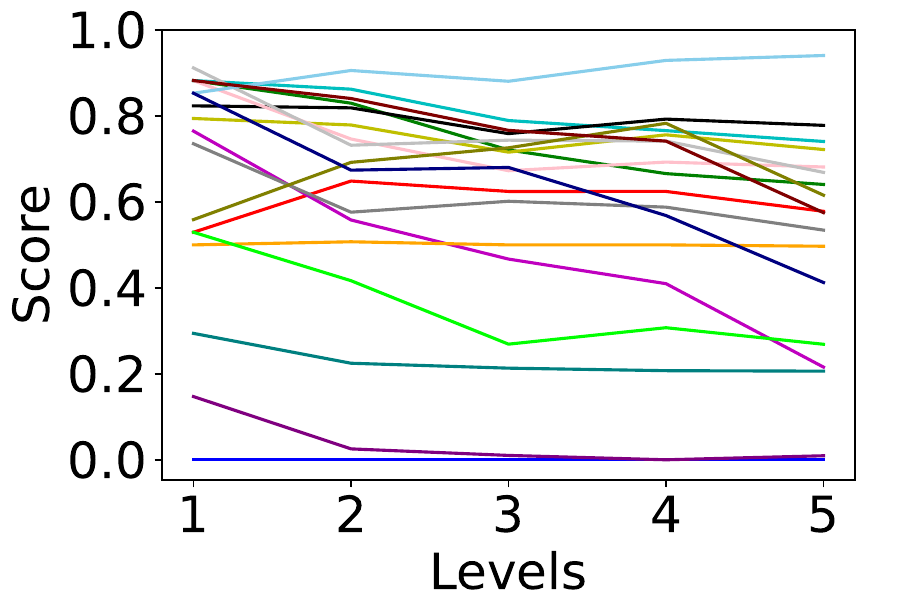}}
  \subfigure[CS-ACM]{
\label{subfig:academic-hard-zero-acc}
  \includegraphics[width=0.24\linewidth]{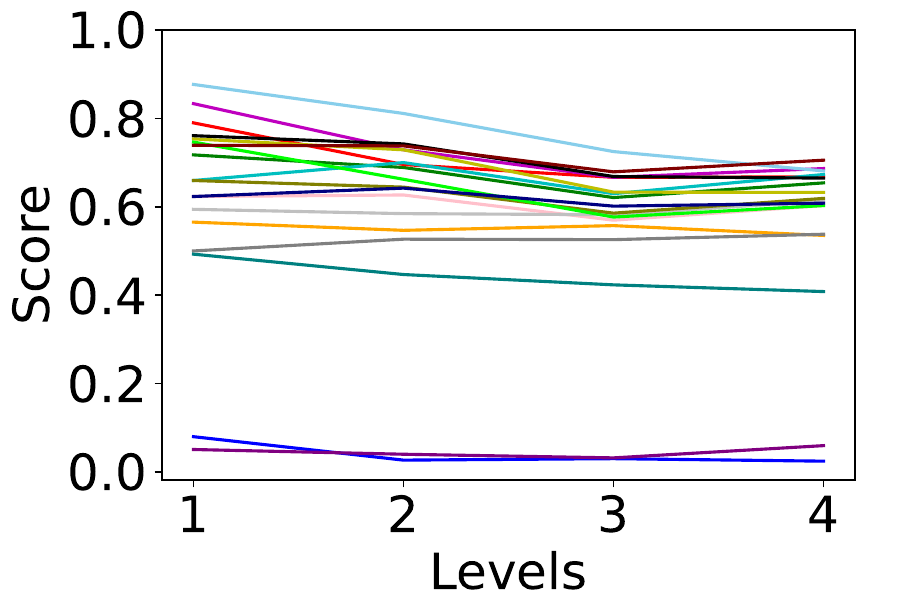}}
  \subfigure[language-Glottolog]{
\label{subfig:language-hard-zero-acc}
  \includegraphics[width=0.24\linewidth]{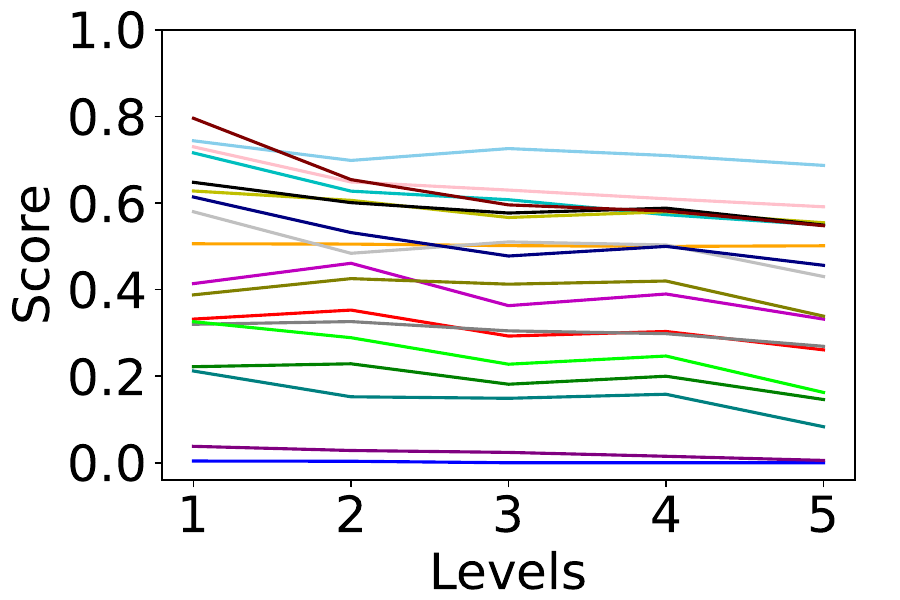}}
  \\
   \subfigure[health-ICD-10-CM]{
\label{subfig:medical-hard-zero-acc}
  \includegraphics[width=0.24\linewidth]{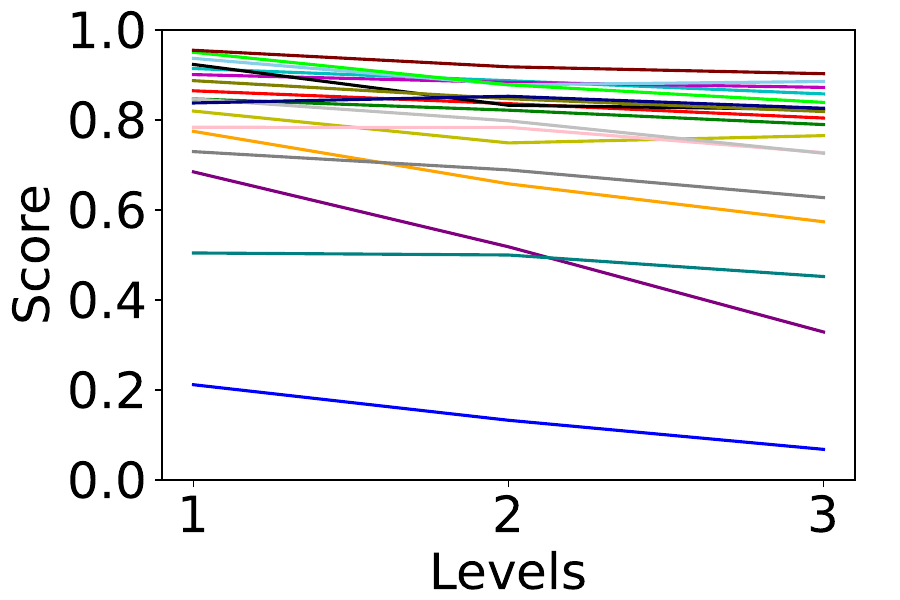}}
  \subfigure[medical-OAE]{
\label{subfig:OAE-hard-zero-acc}
  \includegraphics[width=0.24\linewidth]{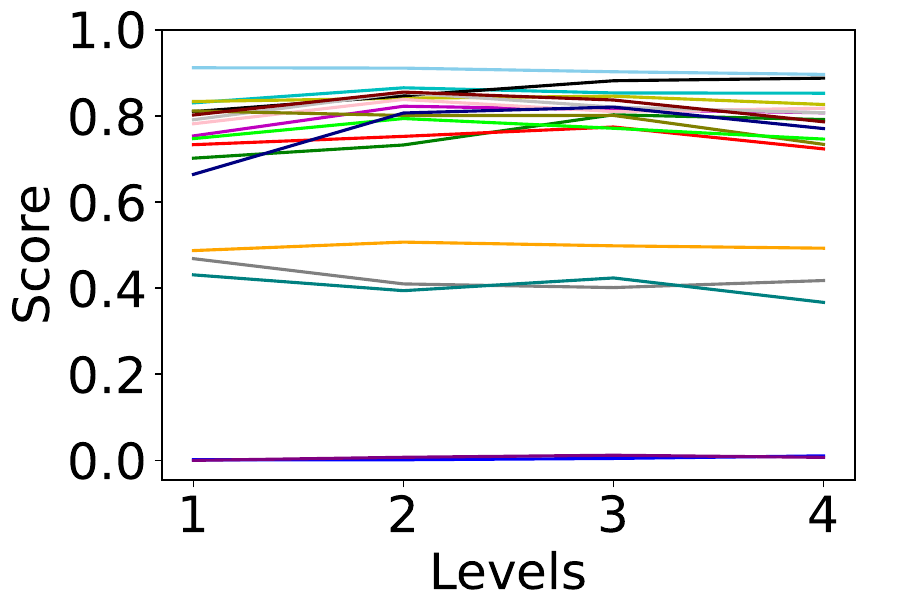}}
  \subfigure[biology-NCBI]{
\label{subfig:biology-hard-zero-acc}
  \includegraphics[width=0.24\linewidth]{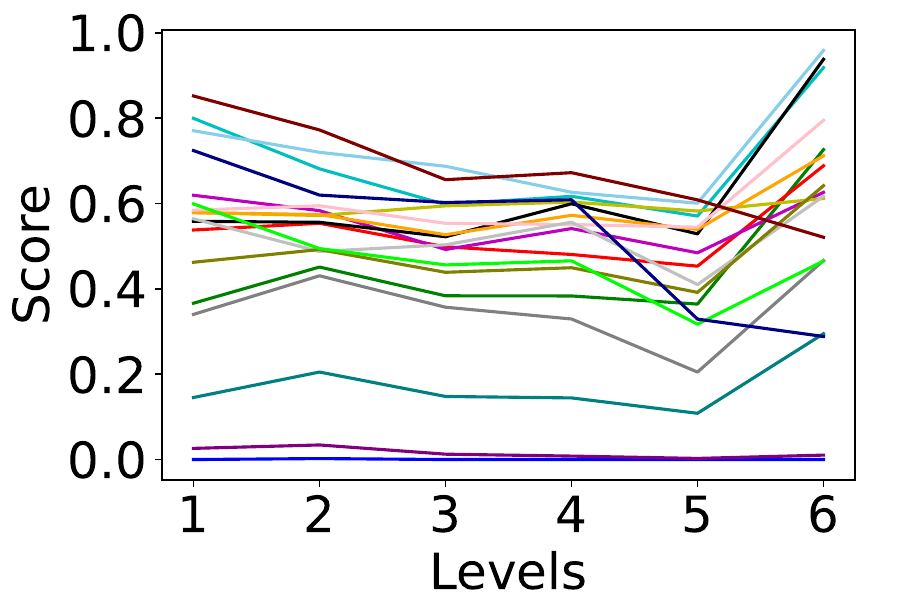}}
  \subfigure{ 
  {\includegraphics[width=0.12\linewidth]{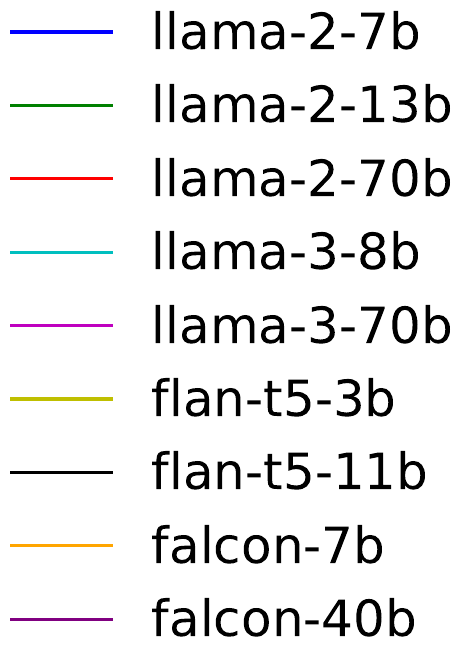}}}
  \subfigure{ 
  {\includegraphics[width=0.12\linewidth]{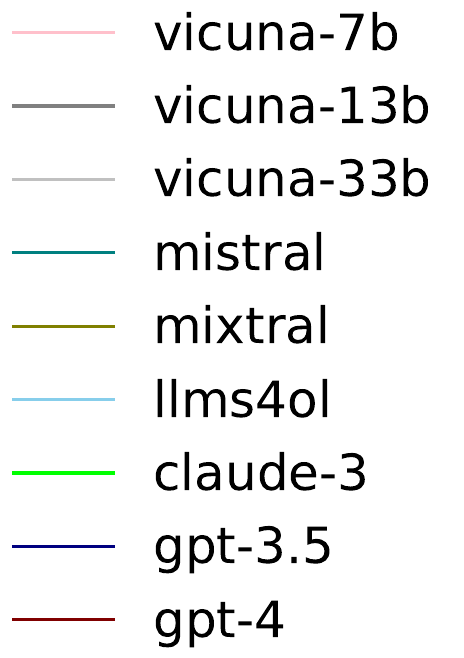}}}
  \vspace{-1.5em}
  \caption{\rev{Accuracies for different levels of questions in hard datasets of different taxonomies under the zero-shot setting.}
}
  \label{fig:hard-zero-acc}
  \vspace{-1.5em}
\end{figure*}


To answer this question, we conducted experiments on each level (level n to level n-1) of the taxonomies, because of the page limit, we only presented the accuracy results of hard datasets in Figure~\ref{fig:hard-zero-acc}. \rev{Since the GeoNames taxonomy only has two concept levels, resulting in only one set of experiments (level 1 to root), we omit the demonstration of its results in the figure.}

\stitle{Accuracy.}
\rev{For common shopping taxonomies, as shown in Figures~\ref{subfig:ebay-hard-zero-acc},~\ref{subfig:amazon-hard-zero-acc}, and~\ref{subfig:google-hard-zero-acc},} despite fluctuation, the accuracy of all LLMs tend to decrease as we go from the shallow levels (root) to deep levels (leaf). Most LLMs can achieve around $80\%$ accuracy in all levels in these taxonomies. A similar root-to-leaf performance decline trend can also be observed on taxonomies \rev{Schema.org, ACM-CCS, Glottolog, and ICD-10-CM as shown in Figures~\ref{subfig:general-hard-zero-acc},~\ref{subfig:academic-hard-zero-acc},~\ref{subfig:language-hard-zero-acc}, and~\ref{subfig:medical-hard-zero-acc}.} \rev{On the general domain taxonomy Schema.org, LLMs4OL, the best LLM achieves over $90\%$ accuracy across different levels, indicating its mastery of general domain knowledge.} We surprisingly observe that the root-to-leaf performance decline trend does not apply to the NCBI taxonomy as shown in Figure~\ref{subfig:biology-hard-zero-acc}: Most LLMs experience sudden performance uplifts at the last level. We notice that the last level in the NCBI taxonomy corresponds to the set of species-to-genus questions. We dived into the NCBI taxonomy database and discovered that this might be due to the data property of biological taxonomy: The names between species and corresponding genus tend to be similar in forms. For example, the ancestor chain of the species \textit{Verbascum chaixii} is \textit{Eukaryota-Strepyophyta-Magnoliopsida-Scrophulariaceae-Verbascum-Verbascum chaixii}, where the names between species and genus are very similar in forms. As a result, LLMs may take advantage of the similarity in forms between species and genus and thus achieve good performance at the last level. Despite the performance uplifts at the last level, the performances of the state-of-the-art LLMs at the middle levels (e.g., level 3, level 4, and level 5) are still very poor: slightly better than random guessing. \rev{A performance uplift trend from root to leaf levels can be observed in the OAE taxonomy, which also has similar name forms between the parent and child concepts near the leaf levels. }

\stitle{\colorbox{black!10}{Finding 2:}}
LLMs tend to present a root-to-leaf performance decline trend in most of the taxonomies. Additional support to improve LLMs' performance on leaf-level entities remains a promising direction for future ontology learning research.



\vspace{-0.5em}
\subsection{Do normal methods that improve LLMs increase the accuracy?}

\label{sec:exp-normal-methods}

\rev{We consider three normal methods that improve LLM reliability discussed by other works~\cite{zheng2023judging, touvron2023llama, babaei2023llms4ol} to see if they work on taxonomies: improving the model size, providing domain-agnostic instruction fine-tuning, conducting domain-specific instruction fine-tuning.}


\stitle{Larger Model Sizes.}\rev{Tables~\ref{tab:exp-main-hard},~\ref{tab:exp-main-easy}, and~\ref{tab:exp-mcq} show LLMs with different sizes and their corresponding performances in different taxonomies.} Since the GPTs and Claude-3 do not release information on model sizes, we only analyze the open-sourced models in this section. Specifically, for the Llama-2 series and Flan-T5 series, we notice that Llama-2-70B outperforms Llama-2-13B and Llama-2-7B and Flan-T5-11B outperforms Flan-T5-3B in most taxonomies, which indicates that increasing the sizes of LLMs can improve the models' performance for Llama-2 and Flan-T5. \rev{However, for the Vicunas and Flacons on the easy and hard datasets: Vicuna-7B outperforms Vicuna-13B in all the taxonomies and achieves better performance than Vicuna-33B in half of the taxonomies; Falcon-7B significantly outperforms Falcon-40B in all taxonomies. Besides, we observe that on the easy and hard datasets, the miss rates of Falcon-40B are significantly higher than those of Falcon-7B, which means Falcon-40B tends to be more conservative in answering hierarchical structure discovery questions and thus generates more ``I don't know'' answers.} This observation coincides with the observation regarding Falcon-40B presented in a previous study~\cite{sun2023head}. 
We attribute this phenomenon to the fact that once the LLM is sufficiently large, the differences in pre-training data and strategies play a vital role in determining the performance of answering hierarchical structure discovery questions in taxonomies.

\stitle{Domain-Agnostic Fine-Tuning.}
\rev{We further compare the Llama-2 series and the Vicuna series as shown in Tables~\ref{tab:exp-main-hard},~\ref{tab:exp-main-easy}, and~\ref{tab:exp-mcq} to consider the effect of domain-agnostic fine-tuning.} As discussed in~\cite{zheng2023judging,chiang2023vicuna}, the Vicuna models are fine-tuned Llama-2 models based on the dialog data in the ShareGPT dataset and thus should produce answers with higher quality. 
Naturally, an interesting question is whether such domain-agnostic fine-tuning can improve the performance of LLMs in answering taxonomy structure questions. 
However, we observe that although Vicuna-7B significantly improves the performance of Llama-2-7B, Vicuna-13B is outperformed by its original model Llama-2-13B on the easy and hard datasets. \rev{On the MCQ dataset, Vicuna-13B improves the performance of Llama-2-13B on some taxonomies only.} The reason might be that the domain-agnostic dialog data fine-tuning may have positive effects on the model to better understand the question format (Vicuna-7B and Llama-2-7B), while it might not improve the performance significantly and stably or even bring negative effects if the model can already well understand the question format because of the miss-match of knowledge coverage between the domain-agnostic fine-tuning data and the domain-specific taxonomy data (Vicuna-13B and Llama-2-13B). 


\stitle{Domain-specific Fine-tuning.}
\rev{Considering the domain-specific fine-tuning, we observe that the instruction-tuned LLMs4OL largely outperforms its backbone model Flan-T5-3B. Specifically, the averaged accuracy over all taxonomies of LLMs4OL boosts the averaged accuracy of Flan-T5-3B by $12.9\%$, $12.9\%$, and $17.0\%$ on the easy, hard, and MCQ datasets, which showcases the significant benefit of performing domain-specific instruction fine-tuning.
}

\stitle{\colorbox{black!10}{Finding 3:}}
Normal methods, including using larger model sizes and domain-agnostic fine-tuning, may not lead to an increase in performance. \rev{The domain-specific fine-tuning leads to a stable and significant performance uplift.} Indeed, the answer quality for the hierarchical structure discovery questions in taxonomies is related to the domain knowledge coverage of the pretraining data. \rev{Introducing domain-specific fine-tuning can increase the domain knowledge coverage of the LLMs. }

\vspace{-1em}

\subsection{Do different prompting settings influence the performance?}
\label{sec:exp-prompting}

\begin{figure*}[t!]
  \vspace{-2em}
\centering
  \subfigure[GPT-4, accuracy]{
\label{subfig:radar-gpt4-acc}
  \includegraphics[width=0.242\linewidth]{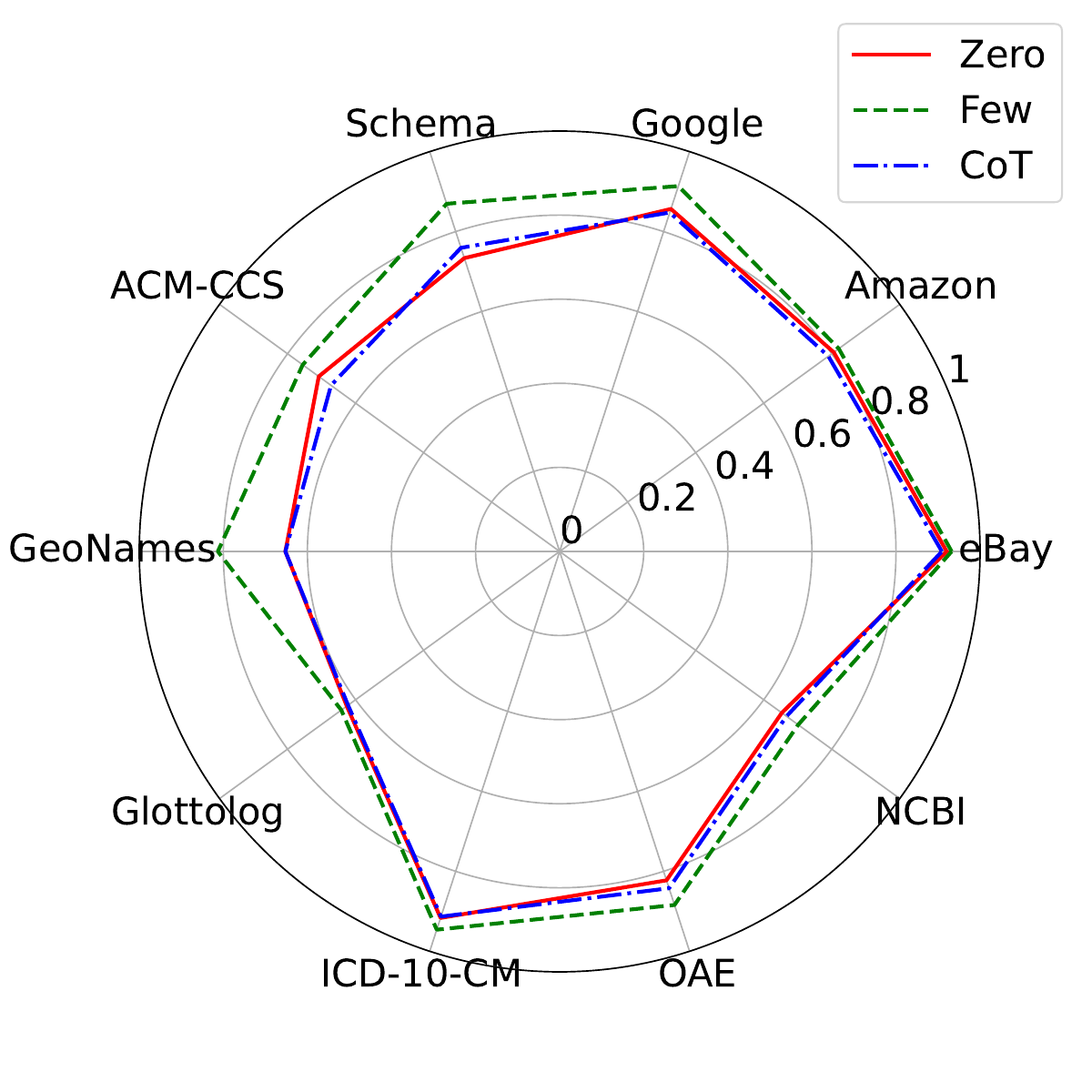}}
  \subfigure[Flan-T5-11B, accuracy]{
\label{subfig:radar-flan-t5-acc}
  \includegraphics[width=0.242\linewidth]{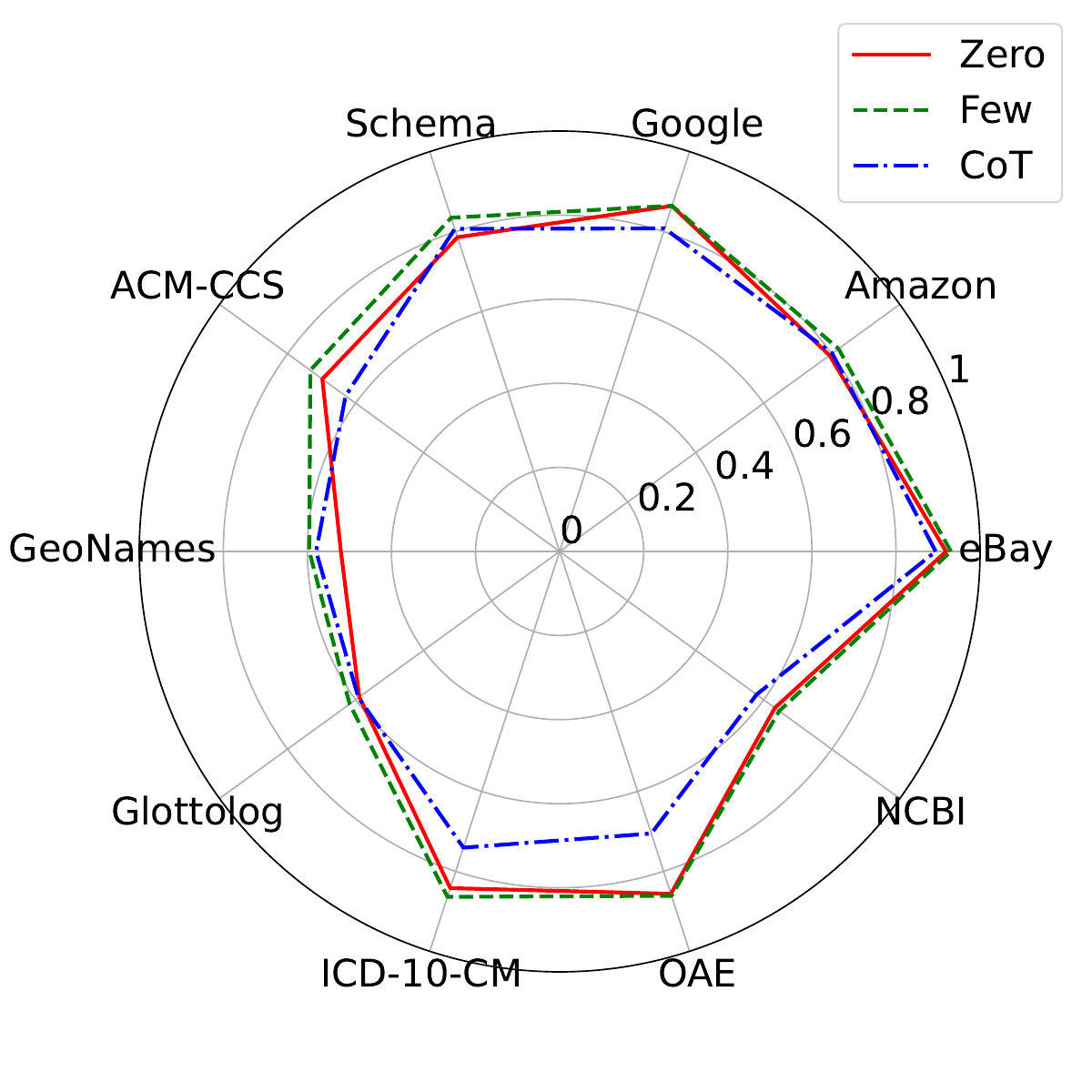}}
  \subfigure[Llama-2-7B, accuracy]{
\label{subfig:radar-llama-acc}
  \includegraphics[width=0.242\linewidth]{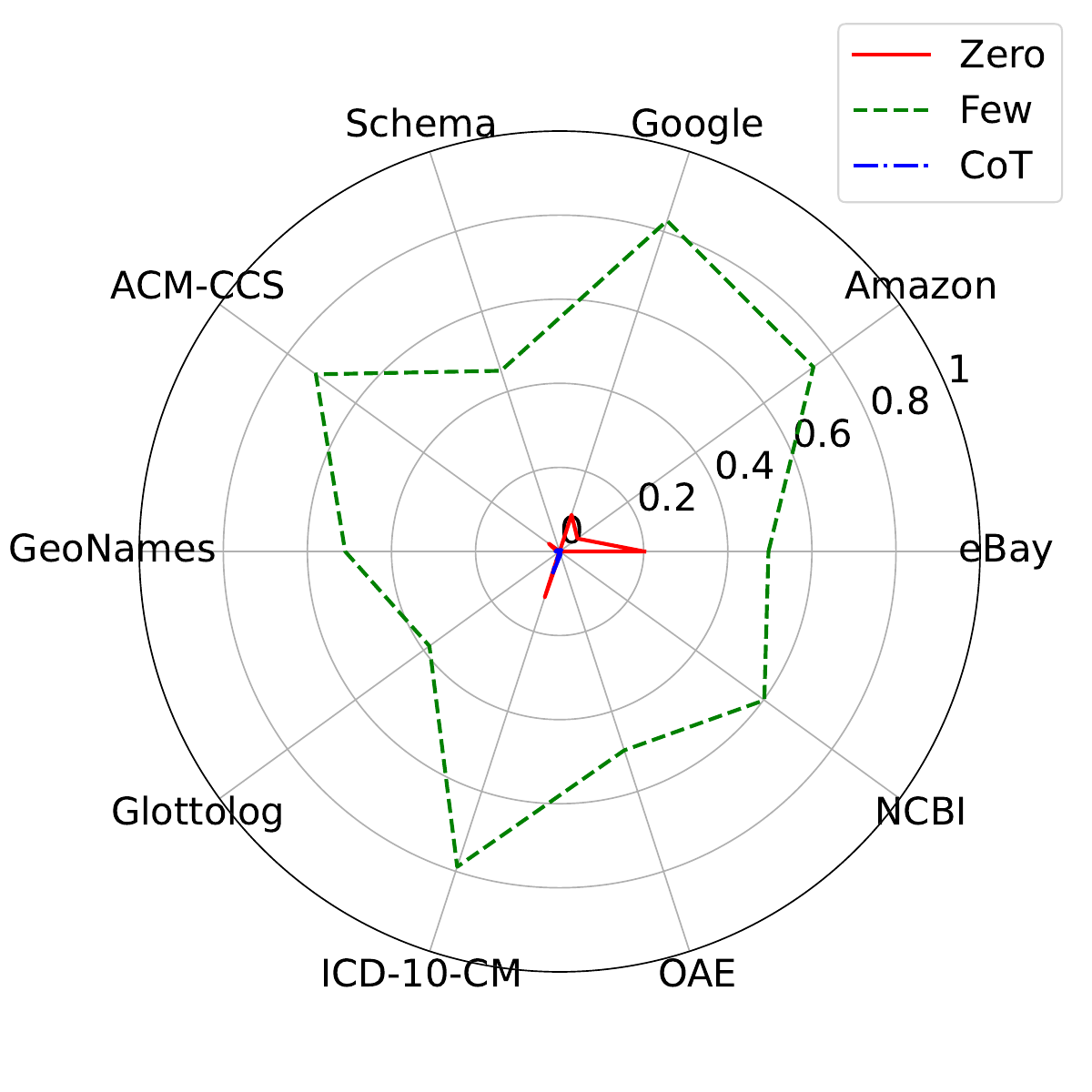}}
  \subfigure[Llama-2-7B, miss rate]{
\label{subfig:radar-llama-miss}
  \includegraphics[width=0.242\linewidth]{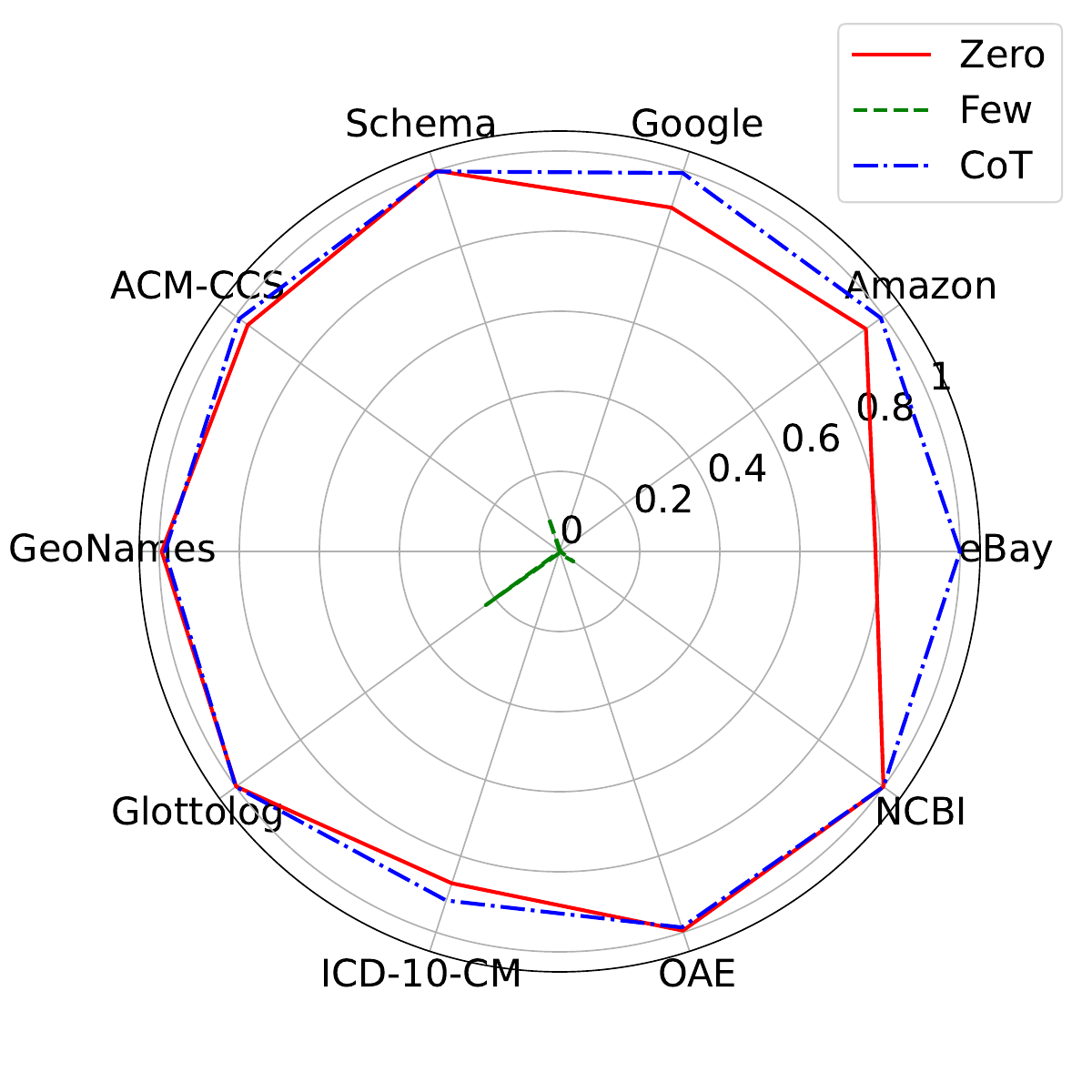}}
  \vspace{-1.5em}
  \caption{\rev{Radar charts for representative LLMs under different prompting settings in hard datasets.}}
  \label{fig:radar}
  \vspace{-1em}
\end{figure*}

\begin{figure}[t!]
  \centering
  \includegraphics[width=1.0\linewidth]{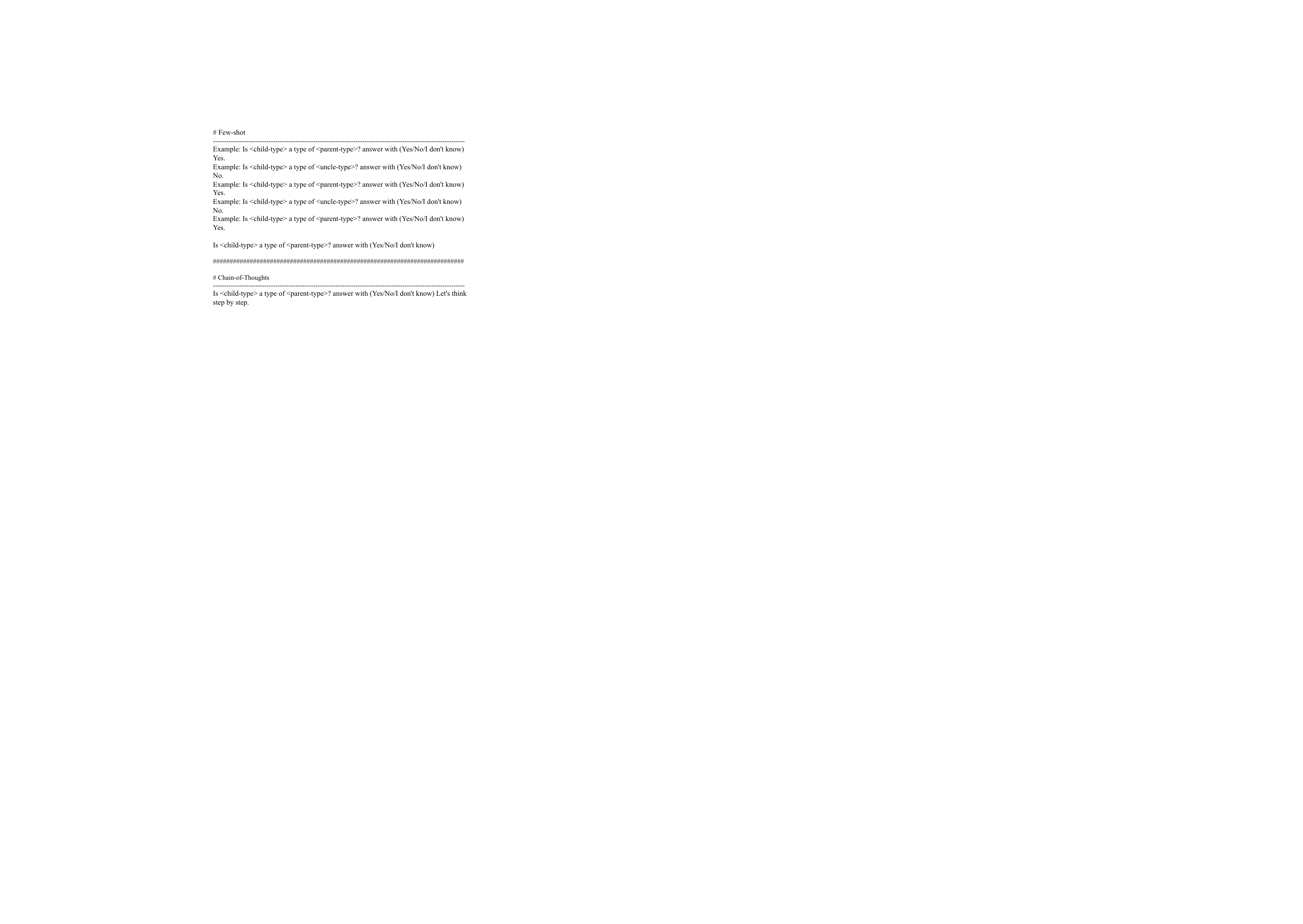}
  \vspace{-2.5em}
  \caption{\rev{Few-shot and Chain-of-Thoughts examples.}}
  \label{fig:few-cot}
  \vspace{-2em}
\end{figure}

Similar to the prompting settings adopted by previous work~\cite{huang2023c}, we introduced two additional prompting settings to further evaluate the performance of LLMs: Few-shot learning, and Chain-of-Thoughts (CoT). As shown in~\cite{touvron2023llama} and~\cite{kojima2022large}, few-shot and CoT prompting techniques can improve LLMs' performance. Therefore, we want to include these prompting settings to see if they can improve LLMs' performance on taxonomies. For the few-shot setting, following~\cite{huang2023c}, we conducted five-shot experiments. To avoid introducing bias in the examples, we sample positive and negative pairs with equal probability.
In addition, to investigate if improving the reasoning ability of LLMs enhances the performance~\cite{wei2022chain}, we conducted chain-of-thoughts (CoT) experiments following~\cite{kojima2022large} by providing an extra prompt ``Let's think step by step.'' at the end of the questions to guide LLMs through more reasoning steps. \rev{Please refer to Figure~\ref{fig:few-cot} for an example of the few-shot and CoT settings of our experiments.} 

\rev{We present the radar charts of the performance of representative LLMs in hard datasets of different taxonomies under zero-shot, few-shot, and CoT prompting settings in Figure~\ref{fig:radar}.}

\stitle{Few-Shot Prompting.}
%
Few-shot prompting can improve the performance of some LLMs in answering hierarchical structure discovery questions in different taxonomies, yet cannot significantly improve the performance of the top-performing LLMs. 
We observe that compared to zero-shot prompting, few-shot prompting reduces the miss rates of LLMs: \rev{the miss rates of Llama-2-7B reduce significantly (Figure~\ref{subfig:radar-llama-miss}), and its corresponding accuracy in turn increases (Figure~\ref{subfig:radar-llama-acc}).} Llama-2-7B benefits from the few-shot prompting: changing from scoring less than or close to $20\%$ accuracy in all taxonomies to achieving comparable performance to Flan-T5-3B on some taxonomies. We believe the miss rates reduce because few-shot prompting provides concrete question-answering pairs so that the models are more confident in imitating the prompt examples provided to generate their best guesses. However, the performance uplift of the LLMs with low miss rates is not significant. For instance, the changes to the performance of Flan-T5-11B in most taxonomies are not significant (Figure~\ref{subfig:radar-flan-t5-acc}), which implies that the effect brought by few-shot prompting mainly lies in reducing the miss rates of the LLMs instead of improving models' answering accuracy. 

\stitle{Chain-of-Thoughts (CoT).}
The CoT prompting harms the performance of some of the LLMs, but the influence brought to the top-performing LLMs is minimal. Comparing the miss rates of zero-shot and CoT prompting, we observe that by introducing CoT prompting, the miss rates of \rev{Llama-2-7B rise (Figure~\ref{subfig:radar-llama-miss}).} We attribute this phenomenon to the fact that the hierarchical structure discovery questions in taxonomies are simple-formed questions, that do not require complex reasoning, and thus CoT may not be helpful for this type of question-answering task, which coincides with the observation in a recent study that CoT is helpful for complex reasoning process~\cite{wei2022chain}. Despite the phenomenon that some LLMs are influenced by CoT prompting, we find that the performance of GPT-4 is stable under CoT prompting: remains unchanged or drops by only a small extent \rev{(Figure~\ref{subfig:radar-gpt4-acc}).}

\stitle{\colorbox{black!10}{Finding 4:}}
The performance changes brought by few-shot and CoT are minimal to the best LLMs such as GPT-4. The main effect of these prompting settings is to influence the miss rates of LLMs, instead of directly improving the accuracy.

\vspace{-0.5em}

\subsection{Instance Typing}
\label{sec:exp-instance}

Other than determining the hierarchical structures of taxonomies, which is the core task we focus on in this paper, we want to further investigate the reliability of LLMs on a taxonomy-related task: instance typing (i.e., the task of determining the types of instance entities under the leaf entities in the taxonomies.) to spark more discussion and thoughts.

\rev{We first define the instances as follows:} The instances in Google and Amazon taxonomies are defined as product names under each leaf entity: we additionally crawled the product names of each leaf entity from Google shopping~\cite{Google-shopping-site} and Browsenodes~\cite{Google-shopping-tax} websites as the instances. As for the ICD-10-CM taxonomy, we define the entities as diseases with different causes, which can be obtained as the fourth-level entities in the taxonomy. The species entities in the NCBI taxonomy are considered as instances. Moreover, we treat the leaf entity language as the instances in the Glottolog taxonomy. \rev{The leaf entity adverse events are considered as the instances for the OAE taxonomy. GeoNames and eBay taxonomies do not provide valid entities. Besides, for eBay Categories, we fail to find a proper way to crawl its product information as entities. The cases of Schema.org and ACM-CCS are complex: Schema.org and ACM-CCS do not have well-defined instances under their leaf entity concepts; besides, there is no appropriate data source for us to crawl proper entities for these taxonomies. As a result, we skip the instance typing experiment for these four taxonomies.}

\begin{figure*}[t!]
\vspace{-2em}
\centering
\subfigure[shopping-Amazon]{
\label{subfig:amazon-instance}
  \includegraphics[width=0.24\linewidth]{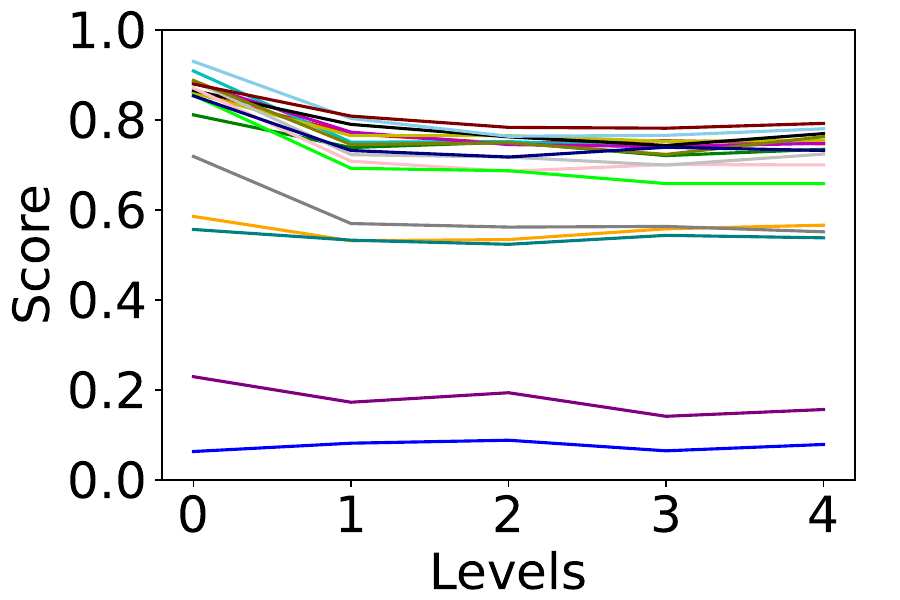}}
\subfigure[shopping-Google]{
\label{subfig:google-instance}
  \includegraphics[width=0.24\linewidth]{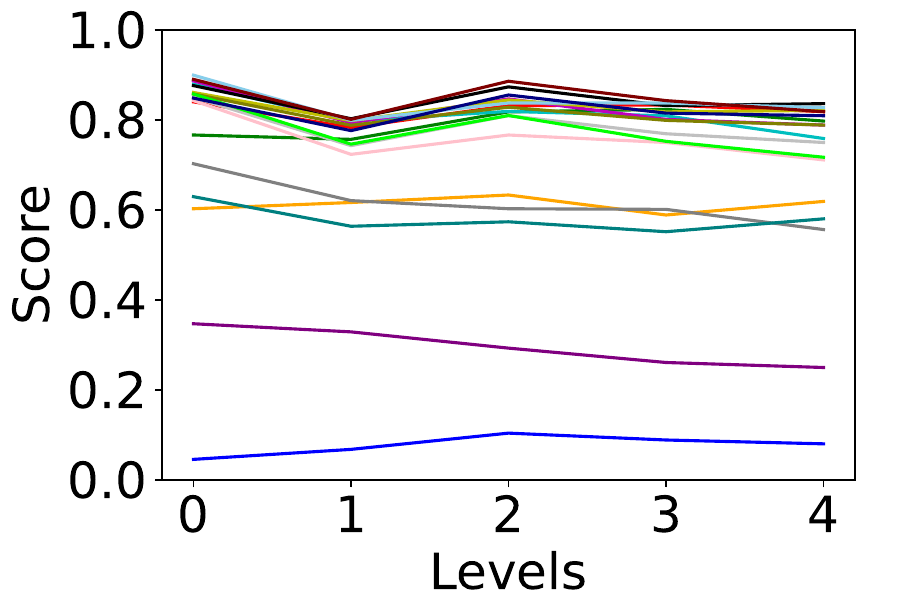}}
  \subfigure[language-Glottolog]{
\label{subfig:language-instance}
  \includegraphics[width=0.24\linewidth]{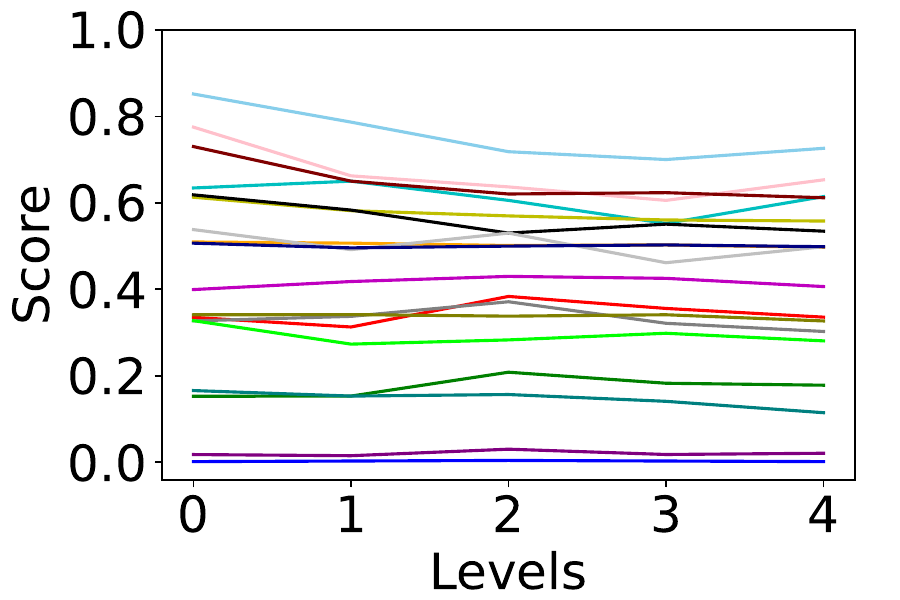}}
  \\
  \subfigure[health-ICD-10-CM]{
\label{subfig:medical-instance}
  \includegraphics[width=0.24\linewidth]{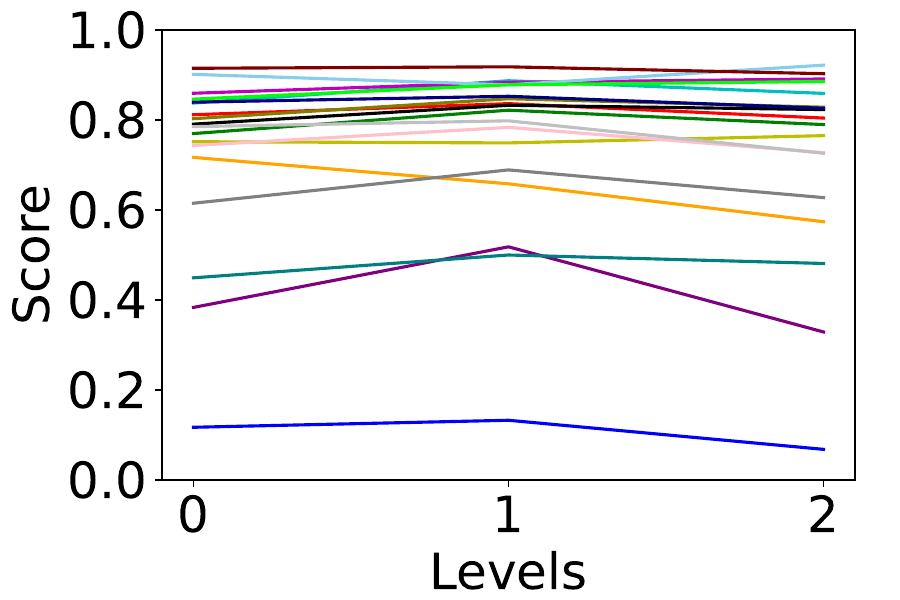}}
  \subfigure[medical-OAE]{
\label{subfig:OAE-instance}
  \includegraphics[width=0.24\linewidth]{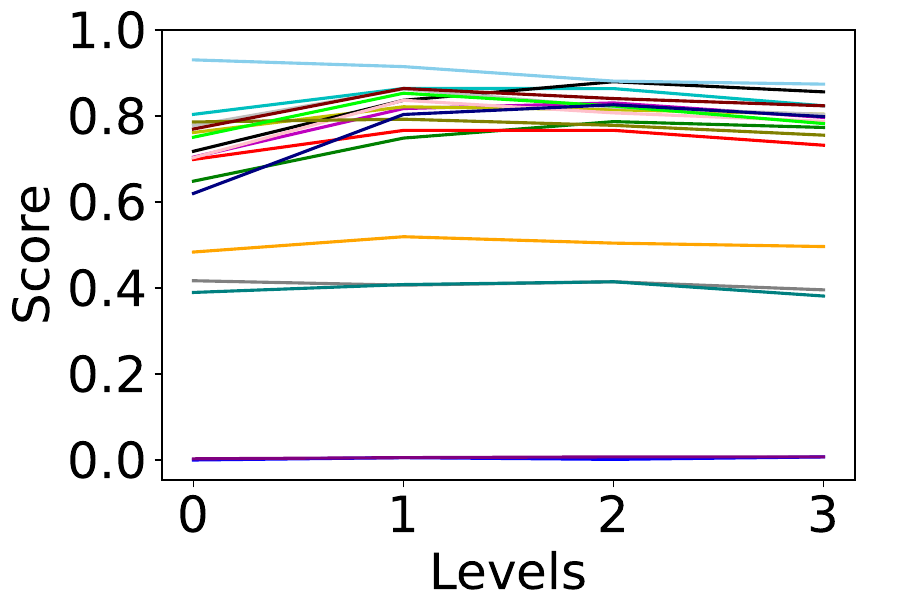}}
  \subfigure[biology-NCBI]{
\label{subfig:biology-instance}
  \includegraphics[width=0.24\linewidth]{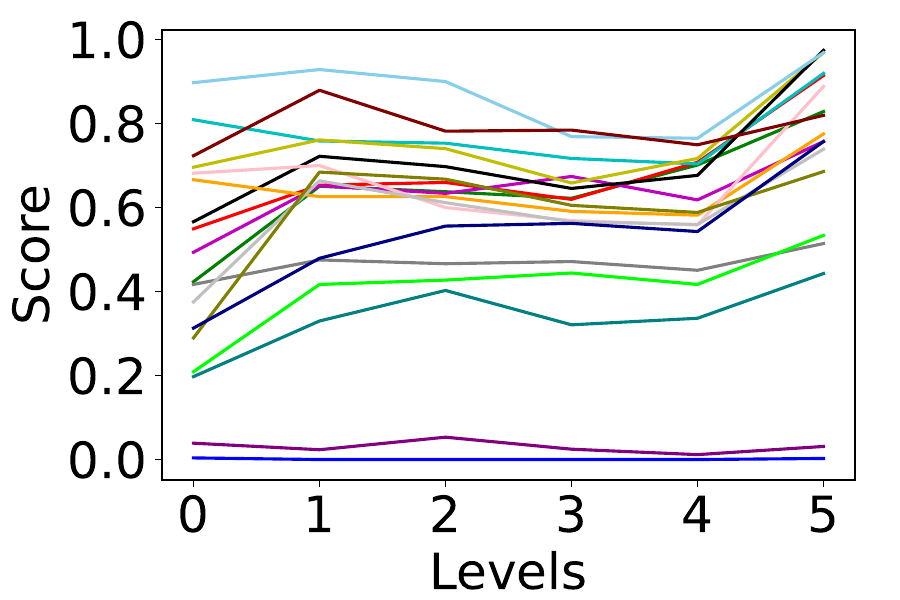}}
  \subfigure{ 
  {\includegraphics[width=0.12\linewidth]{./exp-figures/line/legend1.pdf}}}
  \subfigure{ 
  {\includegraphics[width=0.12\linewidth]{./exp-figures/line/legend2.pdf}}}
  \vspace{-2em}
  \caption{\rev{Instance typing experiments with respect to different levels of questions in hard datasets of different taxonomies.}
}
  \label{fig:instance}
  \vspace{-1em}
\end{figure*}

We adopted the same question templates as shown in Table~\ref{tab:question}. Following a similar True/False question generation manner described in Section~\ref{sec:question-design}, we generated the instance typing pairs in each level. For example, given an instance $i$, which is under an entity $e_k$ in level $k$ of taxonomy, we preserve the following instance typing pairs: $(i \rightarrow e_k)$, $(i \rightarrow e_k.p)$, ... $(i \rightarrow e_k.r)$, mark as the instance typing pairs in level $k$, $k-1$, ..., $0$, where $e_k.p$ and $e_k.r$ are the intermediate parent and root entities of the entity $e_k$. Similar to Section~\ref{sec:question-design}, we generate negative hard samples and negative easy samples. We record the performances of LLMs under the zero-shot prompting setting and present the results on hard datasets in Figure~\ref{fig:instance} due to the similar trends between easy and hard datasets. 

In general, we have the following observations: 1) Similar to the main experiments discussed in Section~\ref{sec:exp-common-to-specialized}, the performance of LLMs presents a common to specialized decline except for the ICD-10-CM and OAE taxonomies due to a similar analysis presented in Section~\ref{sec:exp-common-to-specialized}. 2) The overall instance typing performance drops as we go from root to leaf levels, except for the OAE and NCBI taxonomies whose concept names are highly overlapping near the leaf levels.

\stitle{\colorbox{black!10}{Finding 5:}}
These validate our hypothesis that LLMs are reliable in performing tasks in more common taxonomies, which means instead of manually constructing and maintaining deep and intricate taxonomies in these common domains, we can rely on LLMs to complete most of the ontology learning work. However, in specialized taxonomies, such as NCBI and Glottolog, the better practice is still relying on the traditional tree-like structures.


\vspace{-0.5em}
\section{Discussion}
\label{sec:discussion}

\subsection{The Future of Taxonomy and LLMs} \label{sec:dicussion-future}

The experimental analysis showcases that despite the integration of taxonomy knowledge within the parameters of LLMs, the coverage of their knowledge in specialized domains and deeper parts of taxonomies is still limited. Specifically, the state-of-the-art LLMs demonstrate their mastery of taxonomic knowledge in common domains such as \rev{shopping and general}, however, their performances in more specialized domains such as computer science research, biology, \rev{geography,} and language are unsatisfactory. Besides, the root-to-leaf performance decline almost happens on every LLM in most taxonomies we experimented with, which indicates insufficient coverage of knowledge in deeper levels of taxonomies.

Nevertheless, LLMs have ushered in a paradigm shift in ontology learning, prompting a fundamental reconsideration of the most effective representation of taxonomies. Our vision for the next-generation taxonomy is to combine LLMs with the traditional tree-structure taxonomy to form a novel taxonomy form, where the hierarchical knowledge implicitly resides inside LLMs' weights or explicitly presents as the traditional ``Is-A'' parent-child structure. The harmonious synthesis of these two modalities, harnessing both the cutting-edge advancement of LLMs' knowledge and the reliability of traditional tree structure, presents a captivating and fertile research domain, which we shall explore in greater depth. 

\stitle{Common Taxonomies.}
Common taxonomies that are likely to be covered by the abundant training data of LLMs, such as shopping and general, should be encoded inside the LLMs. 
Despite that in some use cases such as relation display and visualization, there might still be places for the traditional taxonomic structure near root levels to exist, the majority of the use cases (such as entity searching and knowledge reasoning) in common taxonomies can be well handled by LLMs. \rev{We provide a concrete example to study the possibility of taxonomy replacement on Amazon Product Category in Section~\ref{sec:discussion-case}.}
LLMs already demonstrate high reliability as shown in the hierarchical structure discovery and instance typing experiments for these taxonomies. The manually constructed and maintained taxonomies in these domains may not be needed shortly. The few errors (less than $25\%$) that LLMs still have in these taxonomies are likely to get addressed by performing fine-tuning based on existing studies~\cite{hu2023survey, wang2020k}.

\stitle{More Specialized Taxonomies.}
The more specialized taxonomies that domain experts generally use, such as language, computer science research, biology, and geography, are likely to remain in their current tree-structure forms or change to LLM-tree-structure-combined forms. Since the state-of-the-art LLMs are still not ready to provide reliable responses for these more specialized taxonomies, especially near the leaf levels, where the performance of LLMs gets significantly poor or fluctuates. We also discovered that although LLMs can perform well near the root levels of these more specialized taxonomies, their performance near the leaf levels can be significantly worse (Glottolog and ACM-CCS) or unstable (NCBI). Therefore, we recommend that industrial practitioners continue with the current tree-structure taxonomies in specialized domains to ensure reliability; while the research communities should start exploring the possibility of LLM-tree-structure-combined taxonomy forms: the entities near the roots are transformed into LLMs' weights, while the entities near the leaves should remain in the traditional tree-structure form to achieve both high accuracy and minimal maintaining and constructing cost for ontology learning.

\vspace{-1em}
\subsection{Limitations}
\label{sec:limitations}
\rev{As discussed in Section~\ref{sec:dicussion-future}, LLMs without instruction fine-tuning struggle to achieve satisfactory performance at low levels of the specialized taxonomies, which is indeed a limitation of LLMs for taxonomy replacement. In this sense, a possible solution is to replace the traditional taxonomies on some of the levels where the LLMs can achieve high and stable performance. Please refer to Section~\ref{sec:discussion-case} for an example of taxonomy replacement. Besides, the domain-specific instruction fine-tuning~\cite{chung2024scaling} (LLMs4OL) improves the performance of the original LLM (Flan-T5-3B) at low levels of the specialized taxonomies as shown in Figures~\ref{subfig:academic-hard-zero-acc}-\ref{subfig:biology-hard-zero-acc}, which makes it a possible alternative to resolve the limitation.} 

\rev{Despite the fact that domain-specific instruction fine-tuning requires high-quality labeled data and induces high training costs, we believe these issues could be resolved by introducing domain adaptation techniques as discussed by some pilot works~\cite{kocielnik2023can, fang2024source,han2024chatgpt}. However, the effectiveness of these techniques on taxonomies is not yet validated and we think this should be a promising future direction to explore.
}


\vspace{-0.5em}
\subsection{Case Study} \label{sec:discussion-case}
\rev{To provide a concrete example of the integration of traditional taxonomy structure and LLMs, we conducted a case study on the performance and feasibility of the integrated solution with the Amazon Product Category. We replaced the nodes in level-4 or lower of the Amazon Product Category with the Llama-2-70B model, while preserving the nodes in root to level-3 for relation display and visualization purposes. Specifically, suppose there is a level-3 concept named ``Stationery'' and has descendants ``Pen'' and ``Pencil'' and there is a customer who searches for pencil products. Then traditionally, if he/she relies on the traditional taxonomy structure, the query would match the level-4 ``Pencil'' node and then get the product list under the ``Pencil'' category. After we remove the level-4 concepts, his/her query would instead first ask about the parent concept of the query concept ``Pencil'' with an accuracy of over $70\%$ as shown in Figure~\ref{subfig:amazon-hard-zero-acc}. Then the query would find and match the level-3 concept ``Stationery'', and ask Llama-2-70B to return all the pencil products from the set of stationery products.}

\rev{The performance of such replacement is evaluated as follows: given a level 4 or lower concept $e_k$, with a list of products under $e_k$: $l_{e_k}$. We denote the list of products of the siblings of $e_k$ as $l_{e_k.s}$. We record the precision and recall of the product list $\hat{l_{e_k}}$ returned by Llama-2-70B when given the full stationery product list $\{l_{e_k} \cup l_{e_k.s}\}$. We sampled the leaf concepts with a confidence level of 95\% and a margin of error of 5\% similar to the \textbf{Question Generation} in Section~\ref{sec:question-design} to form the experimental dataset.}

\rev{By performing the replacement, we save $25777/43814=59\%$ of the construction and maintenance cost of the taxonomy, as shown in Table~\ref{tab:taxonomies}. The precision and recall of the integrated solution are $0.713$ and $0.792$ respectively. As such, by replacing the level 4 or lower concepts with Llama-2-70B, we saved $59\%$ taxonomy construction and maintenance cost, while achieving an overall precision and recall of over $70\%$. }

\rev{Note that we may replace more layers to achieve lower taxonomy construction and maintenance costs and introduce more advanced fine-tuning techniques for the LLMs or adopt ranking techniques~\cite{li2022learning, liu2009learning} to achieve better precision and recall. The case study serves as a pilot study of the feasibility of the integration of taxonomy and LLMs optimizing and refining the integration solution could be a promising research topic for the community to work on.
}

\subsection{Scalability}
\label{sec:scalability}

\begin{figure}[t!]
  \vspace{-2em}
  \centering
  \includegraphics[width=1.0\linewidth]{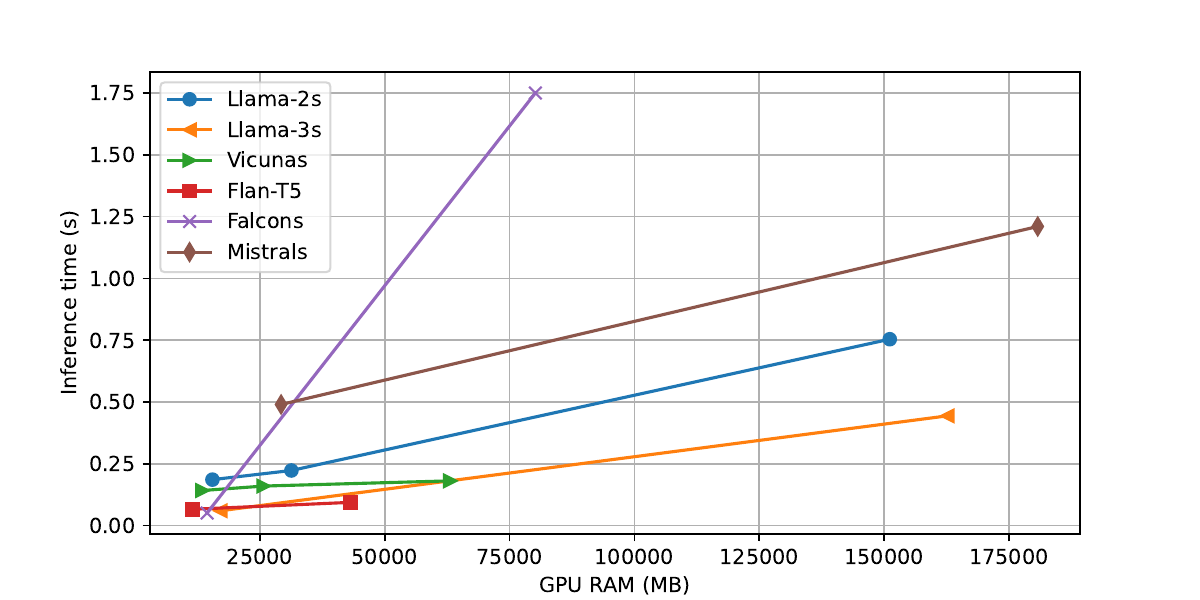}
  \vspace{-2em}
  \caption{\rev{Scalability of different model series.}}
  \label{fig:scalability}
  \vspace{-2em}
\end{figure}

\rev{We analyze the scalability of the LLM series by recording the GPU RAM and the average time costs induced by each LLM model in the corresponding LLM series during the inference of zero-shot taxonomy questions. The experimental results are presented in Figure~\ref{fig:scalability}. Specifically, we considered the scalability of six open-sourced LLM series: Llama-2s, Llama-3s, Vicunas, Flan-T5s, Falcons, and Mistrals. We observe that Flan-T5s, Vicunas, and Llama-3s present relatively good scalability, as the model size grows, the inference time does not increase significantly, which is especially important for their adoption in real-world taxonomy-related applications.}
\section{Related Work}
\label{sec:related}


\stitle{Benchmarks and experimental analysis.} 
Many QA benchmarks were developed to evaluate the ability of language models~\cite{berant2013semantic, joshi2017triviaqa, dubey2019lc, kwiatkowski2019natural, sciavolino-etal-2021-simple, yang2024crag}. However, these benchmarks do not focus on the taxonomy data and common to specialized domain knowledge. Two recent works further the study of long-tail domain knowledge by constructing new QA benchmarks~\cite{mallen-etal-2023-trust, kumar2023automatic}. Sun et al.~\cite{sun2023head} introduced Head-to-tail, a novel benchmark that systematically analyzes the factuality of LLMs over KG entities from common to long-tail. Luo et al.~\cite{luo2023systematic} proposed an automatic question generation method to generate factual questions from common to specialized domains. These works mainly analyze LLMs' performance on KG instead of taxonomy. Although a pilot work LLMs4OL~\cite{babaei2023llms4ol} explored the possibility of utilizing LLMs to perform ontology learning, the domain coverage of their evaluation is rather limited: mainly on the common and bio-medical domains, which cannot comprehensively reflect LLMs' knowledge in various taxonomies from common to specialized domains. Besides, LLMs4OL failed to provide an in-depth analysis of LLMs' performance in different levels of the taxonomies, which is indeed important for the audience to explore whether LLMs can replace taxonomies. As discussed in~\cite{sun2023head}, LLMs are less knowledgeable for the long-tail, nuanced knowledge in KGs. Intuitively, we believe a similar phenomenon should present in different levels of taxonomies, i.e., LLMs are less knowledgeable at the leaf levels of taxonomies and thus should be an issue considered if the users plan to use LLMs in replacement of the traditional taxonomies. However, no existing study on taxonomies considered this direction. Compared with existing works, \sys is the first benchmark that systematically covers the taxonomies instead of KG from common to specialized domains with in-depth root-to-leaf analysis, which is less touched by previous studies.

\stitle{LLM prompting and its settings.} 
As the major way to probe the knowledge of LLMs towards taxonomies, we would like to briefly introduce the LLM prompting methods with different prompting settings. The prompts adopted by us to evaluate the LLMs' performance over taxonomies are named prefix prompts~\cite{lester2021power, li2021prefix, 10.1145/3560815}, which are suitable for the general question-answering scenario. As discussed by~\cite{10.1145/3560815}, the methods to design prompts can be classified into manual template engineering and automated template learning. In our paper, we chose the manual template engineering approach, which is to manually design intuitive templates based on experience. The reason why we did not adopt the advanced automatic template generation approaches is that the primary focus of our work is to provide an initial analysis of LLMs' performance on taxonomies and we believe that manually crafted templates can achieve the goal of reflecting LLMs' knowledge as done by~\cite{sun2023head, petroni2019language}.

The popular prompting settings include zero-shot, few-shot, Chain-of-Thoughts, etc. Zero-shot is straightforward: the LLMs take the input question and return the corresponding answer directly~\cite{romera2015embarrassingly}. Instead of directly querying the LLMs with the question, under few-shot settings, users additionally provide several examples of questions and answers, and then query the LLM with the desired question and receive the response~\cite{schick2020s, brown2020language}. Chain-of-Thoughts (CoT) prompting~\cite{kojima2022large, wei2022chain} guides the LLMs to break down a complex reasoning question into several intermediate steps. By solving the questions step-by-step, the LLMs return more reasonable answers, especially for questions that require complex reasoning. 

\section{Conclusion}
\label{sec:conclusion}

In this paper, we introduced \sys, a novel taxonomy hierarchical structure benchmark that comprehensively evaluates the performance of LLMs over different taxonomies from common to specialized domains, from root to leaf levels. We systematically evaluated the performances of \rev{eighteen} state-of-the-art LLMs under three popular prompting settings: zero-shot, few-shot, and Chain-of-Thoughts at different levels of \rev{ten} representative taxonomies. Four highly concerned research questions were proposed and resolved and we provided valuable insights into future research opportunities for industrial users, LLM developers, and database researchers. Our comprehensive evaluation shows that even the best-performing LLM presents unsatisfactory performances at specialized taxonomies and for entities near the leaf levels. In response, we suggest future research directions to combine the LLMs with traditional taxonomies so as to create novel neural-symbolic taxonomies that have the best of both worlds. 
\clearpage
\balance
\bibliographystyle{ACM-Reference-Format}
\bibliography{sample}


\begin{thebibliography}{80}


\ifx \showCODEN    \undefined \def \showCODEN     #1{\unskip}     \fi
\ifx \showDOI      \undefined \def \showDOI       #1{#1}\fi
\ifx \showISBNx    \undefined \def \showISBNx     #1{\unskip}     \fi
\ifx \showISBNxiii \undefined \def \showISBNxiii  #1{\unskip}     \fi
\ifx \showISSN     \undefined \def \showISSN      #1{\unskip}     \fi
\ifx \showLCCN     \undefined \def \showLCCN      #1{\unskip}     \fi
\ifx \shownote     \undefined \def \shownote      #1{#1}          \fi
\ifx \showarticletitle \undefined \def \showarticletitle #1{#1}   \fi
\ifx \showURL      \undefined \def \showURL       {\relax}        \fi
\providecommand\bibfield[2]{#2}
\providecommand\bibinfo[2]{#2}
\providecommand\natexlab[1]{#1}
\providecommand\showeprint[2][]{arXiv:#2}

\bibitem[\protect\citeauthoryear{??}{ccs}{2012a}]%
        {ccs-concept-code}
 \bibinfo{year}{2012}\natexlab{a}.
\newblock \bibinfo{title}{ACM CCS Concept 2012 link}.
\newblock
\newblock
\urldef\tempurl%
\url{https://dl.acm.org/pb-assets/dl_ccs/acm_ccs2012-1626988337597.xml}
\showURL{%
Retrieved Jan 10, 2024 from \tempurl}


\bibitem[\protect\citeauthoryear{??}{ccs}{2012b}]%
        {ccs-concept}
 \bibinfo{year}{2012}\natexlab{b}.
\newblock \bibinfo{title}{ACM Computing Classification System}.
\newblock
\newblock
\urldef\tempurl%
\url{https://dl.acm.org/ccs}
\showURL{%
Retrieved Jan 10, 2024 from \tempurl}


\bibitem[\protect\citeauthoryear{??}{Ama}{2019}]%
        {Amazon-shopping-tax}
 \bibinfo{year}{2019}\natexlab{}.
\newblock \bibinfo{title}{Amazon's Product Category}.
\newblock
\newblock
\urldef\tempurl%
\url{https://www.browsenodes.com/}
\showURL{%
Retrieved Jan 10, 2024 from \tempurl}


\bibitem[\protect\citeauthoryear{??}{Goo}{2021}]%
        {Google-shopping-tax}
 \bibinfo{year}{2021}\natexlab{}.
\newblock \bibinfo{title}{Google Product Category}.
\newblock
\newblock
\urldef\tempurl%
\url{https://www.google.com/basepages/producttype/taxonomy.en-US.txt}
\showURL{%
Retrieved Jan 10, 2024 from \tempurl}


\bibitem[\protect\citeauthoryear{??}{ICD}{2021}]%
        {ICD-python}
 \bibinfo{year}{2021}\natexlab{}.
\newblock \bibinfo{title}{ICD-10-CM package}.
\newblock
\newblock
\urldef\tempurl%
\url{https://pypi.org/project/simple-icd-10/}
\showURL{%
Retrieved Jan 10, 2024 from \tempurl}


\bibitem[\protect\citeauthoryear{??}{Goo}{2022}]%
        {Google-shopping-stat}
 \bibinfo{year}{2022}\natexlab{}.
\newblock \bibinfo{title}{Google Shopping Statistics}.
\newblock
\newblock
\urldef\tempurl%
\url{https://www.statista.com/statistics/1341380/most-well-known-price-comparison-portals-in-the-united-states/}
\showURL{%
Retrieved Jan 10, 2024 from \tempurl}


\bibitem[\protect\citeauthoryear{??}{OAE}{2022}]%
        {OAE}
 \bibinfo{year}{2022}\natexlab{}.
\newblock \bibinfo{title}{OAE-website}.
\newblock
\newblock
\urldef\tempurl%
\url{https://bioportal.bioontology.org/ontologies/OAE}
\showURL{%
Retrieved Apr 17, 2024 from \tempurl}


\bibitem[\protect\citeauthoryear{??}{Ama}{2023}]%
        {Amazon-shopping-stat}
 \bibinfo{year}{2023}\natexlab{}.
\newblock \bibinfo{title}{Amazon's Product Category statistics}.
\newblock
\newblock
\urldef\tempurl%
\url{https://www.statista.com/forecasts/997230/most-popular-online-shops-in-the-us}
\showURL{%
Retrieved Jan 10, 2024 from \tempurl}


\bibitem[\protect\citeauthoryear{??}{Glo}{2023}]%
        {Glottolog-release}
 \bibinfo{year}{2023}\natexlab{}.
\newblock \bibinfo{title}{Glottolog-4.8}.
\newblock
\newblock
\urldef\tempurl%
\url{https://glottolog.org/meta/downloads}
\showURL{%
Retrieved Jan 10, 2024 from \tempurl}


\bibitem[\protect\citeauthoryear{??}{ICD}{2023a}]%
        {ICD-public}
 \bibinfo{year}{2023}\natexlab{a}.
\newblock \bibinfo{title}{ICD-10-CM for public}.
\newblock
\newblock
\urldef\tempurl%
\url{https://www.verywellhealth.com/icd-10-codes-5271405}
\showURL{%
Retrieved Jan 10, 2024 from \tempurl}


\bibitem[\protect\citeauthoryear{??}{ICD}{2023b}]%
        {ICD-tax}
 \bibinfo{year}{2023}\natexlab{b}.
\newblock \bibinfo{title}{ICD-10-CM taxonomy information}.
\newblock
\newblock
\urldef\tempurl%
\url{https://www.cdc.gov/nchs/icd/icd-10-cm.htm}
\showURL{%
Retrieved Jan 10, 2024 from \tempurl}


\bibitem[\protect\citeauthoryear{??}{ncb}{2023}]%
        {ncbi-data}
 \bibinfo{year}{2023}\natexlab{}.
\newblock \bibinfo{title}{NCBI data download}.
\newblock
\newblock
\urldef\tempurl%
\url{https://ftp.ncbi.nlm.nih.gov/pub/taxonomy/}
\showURL{%
Retrieved Jan 10, 2024 from \tempurl}


\bibitem[\protect\citeauthoryear{??}{Qua}{2023}]%
        {Qualtrics}
 \bibinfo{year}{2023}\natexlab{}.
\newblock \bibinfo{title}{Qualtrics}.
\newblock
\newblock
\urldef\tempurl%
\url{https://www.qualtrics.com/au/experience-management/research/determine-sample-size/?rid=ip&prevsite=en&newsite=au&geo=HK&geomatch=au}
\showURL{%
Retrieved Jan 10, 2024 from \tempurl}


\bibitem[\protect\citeauthoryear{??}{Cla}{2024}]%
        {Claude-v3-documentation}
 \bibinfo{year}{2024}\natexlab{}.
\newblock \bibinfo{title}{Claude-v3-documentation}.
\newblock
\newblock
\urldef\tempurl%
\url{https://www.anthropic.com/news/claude-3-family}
\showURL{%
Retrieved Apr 18, 2024 from \tempurl}


\bibitem[\protect\citeauthoryear{??}{eba}{2024}]%
        {ebay}
 \bibinfo{year}{2024}\natexlab{}.
\newblock \bibinfo{title}{eBay}.
\newblock
\newblock
\urldef\tempurl%
\url{https://www.ebay.com/n/all-categories}
\showURL{%
Retrieved Apr 17, 2024 from \tempurl}


\bibitem[\protect\citeauthoryear{??}{geo}{2024}]%
        {geonames}
 \bibinfo{year}{2024}\natexlab{}.
\newblock \bibinfo{title}{geonames-website}.
\newblock
\newblock
\urldef\tempurl%
\url{https://www.geonames.org/export/codes.html}
\showURL{%
Retrieved Apr 17, 2024 from \tempurl}


\bibitem[\protect\citeauthoryear{??}{Goo}{2024}]%
        {Google-shopping-site}
 \bibinfo{year}{2024}\natexlab{}.
\newblock \bibinfo{title}{Google Shopping Website}.
\newblock
\newblock
\urldef\tempurl%
\url{https://shopping.google.com/}
\showURL{%
Retrieved Jan 10, 2024 from \tempurl}


\bibitem[\protect\citeauthoryear{??}{llm}{2024}]%
        {llms4ol-code}
 \bibinfo{year}{2024}\natexlab{}.
\newblock \bibinfo{title}{LLMs4OL-code}.
\newblock
\newblock
\urldef\tempurl%
\url{https://github.com/HamedBabaei/LLMs4OL/tree/main/TaskB}
\showURL{%
Retrieved Apr 17, 2024 from \tempurl}


\bibitem[\protect\citeauthoryear{??}{sch}{2024}]%
        {schemaorg}
 \bibinfo{year}{2024}\natexlab{}.
\newblock \bibinfo{title}{schema-website}.
\newblock
\newblock
\urldef\tempurl%
\url{https://github.com/schemaorg/schemaorg/blob/main/data/releases/26.0/schemaorg-current-https-types.csv}
\showURL{%
Retrieved Apr 17, 2024 from \tempurl}


\bibitem[\protect\citeauthoryear{??}{tax}{2024}]%
        {taxo-exp}
 \bibinfo{year}{2024}\natexlab{}.
\newblock \bibinfo{title}{TaxoGlimpse experimental results}.
\newblock
\newblock
\urldef\tempurl%
\url{https://github.com/ysunbp/TaxoGlimpse/tree/main/exp-results}
\showURL{%
Retrieved Apr 17, 2024 from \tempurl}


\bibitem[\protect\citeauthoryear{Achiam, Adler, Agarwal, Ahmad, Akkaya, Aleman, Almeida, Altenschmidt, Altman, Anadkat, et~al\mbox{.}}{Achiam et~al\mbox{.}}{2023}]%
        {achiam2023gpt}
\bibfield{author}{\bibinfo{person}{Josh Achiam}, \bibinfo{person}{Steven Adler}, \bibinfo{person}{Sandhini Agarwal}, \bibinfo{person}{Lama Ahmad}, \bibinfo{person}{Ilge Akkaya}, \bibinfo{person}{Florencia~Leoni Aleman}, \bibinfo{person}{Diogo Almeida}, \bibinfo{person}{Janko Altenschmidt}, \bibinfo{person}{Sam Altman}, \bibinfo{person}{Shyamal Anadkat}, {et~al\mbox{.}}} \bibinfo{year}{2023}\natexlab{}.
\newblock \showarticletitle{GPT-4 Technical Report}.
\newblock \bibinfo{journal}{\emph{arXiv preprint arXiv:2303.08774}} (\bibinfo{year}{2023}).
\newblock


\bibitem[\protect\citeauthoryear{AI@Meta}{AI@Meta}{2024}]%
        {llama3modelcard}
\bibfield{author}{\bibinfo{person}{AI@Meta}.} \bibinfo{year}{2024}\natexlab{}.
\newblock \showarticletitle{Llama 3 Model Card}.
\newblock  (\bibinfo{year}{2024}).
\newblock
\urldef\tempurl%
\url{https://github.com/meta-llama/llama3/blob/main/MODEL_CARD.md}
\showURL{%
\tempurl}


\bibitem[\protect\citeauthoryear{Almazrouei, Alobeidli, Alshamsi, Cappelli, Cojocaru, Debbah, Goffinet, Hesslow, Launay, Malartic, et~al\mbox{.}}{Almazrouei et~al\mbox{.}}{2023}]%
        {almazrouei2023falcon}
\bibfield{author}{\bibinfo{person}{Ebtesam Almazrouei}, \bibinfo{person}{Hamza Alobeidli}, \bibinfo{person}{Abdulaziz Alshamsi}, \bibinfo{person}{Alessandro Cappelli}, \bibinfo{person}{Ruxandra Cojocaru}, \bibinfo{person}{M{\'e}rouane Debbah}, \bibinfo{person}{{\'E}tienne Goffinet}, \bibinfo{person}{Daniel Hesslow}, \bibinfo{person}{Julien Launay}, \bibinfo{person}{Quentin Malartic}, {et~al\mbox{.}}} \bibinfo{year}{2023}\natexlab{}.
\newblock \showarticletitle{The falcon series of open language models}.
\newblock \bibinfo{journal}{\emph{arXiv preprint arXiv:2311.16867}} (\bibinfo{year}{2023}).
\newblock


\bibitem[\protect\citeauthoryear{Babaei~Giglou, D’Souza, and Auer}{Babaei~Giglou et~al\mbox{.}}{2023}]%
        {babaei2023llms4ol}
\bibfield{author}{\bibinfo{person}{Hamed Babaei~Giglou}, \bibinfo{person}{Jennifer D’Souza}, {and} \bibinfo{person}{S{\"o}ren Auer}.} \bibinfo{year}{2023}\natexlab{}.
\newblock \showarticletitle{LLMs4OL: Large language models for ontology learning}. In \bibinfo{booktitle}{\emph{International Semantic Web Conference}}. Springer, \bibinfo{pages}{408--427}.
\newblock


\bibitem[\protect\citeauthoryear{Bang, Cahyawijaya, Lee, Dai, Su, Wilie, Lovenia, Ji, Yu, Chung, et~al\mbox{.}}{Bang et~al\mbox{.}}{2023}]%
        {bang2023multitask}
\bibfield{author}{\bibinfo{person}{Yejin Bang}, \bibinfo{person}{Samuel Cahyawijaya}, \bibinfo{person}{Nayeon Lee}, \bibinfo{person}{Wenliang Dai}, \bibinfo{person}{Dan Su}, \bibinfo{person}{Bryan Wilie}, \bibinfo{person}{Holy Lovenia}, \bibinfo{person}{Ziwei Ji}, \bibinfo{person}{Tiezheng Yu}, \bibinfo{person}{Willy Chung}, {et~al\mbox{.}}} \bibinfo{year}{2023}\natexlab{}.
\newblock \showarticletitle{A multitask, multilingual, multimodal evaluation of chatgpt on reasoning, hallucination, and interactivity}.
\newblock \bibinfo{journal}{\emph{arXiv preprint arXiv:2302.04023}} (\bibinfo{year}{2023}).
\newblock


\bibitem[\protect\citeauthoryear{Berant, Chou, Frostig, and Liang}{Berant et~al\mbox{.}}{2013}]%
        {berant2013semantic}
\bibfield{author}{\bibinfo{person}{Jonathan Berant}, \bibinfo{person}{Andrew Chou}, \bibinfo{person}{Roy Frostig}, {and} \bibinfo{person}{Percy Liang}.} \bibinfo{year}{2013}\natexlab{}.
\newblock \showarticletitle{Semantic parsing on freebase from question-answer pairs}. In \bibinfo{booktitle}{\emph{Proceedings of the 2013 conference on empirical methods in natural language processing}}. \bibinfo{pages}{1533--1544}.
\newblock


\bibitem[\protect\citeauthoryear{Brown, Mann, Ryder, Subbiah, Kaplan, Dhariwal, Neelakantan, Shyam, Sastry, Askell, et~al\mbox{.}}{Brown et~al\mbox{.}}{2020}]%
        {brown2020language}
\bibfield{author}{\bibinfo{person}{Tom Brown}, \bibinfo{person}{Benjamin Mann}, \bibinfo{person}{Nick Ryder}, \bibinfo{person}{Melanie Subbiah}, \bibinfo{person}{Jared~D Kaplan}, \bibinfo{person}{Prafulla Dhariwal}, \bibinfo{person}{Arvind Neelakantan}, \bibinfo{person}{Pranav Shyam}, \bibinfo{person}{Girish Sastry}, \bibinfo{person}{Amanda Askell}, {et~al\mbox{.}}} \bibinfo{year}{2020}\natexlab{}.
\newblock \showarticletitle{Language models are few-shot learners}.
\newblock \bibinfo{journal}{\emph{Advances in neural information processing systems}}  \bibinfo{volume}{33} (\bibinfo{year}{2020}), \bibinfo{pages}{1877--1901}.
\newblock


\bibitem[\protect\citeauthoryear{Caines, Bentz, Alikaniotis, Katushemererwe, and Buttery}{Caines et~al\mbox{.}}{2016}]%
        {caines2016glottolog}
\bibfield{author}{\bibinfo{person}{Andrew Caines}, \bibinfo{person}{Christian Bentz}, \bibinfo{person}{Dimitrios Alikaniotis}, \bibinfo{person}{Fridah Katushemererwe}, {and} \bibinfo{person}{Paula Buttery}.} \bibinfo{year}{2016}\natexlab{}.
\newblock \showarticletitle{The Glottolog data explorer: Mapping the world’s languages}.
\newblock \bibinfo{journal}{\emph{Proceedings of VisLR II: Visualization as Added Value in the Development, Use and Evaluation of Language Resources}} (\bibinfo{year}{2016}), \bibinfo{pages}{38--53}.
\newblock


\bibitem[\protect\citeauthoryear{Chiang, Li, Lin, Sheng, Wu, Zhang, Zheng, Zhuang, Zhuang, Gonzalez, et~al\mbox{.}}{Chiang et~al\mbox{.}}{2023}]%
        {chiang2023vicuna}
\bibfield{author}{\bibinfo{person}{Wei-Lin Chiang}, \bibinfo{person}{Zhuohan Li}, \bibinfo{person}{Zi Lin}, \bibinfo{person}{Ying Sheng}, \bibinfo{person}{Zhanghao Wu}, \bibinfo{person}{Hao Zhang}, \bibinfo{person}{Lianmin Zheng}, \bibinfo{person}{Siyuan Zhuang}, \bibinfo{person}{Yonghao Zhuang}, \bibinfo{person}{Joseph~E Gonzalez}, {et~al\mbox{.}}} \bibinfo{year}{2023}\natexlab{}.
\newblock \showarticletitle{Vicuna: An open-source chatbot impressing gpt-4 with 90\%* chatgpt quality}.
\newblock \bibinfo{journal}{\emph{See https://vicuna. lmsys. org (accessed 14 Apr 2023)}} (\bibinfo{year}{2023}).
\newblock


\bibitem[\protect\citeauthoryear{Choi}{Choi}{2023}]%
        {choi2023common}
\bibfield{author}{\bibinfo{person}{Yejin Choi}.} \bibinfo{year}{2023}\natexlab{}.
\newblock \showarticletitle{Common Sense: The Dark Matter of Language and Intelligence}. In \bibinfo{booktitle}{\emph{Proceedings of the 2023 International Conference on Autonomous Agents and Multiagent Systems}}. \bibinfo{pages}{2--2}.
\newblock


\bibitem[\protect\citeauthoryear{Chung, Hou, Longpre, Zoph, Tay, Fedus, Li, Wang, Dehghani, Brahma, et~al\mbox{.}}{Chung et~al\mbox{.}}{2022}]%
        {chung2022scaling}
\bibfield{author}{\bibinfo{person}{Hyung~Won Chung}, \bibinfo{person}{Le Hou}, \bibinfo{person}{Shayne Longpre}, \bibinfo{person}{Barret Zoph}, \bibinfo{person}{Yi Tay}, \bibinfo{person}{William Fedus}, \bibinfo{person}{Yunxuan Li}, \bibinfo{person}{Xuezhi Wang}, \bibinfo{person}{Mostafa Dehghani}, \bibinfo{person}{Siddhartha Brahma}, {et~al\mbox{.}}} \bibinfo{year}{2022}\natexlab{}.
\newblock \showarticletitle{Scaling instruction-finetuned language models}.
\newblock \bibinfo{journal}{\emph{arXiv preprint arXiv:2210.11416}} (\bibinfo{year}{2022}).
\newblock


\bibitem[\protect\citeauthoryear{Chung, Hou, Longpre, Zoph, Tay, Fedus, Li, Wang, Dehghani, Brahma, et~al\mbox{.}}{Chung et~al\mbox{.}}{2024}]%
        {chung2024scaling}
\bibfield{author}{\bibinfo{person}{Hyung~Won Chung}, \bibinfo{person}{Le Hou}, \bibinfo{person}{Shayne Longpre}, \bibinfo{person}{Barret Zoph}, \bibinfo{person}{Yi Tay}, \bibinfo{person}{William Fedus}, \bibinfo{person}{Yunxuan Li}, \bibinfo{person}{Xuezhi Wang}, \bibinfo{person}{Mostafa Dehghani}, \bibinfo{person}{Siddhartha Brahma}, {et~al\mbox{.}}} \bibinfo{year}{2024}\natexlab{}.
\newblock \showarticletitle{Scaling instruction-finetuned language models}.
\newblock \bibinfo{journal}{\emph{Journal of Machine Learning Research}} \bibinfo{volume}{25}, \bibinfo{number}{70} (\bibinfo{year}{2024}), \bibinfo{pages}{1--53}.
\newblock


\bibitem[\protect\citeauthoryear{Dubey, Banerjee, Abdelkawi, and Lehmann}{Dubey et~al\mbox{.}}{2019}]%
        {dubey2019lc}
\bibfield{author}{\bibinfo{person}{Mohnish Dubey}, \bibinfo{person}{Debayan Banerjee}, \bibinfo{person}{Abdelrahman Abdelkawi}, {and} \bibinfo{person}{Jens Lehmann}.} \bibinfo{year}{2019}\natexlab{}.
\newblock \showarticletitle{Lc-quad 2.0: A large dataset for complex question answering over wikidata and dbpedia}. In \bibinfo{booktitle}{\emph{The Semantic Web--ISWC 2019: 18th International Semantic Web Conference, Auckland, New Zealand, October 26--30, 2019, Proceedings, Part II 18}}. Springer, \bibinfo{pages}{69--78}.
\newblock


\bibitem[\protect\citeauthoryear{Fang, Yap, Lin, Zhu, and Liu}{Fang et~al\mbox{.}}{2024}]%
        {fang2024source}
\bibfield{author}{\bibinfo{person}{Yuqi Fang}, \bibinfo{person}{Pew-Thian Yap}, \bibinfo{person}{Weili Lin}, \bibinfo{person}{Hongtu Zhu}, {and} \bibinfo{person}{Mingxia Liu}.} \bibinfo{year}{2024}\natexlab{}.
\newblock \showarticletitle{Source-free unsupervised domain adaptation: A survey}.
\newblock \bibinfo{journal}{\emph{Neural Networks}} (\bibinfo{year}{2024}), \bibinfo{pages}{106230}.
\newblock


\bibitem[\protect\citeauthoryear{Federhen}{Federhen}{2012}]%
        {federhen2012ncbi}
\bibfield{author}{\bibinfo{person}{Scott Federhen}.} \bibinfo{year}{2012}\natexlab{}.
\newblock \showarticletitle{The NCBI taxonomy database}.
\newblock \bibinfo{journal}{\emph{Nucleic acids research}} \bibinfo{volume}{40}, \bibinfo{number}{D1} (\bibinfo{year}{2012}), \bibinfo{pages}{D136--D143}.
\newblock


\bibitem[\protect\citeauthoryear{Guha, Brickley, and Macbeth}{Guha et~al\mbox{.}}{2016}]%
        {guha2016schema}
\bibfield{author}{\bibinfo{person}{Ramanathan~V Guha}, \bibinfo{person}{Dan Brickley}, {and} \bibinfo{person}{Steve Macbeth}.} \bibinfo{year}{2016}\natexlab{}.
\newblock \showarticletitle{Schema. org: evolution of structured data on the web}.
\newblock \bibinfo{journal}{\emph{Commun. ACM}} \bibinfo{volume}{59}, \bibinfo{number}{2} (\bibinfo{year}{2016}), \bibinfo{pages}{44--51}.
\newblock


\bibitem[\protect\citeauthoryear{Hammarstr{\"o}m, Forkel, Haspelmath, and Bank}{Hammarstr{\"o}m et~al\mbox{.}}{2023}]%
        {hammarstrom2023glottolog}
\bibfield{author}{\bibinfo{person}{Harald Hammarstr{\"o}m}, \bibinfo{person}{Robert Forkel}, \bibinfo{person}{Martin Haspelmath}, {and} \bibinfo{person}{Sebastian Bank}.} \bibinfo{year}{2023}\natexlab{}.
\newblock \showarticletitle{Glottolog 4.8}.
\newblock  (\bibinfo{year}{2023}).
\newblock


\bibitem[\protect\citeauthoryear{Hammarström and Forkel}{Hammarström and Forkel}{2022}]%
        {swj-glottocodes}
\bibfield{author}{\bibinfo{person}{Harald Hammarström} {and} \bibinfo{person}{Robert Forkel}.} \bibinfo{year}{2022}\natexlab{}.
\newblock \showarticletitle{Glottocodes: Identifiers Linking Families, Languages and Dialects to Comprehensive Reference Information}.
\newblock \bibinfo{journal}{\emph{Semantic Web Journal}} \bibinfo{volume}{13}, \bibinfo{number}{6} (\bibinfo{year}{2022}), \bibinfo{pages}{917--924}.
\newblock
\urldef\tempurl%
\url{https://content.iospress.com/articles/semantic-web/sw212843}
\showURL{%
\tempurl}


\bibitem[\protect\citeauthoryear{Han, Kocielnik, Saravanan, Jiang, Sharir, and Anandkumar}{Han et~al\mbox{.}}{2024}]%
        {han2024chatgpt}
\bibfield{author}{\bibinfo{person}{Pengrui Han}, \bibinfo{person}{Rafal Kocielnik}, \bibinfo{person}{Adhithya Saravanan}, \bibinfo{person}{Roy Jiang}, \bibinfo{person}{Or Sharir}, {and} \bibinfo{person}{Anima Anandkumar}.} \bibinfo{year}{2024}\natexlab{}.
\newblock \showarticletitle{ChatGPT Based Data Augmentation for Improved Parameter-Efficient Debiasing of LLMs}.
\newblock \bibinfo{journal}{\emph{arXiv preprint arXiv:2402.11764}} (\bibinfo{year}{2024}).
\newblock


\bibitem[\protect\citeauthoryear{He, Sarntivijai, Lin, Xiang, Guo, Zhang, Jagannathan, Toldo, Tao, and Smith}{He et~al\mbox{.}}{2014}]%
        {he2014oae}
\bibfield{author}{\bibinfo{person}{Yongqun He}, \bibinfo{person}{Sirarat Sarntivijai}, \bibinfo{person}{Yu Lin}, \bibinfo{person}{Zuoshuang Xiang}, \bibinfo{person}{Abra Guo}, \bibinfo{person}{Shelley Zhang}, \bibinfo{person}{Desikan Jagannathan}, \bibinfo{person}{Luca Toldo}, \bibinfo{person}{Cui Tao}, {and} \bibinfo{person}{Barry Smith}.} \bibinfo{year}{2014}\natexlab{}.
\newblock \showarticletitle{OAE: the ontology of adverse events}.
\newblock \bibinfo{journal}{\emph{Journal of biomedical semantics}}  \bibinfo{volume}{5} (\bibinfo{year}{2014}), \bibinfo{pages}{1--13}.
\newblock


\bibitem[\protect\citeauthoryear{Hu, Liu, Zhao, Hou, Nie, and Li}{Hu et~al\mbox{.}}{2023}]%
        {hu2023survey}
\bibfield{author}{\bibinfo{person}{Linmei Hu}, \bibinfo{person}{Zeyi Liu}, \bibinfo{person}{Ziwang Zhao}, \bibinfo{person}{Lei Hou}, \bibinfo{person}{Liqiang Nie}, {and} \bibinfo{person}{Juanzi Li}.} \bibinfo{year}{2023}\natexlab{}.
\newblock \showarticletitle{A survey of knowledge enhanced pre-trained language models}.
\newblock \bibinfo{journal}{\emph{IEEE Transactions on Knowledge and Data Engineering}} (\bibinfo{year}{2023}).
\newblock


\bibitem[\protect\citeauthoryear{Huang, Ren, Zhao, He, Wen, and Dong}{Huang et~al\mbox{.}}{2019}]%
        {huang2019taxonomy}
\bibfield{author}{\bibinfo{person}{Jin Huang}, \bibinfo{person}{Zhaochun Ren}, \bibinfo{person}{Wayne~Xin Zhao}, \bibinfo{person}{Gaole He}, \bibinfo{person}{Ji-Rong Wen}, {and} \bibinfo{person}{Daxiang Dong}.} \bibinfo{year}{2019}\natexlab{}.
\newblock \showarticletitle{Taxonomy-aware multi-hop reasoning networks for sequential recommendation}. In \bibinfo{booktitle}{\emph{Proceedings of the twelfth ACM international conference on web search and data mining}}. \bibinfo{pages}{573--581}.
\newblock


\bibitem[\protect\citeauthoryear{Huang, Bai, Zhu, Zhang, Zhang, Su, Liu, Lv, Zhang, Lei, et~al\mbox{.}}{Huang et~al\mbox{.}}{2023}]%
        {huang2023c}
\bibfield{author}{\bibinfo{person}{Yuzhen Huang}, \bibinfo{person}{Yuzhuo Bai}, \bibinfo{person}{Zhihao Zhu}, \bibinfo{person}{Junlei Zhang}, \bibinfo{person}{Jinghan Zhang}, \bibinfo{person}{Tangjun Su}, \bibinfo{person}{Junteng Liu}, \bibinfo{person}{Chuancheng Lv}, \bibinfo{person}{Yikai Zhang}, \bibinfo{person}{Jiayi Lei}, {et~al\mbox{.}}} \bibinfo{year}{2023}\natexlab{}.
\newblock \showarticletitle{C-eval: A multi-level multi-discipline chinese evaluation suite for foundation models}.
\newblock \bibinfo{journal}{\emph{arXiv preprint arXiv:2305.08322}} (\bibinfo{year}{2023}).
\newblock


\bibitem[\protect\citeauthoryear{Jiang, Sablayrolles, Mensch, Bamford, Chaplot, Casas, Bressand, Lengyel, Lample, Saulnier, et~al\mbox{.}}{Jiang et~al\mbox{.}}{2023}]%
        {jiang2023mistral}
\bibfield{author}{\bibinfo{person}{Albert~Q Jiang}, \bibinfo{person}{Alexandre Sablayrolles}, \bibinfo{person}{Arthur Mensch}, \bibinfo{person}{Chris Bamford}, \bibinfo{person}{Devendra~Singh Chaplot}, \bibinfo{person}{Diego de~las Casas}, \bibinfo{person}{Florian Bressand}, \bibinfo{person}{Gianna Lengyel}, \bibinfo{person}{Guillaume Lample}, \bibinfo{person}{Lucile Saulnier}, {et~al\mbox{.}}} \bibinfo{year}{2023}\natexlab{}.
\newblock \showarticletitle{Mistral 7B}.
\newblock \bibinfo{journal}{\emph{arXiv preprint arXiv:2310.06825}} (\bibinfo{year}{2023}).
\newblock


\bibitem[\protect\citeauthoryear{Jiang, Sablayrolles, Roux, Mensch, Savary, Bamford, Chaplot, Casas, Hanna, Bressand, et~al\mbox{.}}{Jiang et~al\mbox{.}}{2024}]%
        {jiang2024mixtral}
\bibfield{author}{\bibinfo{person}{Albert~Q Jiang}, \bibinfo{person}{Alexandre Sablayrolles}, \bibinfo{person}{Antoine Roux}, \bibinfo{person}{Arthur Mensch}, \bibinfo{person}{Blanche Savary}, \bibinfo{person}{Chris Bamford}, \bibinfo{person}{Devendra~Singh Chaplot}, \bibinfo{person}{Diego de~las Casas}, \bibinfo{person}{Emma~Bou Hanna}, \bibinfo{person}{Florian Bressand}, {et~al\mbox{.}}} \bibinfo{year}{2024}\natexlab{}.
\newblock \showarticletitle{Mixtral of experts}.
\newblock \bibinfo{journal}{\emph{arXiv preprint arXiv:2401.04088}} (\bibinfo{year}{2024}).
\newblock


\bibitem[\protect\citeauthoryear{Joshi, Choi, Weld, and Zettlemoyer}{Joshi et~al\mbox{.}}{2017}]%
        {joshi2017triviaqa}
\bibfield{author}{\bibinfo{person}{Mandar Joshi}, \bibinfo{person}{Eunsol Choi}, \bibinfo{person}{Daniel~S Weld}, {and} \bibinfo{person}{Luke Zettlemoyer}.} \bibinfo{year}{2017}\natexlab{}.
\newblock \showarticletitle{Triviaqa: A large scale distantly supervised challenge dataset for reading comprehension}.
\newblock \bibinfo{journal}{\emph{arXiv preprint arXiv:1705.03551}} (\bibinfo{year}{2017}).
\newblock


\bibitem[\protect\citeauthoryear{Karamanolakis, Ma, and Dong}{Karamanolakis et~al\mbox{.}}{2020}]%
        {karamanolakis2020txtract}
\bibfield{author}{\bibinfo{person}{Giannis Karamanolakis}, \bibinfo{person}{Jun Ma}, {and} \bibinfo{person}{Xin~Luna Dong}.} \bibinfo{year}{2020}\natexlab{}.
\newblock \showarticletitle{TXtract: Taxonomy-Aware Knowledge Extraction for Thousands of Product Categories}. In \bibinfo{booktitle}{\emph{Proceedings of the 58th Annual Meeting of the Association for Computational Linguistics}}. \bibinfo{pages}{8489--8502}.
\newblock


\bibitem[\protect\citeauthoryear{Kocielnik, Kangaslahti, Prabhumoye, Hari, Alvarez, and Anandkumar}{Kocielnik et~al\mbox{.}}{2023}]%
        {kocielnik2023can}
\bibfield{author}{\bibinfo{person}{Rafal Kocielnik}, \bibinfo{person}{Sara Kangaslahti}, \bibinfo{person}{Shrimai Prabhumoye}, \bibinfo{person}{Meena Hari}, \bibinfo{person}{Michael Alvarez}, {and} \bibinfo{person}{Anima Anandkumar}.} \bibinfo{year}{2023}\natexlab{}.
\newblock \showarticletitle{Can you label less by using out-of-domain data? Active \& transfer learning with few-shot instructions}. In \bibinfo{booktitle}{\emph{Transfer Learning for Natural Language Processing Workshop}}. PMLR, \bibinfo{pages}{22--32}.
\newblock


\bibitem[\protect\citeauthoryear{Kojima, Gu, Reid, Matsuo, and Iwasawa}{Kojima et~al\mbox{.}}{2022}]%
        {kojima2022large}
\bibfield{author}{\bibinfo{person}{Takeshi Kojima}, \bibinfo{person}{Shixiang~Shane Gu}, \bibinfo{person}{Machel Reid}, \bibinfo{person}{Yutaka Matsuo}, {and} \bibinfo{person}{Yusuke Iwasawa}.} \bibinfo{year}{2022}\natexlab{}.
\newblock \showarticletitle{Large language models are zero-shot reasoners}.
\newblock \bibinfo{journal}{\emph{Advances in neural information processing systems}}  \bibinfo{volume}{35} (\bibinfo{year}{2022}), \bibinfo{pages}{22199--22213}.
\newblock


\bibitem[\protect\citeauthoryear{Kumar, Kim, Ravi, Sun, Faloutsos, Salakhutdinov, and Yoon}{Kumar et~al\mbox{.}}{2023}]%
        {kumar2023automatic}
\bibfield{author}{\bibinfo{person}{Rohan Kumar}, \bibinfo{person}{Youngmin Kim}, \bibinfo{person}{Sunitha Ravi}, \bibinfo{person}{Haitian Sun}, \bibinfo{person}{Christos Faloutsos}, \bibinfo{person}{Ruslan Salakhutdinov}, {and} \bibinfo{person}{Minji Yoon}.} \bibinfo{year}{2023}\natexlab{}.
\newblock \showarticletitle{Automatic Question-Answer Generation for Long-Tail Knowledge}.
\newblock  (\bibinfo{year}{2023}).
\newblock


\bibitem[\protect\citeauthoryear{Kwiatkowski, Palomaki, Redfield, Collins, Parikh, Alberti, Epstein, Polosukhin, Devlin, Lee, et~al\mbox{.}}{Kwiatkowski et~al\mbox{.}}{2019}]%
        {kwiatkowski2019natural}
\bibfield{author}{\bibinfo{person}{Tom Kwiatkowski}, \bibinfo{person}{Jennimaria Palomaki}, \bibinfo{person}{Olivia Redfield}, \bibinfo{person}{Michael Collins}, \bibinfo{person}{Ankur Parikh}, \bibinfo{person}{Chris Alberti}, \bibinfo{person}{Danielle Epstein}, \bibinfo{person}{Illia Polosukhin}, \bibinfo{person}{Jacob Devlin}, \bibinfo{person}{Kenton Lee}, {et~al\mbox{.}}} \bibinfo{year}{2019}\natexlab{}.
\newblock \showarticletitle{Natural questions: a benchmark for question answering research}.
\newblock \bibinfo{journal}{\emph{Transactions of the Association for Computational Linguistics}}  \bibinfo{volume}{7} (\bibinfo{year}{2019}), \bibinfo{pages}{453--466}.
\newblock


\bibitem[\protect\citeauthoryear{Lester, Al-Rfou, and Constant}{Lester et~al\mbox{.}}{2021}]%
        {lester2021power}
\bibfield{author}{\bibinfo{person}{Brian Lester}, \bibinfo{person}{Rami Al-Rfou}, {and} \bibinfo{person}{Noah Constant}.} \bibinfo{year}{2021}\natexlab{}.
\newblock \showarticletitle{The power of scale for parameter-efficient prompt tuning}.
\newblock \bibinfo{journal}{\emph{arXiv preprint arXiv:2104.08691}} (\bibinfo{year}{2021}).
\newblock


\bibitem[\protect\citeauthoryear{Li}{Li}{2022}]%
        {li2022learning}
\bibfield{author}{\bibinfo{person}{Hang Li}.} \bibinfo{year}{2022}\natexlab{}.
\newblock \bibinfo{booktitle}{\emph{Learning to rank for information retrieval and natural language processing}}.
\newblock \bibinfo{publisher}{Springer Nature}.
\newblock


\bibitem[\protect\citeauthoryear{Li and Liang}{Li and Liang}{2021}]%
        {li2021prefix}
\bibfield{author}{\bibinfo{person}{Xiang~Lisa Li} {and} \bibinfo{person}{Percy Liang}.} \bibinfo{year}{2021}\natexlab{}.
\newblock \showarticletitle{Prefix-tuning: Optimizing continuous prompts for generation}.
\newblock \bibinfo{journal}{\emph{arXiv preprint arXiv:2101.00190}} (\bibinfo{year}{2021}).
\newblock


\bibitem[\protect\citeauthoryear{Liu, Yuan, Fu, Jiang, Hayashi, and Neubig}{Liu et~al\mbox{.}}{2023}]%
        {10.1145/3560815}
\bibfield{author}{\bibinfo{person}{Pengfei Liu}, \bibinfo{person}{Weizhe Yuan}, \bibinfo{person}{Jinlan Fu}, \bibinfo{person}{Zhengbao Jiang}, \bibinfo{person}{Hiroaki Hayashi}, {and} \bibinfo{person}{Graham Neubig}.} \bibinfo{year}{2023}\natexlab{}.
\newblock \showarticletitle{Pre-train, Prompt, and Predict: A Systematic Survey of Prompting Methods in Natural Language Processing}.
\newblock \bibinfo{journal}{\emph{ACM Comput. Surv.}} \bibinfo{volume}{55}, \bibinfo{number}{9}, Article \bibinfo{articleno}{195} (\bibinfo{date}{jan} \bibinfo{year}{2023}), \bibinfo{numpages}{35}~pages.
\newblock
\showISSN{0360-0300}
\urldef\tempurl%
\url{https://doi.org/10.1145/3560815}
\showDOI{\tempurl}


\bibitem[\protect\citeauthoryear{Liu et~al\mbox{.}}{Liu et~al\mbox{.}}{2009}]%
        {liu2009learning}
\bibfield{author}{\bibinfo{person}{Tie-Yan Liu} {et~al\mbox{.}}} \bibinfo{year}{2009}\natexlab{}.
\newblock \showarticletitle{Learning to rank for information retrieval}.
\newblock \bibinfo{journal}{\emph{Foundations and Trends{\textregistered} in Information Retrieval}} \bibinfo{volume}{3}, \bibinfo{number}{3} (\bibinfo{year}{2009}), \bibinfo{pages}{225--331}.
\newblock


\bibitem[\protect\citeauthoryear{Luo, Vu, Phung, and Haf}{Luo et~al\mbox{.}}{2023}]%
        {luo2023systematic}
\bibfield{author}{\bibinfo{person}{Linhao Luo}, \bibinfo{person}{Trang Vu}, \bibinfo{person}{Dinh Phung}, {and} \bibinfo{person}{Reza Haf}.} \bibinfo{year}{2023}\natexlab{}.
\newblock \showarticletitle{Systematic Assessment of Factual Knowledge in Large Language Models}. In \bibinfo{booktitle}{\emph{Findings of the Association for Computational Linguistics: EMNLP 2023}}. \bibinfo{pages}{13272--13286}.
\newblock


\bibitem[\protect\citeauthoryear{Mallen, Asai, Zhong, Das, Khashabi, and Hajishirzi}{Mallen et~al\mbox{.}}{2023}]%
        {mallen-etal-2023-trust}
\bibfield{author}{\bibinfo{person}{Alex Mallen}, \bibinfo{person}{Akari Asai}, \bibinfo{person}{Victor Zhong}, \bibinfo{person}{Rajarshi Das}, \bibinfo{person}{Daniel Khashabi}, {and} \bibinfo{person}{Hannaneh Hajishirzi}.} \bibinfo{year}{2023}\natexlab{}.
\newblock \showarticletitle{When Not to Trust Language Models: Investigating Effectiveness of Parametric and Non-Parametric Memories}. In \bibinfo{booktitle}{\emph{Proceedings of the 61st Annual Meeting of the Association for Computational Linguistics (Volume 1: Long Papers)}}, \bibfield{editor}{\bibinfo{person}{Anna Rogers}, \bibinfo{person}{Jordan Boyd-Graber}, {and} \bibinfo{person}{Naoaki Okazaki}} (Eds.). \bibinfo{publisher}{Association for Computational Linguistics}, \bibinfo{address}{Toronto, Canada}, \bibinfo{pages}{9802--9822}.
\newblock
\urldef\tempurl%
\url{https://doi.org/10.18653/v1/2023.acl-long.546}
\showDOI{\tempurl}


\bibitem[\protect\citeauthoryear{Neuhaus}{Neuhaus}{2023}]%
        {neuhaus2023ontologies}
\bibfield{author}{\bibinfo{person}{Fabian Neuhaus}.} \bibinfo{year}{2023}\natexlab{}.
\newblock \showarticletitle{Ontologies in the era of large language models--a perspective}.
\newblock \bibinfo{journal}{\emph{Applied Ontology}} \bibinfo{volume}{18}, \bibinfo{number}{4} (\bibinfo{year}{2023}), \bibinfo{pages}{399--407}.
\newblock


\bibitem[\protect\citeauthoryear{Nordhoff and Hammarstr{\"o}m}{Nordhoff and Hammarstr{\"o}m}{2011}]%
        {nordhoff2011glottolog}
\bibfield{author}{\bibinfo{person}{Sebastian Nordhoff} {and} \bibinfo{person}{Harald Hammarstr{\"o}m}.} \bibinfo{year}{2011}\natexlab{}.
\newblock \showarticletitle{Glottolog/Langdoc: Defining dialects, languages, and language families as collections of resources}. In \bibinfo{booktitle}{\emph{First International Workshop on Linked Science 2011-In conjunction with the International Semantic Web Conference (ISWC 2011)}}.
\newblock


\bibitem[\protect\citeauthoryear{Papadimitriou, Tsaparas, Fuxman, and Getoor}{Papadimitriou et~al\mbox{.}}{2012}]%
        {papadimitriou2012taci}
\bibfield{author}{\bibinfo{person}{Panagiotis Papadimitriou}, \bibinfo{person}{Panayiotis Tsaparas}, \bibinfo{person}{Ariel Fuxman}, {and} \bibinfo{person}{Lise Getoor}.} \bibinfo{year}{2012}\natexlab{}.
\newblock \showarticletitle{TACI: Taxonomy-aware catalog integration}.
\newblock \bibinfo{journal}{\emph{IEEE Transactions on knowledge and data engineering}} \bibinfo{volume}{25}, \bibinfo{number}{7} (\bibinfo{year}{2012}), \bibinfo{pages}{1643--1655}.
\newblock


\bibitem[\protect\citeauthoryear{Petroni, Rockt{\"a}schel, Lewis, Bakhtin, Wu, Miller, and Riedel}{Petroni et~al\mbox{.}}{2019}]%
        {petroni2019language}
\bibfield{author}{\bibinfo{person}{Fabio Petroni}, \bibinfo{person}{Tim Rockt{\"a}schel}, \bibinfo{person}{Patrick Lewis}, \bibinfo{person}{Anton Bakhtin}, \bibinfo{person}{Yuxiang Wu}, \bibinfo{person}{Alexander~H Miller}, {and} \bibinfo{person}{Sebastian Riedel}.} \bibinfo{year}{2019}\natexlab{}.
\newblock \showarticletitle{Language models as knowledge bases?}
\newblock \bibinfo{journal}{\emph{arXiv preprint arXiv:1909.01066}} (\bibinfo{year}{2019}).
\newblock


\bibitem[\protect\citeauthoryear{Romera-Paredes and Torr}{Romera-Paredes and Torr}{2015}]%
        {romera2015embarrassingly}
\bibfield{author}{\bibinfo{person}{Bernardino Romera-Paredes} {and} \bibinfo{person}{Philip Torr}.} \bibinfo{year}{2015}\natexlab{}.
\newblock \showarticletitle{An embarrassingly simple approach to zero-shot learning}. In \bibinfo{booktitle}{\emph{International conference on machine learning}}. PMLR, \bibinfo{pages}{2152--2161}.
\newblock


\bibitem[\protect\citeauthoryear{Sayers, Cavanaugh, Clark, Ostell, Pruitt, and Karsch-Mizrachi}{Sayers et~al\mbox{.}}{2019}]%
        {sayers2019genbank}
\bibfield{author}{\bibinfo{person}{Eric~W Sayers}, \bibinfo{person}{Mark Cavanaugh}, \bibinfo{person}{Karen Clark}, \bibinfo{person}{James Ostell}, \bibinfo{person}{Kim~D Pruitt}, {and} \bibinfo{person}{Ilene Karsch-Mizrachi}.} \bibinfo{year}{2019}\natexlab{}.
\newblock \showarticletitle{GenBank}.
\newblock \bibinfo{journal}{\emph{Nucleic acids research}} \bibinfo{volume}{47}, \bibinfo{number}{Database issue} (\bibinfo{year}{2019}), \bibinfo{pages}{D94}.
\newblock


\bibitem[\protect\citeauthoryear{Schick and Sch{\"u}tze}{Schick and Sch{\"u}tze}{2020}]%
        {schick2020s}
\bibfield{author}{\bibinfo{person}{Timo Schick} {and} \bibinfo{person}{Hinrich Sch{\"u}tze}.} \bibinfo{year}{2020}\natexlab{}.
\newblock \showarticletitle{It's not just size that matters: Small language models are also few-shot learners}.
\newblock \bibinfo{journal}{\emph{arXiv preprint arXiv:2009.07118}} (\bibinfo{year}{2020}).
\newblock


\bibitem[\protect\citeauthoryear{Schoch, Ciufo, Domrachev, Hotton, Kannan, Khovanskaya, Leipe, Mcveigh, O’Neill, Robbertse, et~al\mbox{.}}{Schoch et~al\mbox{.}}{2020}]%
        {schoch2020ncbi}
\bibfield{author}{\bibinfo{person}{Conrad~L Schoch}, \bibinfo{person}{Stacy Ciufo}, \bibinfo{person}{Mikhail Domrachev}, \bibinfo{person}{Carol~L Hotton}, \bibinfo{person}{Sivakumar Kannan}, \bibinfo{person}{Rogneda Khovanskaya}, \bibinfo{person}{Detlef Leipe}, \bibinfo{person}{Richard Mcveigh}, \bibinfo{person}{Kathleen O’Neill}, \bibinfo{person}{Barbara Robbertse}, {et~al\mbox{.}}} \bibinfo{year}{2020}\natexlab{}.
\newblock \showarticletitle{NCBI Taxonomy: a comprehensive update on curation, resources and tools}.
\newblock \bibinfo{journal}{\emph{Database}}  \bibinfo{volume}{2020} (\bibinfo{year}{2020}), \bibinfo{pages}{baaa062}.
\newblock


\bibitem[\protect\citeauthoryear{Sciavolino, Zhong, Lee, and Chen}{Sciavolino et~al\mbox{.}}{2021}]%
        {sciavolino-etal-2021-simple}
\bibfield{author}{\bibinfo{person}{Christopher Sciavolino}, \bibinfo{person}{Zexuan Zhong}, \bibinfo{person}{Jinhyuk Lee}, {and} \bibinfo{person}{Danqi Chen}.} \bibinfo{year}{2021}\natexlab{}.
\newblock \showarticletitle{Simple Entity-Centric Questions Challenge Dense Retrievers}. In \bibinfo{booktitle}{\emph{Proceedings of the 2021 Conference on Empirical Methods in Natural Language Processing}}, \bibfield{editor}{\bibinfo{person}{Marie-Francine Moens}, \bibinfo{person}{Xuanjing Huang}, \bibinfo{person}{Lucia Specia}, {and} \bibinfo{person}{Scott Wen-tau Yih}} (Eds.). \bibinfo{publisher}{Association for Computational Linguistics}, \bibinfo{address}{Online and Punta Cana, Dominican Republic}, \bibinfo{pages}{6138--6148}.
\newblock
\urldef\tempurl%
\url{https://doi.org/10.18653/v1/2021.emnlp-main.496}
\showDOI{\tempurl}


\bibitem[\protect\citeauthoryear{Sen}{Sen}{2019}]%
        {sen2019knowledge}
\bibfield{author}{\bibinfo{person}{Yasemin Sen}.} \bibinfo{year}{2019}\natexlab{}.
\newblock \showarticletitle{Knowledge as a valuable asset of organizations: Taxonomy, management and implications}.
\newblock In \bibinfo{booktitle}{\emph{Management science: Foundations and innovations}}. \bibinfo{publisher}{Springer}, \bibinfo{pages}{29--48}.
\newblock


\bibitem[\protect\citeauthoryear{Suchanek, Kasneci, and Weikum}{Suchanek et~al\mbox{.}}{2007}]%
        {suchanek2007yago}
\bibfield{author}{\bibinfo{person}{Fabian~M Suchanek}, \bibinfo{person}{Gjergji Kasneci}, {and} \bibinfo{person}{Gerhard Weikum}.} \bibinfo{year}{2007}\natexlab{}.
\newblock \showarticletitle{Yago: a core of semantic knowledge}. In \bibinfo{booktitle}{\emph{Proceedings of the 16th international conference on World Wide Web}}. \bibinfo{pages}{697--706}.
\newblock


\bibitem[\protect\citeauthoryear{Sun, Xu, Zha, Liu, and Dong}{Sun et~al\mbox{.}}{2023b}]%
        {sun2023head}
\bibfield{author}{\bibinfo{person}{Kai Sun}, \bibinfo{person}{Yifan~Ethan Xu}, \bibinfo{person}{Hanwen Zha}, \bibinfo{person}{Yue Liu}, {and} \bibinfo{person}{Xin~Luna Dong}.} \bibinfo{year}{2023}\natexlab{b}.
\newblock \showarticletitle{Head-to-tail: How knowledgeable are large language models (llm)? AKA will llms replace knowledge graphs?}
\newblock \bibinfo{journal}{\emph{arXiv preprint arXiv:2308.10168}} (\bibinfo{year}{2023}).
\newblock


\bibitem[\protect\citeauthoryear{Sun, Xin, and Chen}{Sun et~al\mbox{.}}{2023a}]%
        {sun2023reca}
\bibfield{author}{\bibinfo{person}{Yushi Sun}, \bibinfo{person}{Hao Xin}, {and} \bibinfo{person}{Lei Chen}.} \bibinfo{year}{2023}\natexlab{a}.
\newblock \showarticletitle{Reca: Related tables enhanced column semantic type annotation framework}.
\newblock \bibinfo{journal}{\emph{Proceedings of the VLDB Endowment}} \bibinfo{volume}{16}, \bibinfo{number}{6} (\bibinfo{year}{2023}), \bibinfo{pages}{1319--1331}.
\newblock


\bibitem[\protect\citeauthoryear{Tan, Min, Li, Li, Hu, Chen, and Qi}{Tan et~al\mbox{.}}{2023}]%
        {tan2023evaluation}
\bibfield{author}{\bibinfo{person}{Yiming Tan}, \bibinfo{person}{Dehai Min}, \bibinfo{person}{Yu Li}, \bibinfo{person}{Wenbo Li}, \bibinfo{person}{Nan Hu}, \bibinfo{person}{Yongrui Chen}, {and} \bibinfo{person}{Guilin Qi}.} \bibinfo{year}{2023}\natexlab{}.
\newblock \showarticletitle{Evaluation of ChatGPT as a question answering system for answering complex questions}.
\newblock  (\bibinfo{year}{2023}).
\newblock


\bibitem[\protect\citeauthoryear{Touvron, Martin, Stone, Albert, Almahairi, Babaei, Bashlykov, Batra, Bhargava, Bhosale, et~al\mbox{.}}{Touvron et~al\mbox{.}}{2023}]%
        {touvron2023llama}
\bibfield{author}{\bibinfo{person}{Hugo Touvron}, \bibinfo{person}{Louis Martin}, \bibinfo{person}{Kevin Stone}, \bibinfo{person}{Peter Albert}, \bibinfo{person}{Amjad Almahairi}, \bibinfo{person}{Yasmine Babaei}, \bibinfo{person}{Nikolay Bashlykov}, \bibinfo{person}{Soumya Batra}, \bibinfo{person}{Prajjwal Bhargava}, \bibinfo{person}{Shruti Bhosale}, {et~al\mbox{.}}} \bibinfo{year}{2023}\natexlab{}.
\newblock \showarticletitle{Llama 2: Open foundation and fine-tuned chat models}.
\newblock \bibinfo{journal}{\emph{arXiv preprint arXiv:2307.09288}} (\bibinfo{year}{2023}).
\newblock


\bibitem[\protect\citeauthoryear{Wadhwa, Amir, and Wallace}{Wadhwa et~al\mbox{.}}{2023}]%
        {wadhwa2023revisiting}
\bibfield{author}{\bibinfo{person}{Somin Wadhwa}, \bibinfo{person}{Silvio Amir}, {and} \bibinfo{person}{Byron~C Wallace}.} \bibinfo{year}{2023}\natexlab{}.
\newblock \showarticletitle{Revisiting relation extraction in the era of large language models}.
\newblock \bibinfo{journal}{\emph{arXiv preprint arXiv:2305.05003}} (\bibinfo{year}{2023}).
\newblock


\bibitem[\protect\citeauthoryear{Wang, Tang, Duan, Wei, Huang, Cao, Jiang, Zhou, et~al\mbox{.}}{Wang et~al\mbox{.}}{2020}]%
        {wang2020k}
\bibfield{author}{\bibinfo{person}{Ruize Wang}, \bibinfo{person}{Duyu Tang}, \bibinfo{person}{Nan Duan}, \bibinfo{person}{Zhongyu Wei}, \bibinfo{person}{Xuanjing Huang}, \bibinfo{person}{Guihong Cao}, \bibinfo{person}{Daxin Jiang}, \bibinfo{person}{Ming Zhou}, {et~al\mbox{.}}} \bibinfo{year}{2020}\natexlab{}.
\newblock \showarticletitle{K-adapter: Infusing knowledge into pre-trained models with adapters}.
\newblock \bibinfo{journal}{\emph{arXiv preprint arXiv:2002.01808}} (\bibinfo{year}{2020}).
\newblock


\bibitem[\protect\citeauthoryear{Wei, Wang, Schuurmans, Bosma, Xia, Chi, Le, Zhou, et~al\mbox{.}}{Wei et~al\mbox{.}}{2022}]%
        {wei2022chain}
\bibfield{author}{\bibinfo{person}{Jason Wei}, \bibinfo{person}{Xuezhi Wang}, \bibinfo{person}{Dale Schuurmans}, \bibinfo{person}{Maarten Bosma}, \bibinfo{person}{Fei Xia}, \bibinfo{person}{Ed Chi}, \bibinfo{person}{Quoc~V Le}, \bibinfo{person}{Denny Zhou}, {et~al\mbox{.}}} \bibinfo{year}{2022}\natexlab{}.
\newblock \showarticletitle{Chain-of-thought prompting elicits reasoning in large language models}.
\newblock \bibinfo{journal}{\emph{Advances in Neural Information Processing Systems}}  \bibinfo{volume}{35} (\bibinfo{year}{2022}), \bibinfo{pages}{24824--24837}.
\newblock


\bibitem[\protect\citeauthoryear{Yang, Sun, Xin, Sun, Bhalla, Chen, Choudhary, Gui, Jiang, Jiang, et~al\mbox{.}}{Yang et~al\mbox{.}}{2024}]%
        {yang2024crag}
\bibfield{author}{\bibinfo{person}{Xiao Yang}, \bibinfo{person}{Kai Sun}, \bibinfo{person}{Hao Xin}, \bibinfo{person}{Yushi Sun}, \bibinfo{person}{Nikita Bhalla}, \bibinfo{person}{Xiangsen Chen}, \bibinfo{person}{Sajal Choudhary}, \bibinfo{person}{Rongze~Daniel Gui}, \bibinfo{person}{Ziran~Will Jiang}, \bibinfo{person}{Ziyu Jiang}, {et~al\mbox{.}}} \bibinfo{year}{2024}\natexlab{}.
\newblock \showarticletitle{CRAG-Comprehensive RAG Benchmark}.
\newblock \bibinfo{journal}{\emph{arXiv preprint arXiv:2406.04744}} (\bibinfo{year}{2024}).
\newblock


\bibitem[\protect\citeauthoryear{Zhang, Ladhak, Durmus, Liang, McKeown, and Hashimoto}{Zhang et~al\mbox{.}}{2023}]%
        {zhang2023benchmarking}
\bibfield{author}{\bibinfo{person}{Tianyi Zhang}, \bibinfo{person}{Faisal Ladhak}, \bibinfo{person}{Esin Durmus}, \bibinfo{person}{Percy Liang}, \bibinfo{person}{Kathleen McKeown}, {and} \bibinfo{person}{Tatsunori~B Hashimoto}.} \bibinfo{year}{2023}\natexlab{}.
\newblock \showarticletitle{Benchmarking large language models for news summarization}.
\newblock \bibinfo{journal}{\emph{arXiv preprint arXiv:2301.13848}} (\bibinfo{year}{2023}).
\newblock


\bibitem[\protect\citeauthoryear{Zheng, Chiang, Sheng, Zhuang, Wu, Zhuang, Lin, Li, Li, Xing, et~al\mbox{.}}{Zheng et~al\mbox{.}}{2023}]%
        {zheng2023judging}
\bibfield{author}{\bibinfo{person}{Lianmin Zheng}, \bibinfo{person}{Wei-Lin Chiang}, \bibinfo{person}{Ying Sheng}, \bibinfo{person}{Siyuan Zhuang}, \bibinfo{person}{Zhanghao Wu}, \bibinfo{person}{Yonghao Zhuang}, \bibinfo{person}{Zi Lin}, \bibinfo{person}{Zhuohan Li}, \bibinfo{person}{Dacheng Li}, \bibinfo{person}{Eric Xing}, {et~al\mbox{.}}} \bibinfo{year}{2023}\natexlab{}.
\newblock \showarticletitle{Judging LLM-as-a-judge with MT-Bench and Chatbot Arena}.
\newblock \bibinfo{journal}{\emph{arXiv preprint arXiv:2306.05685}} (\bibinfo{year}{2023}).
\newblock


\bibitem[\protect\citeauthoryear{Zhu, Yuan, Wang, Liu, Liu, Deng, Dou, and Wen}{Zhu et~al\mbox{.}}{2023}]%
        {zhu2023large}
\bibfield{author}{\bibinfo{person}{Yutao Zhu}, \bibinfo{person}{Huaying Yuan}, \bibinfo{person}{Shuting Wang}, \bibinfo{person}{Jiongnan Liu}, \bibinfo{person}{Wenhan Liu}, \bibinfo{person}{Chenlong Deng}, \bibinfo{person}{Zhicheng Dou}, {and} \bibinfo{person}{Ji-Rong Wen}.} \bibinfo{year}{2023}\natexlab{}.
\newblock \showarticletitle{Large language models for information retrieval: A survey}.
\newblock \bibinfo{journal}{\emph{arXiv preprint arXiv:2308.07107}} (\bibinfo{year}{2023}).
\newblock


\end{thebibliography}
\clearpage





\nobalance
\section{Supplementary Materials}
\label{sec:supp}
This section contains the supplementary materials for our paper. In Section~\ref{supp:exp-results}, we present the experimental results on Easy and MCQ datasets.

\newpage
\subsection{Experimental results}
\label{supp:exp-results}
We present the experimental results on Easy and MCQ datasets in Tables~\ref{tab:exp-main-easy}, and~\ref{tab:exp-mcq}.

\begin{table*}[hbtp]
  \caption{\rev{Overall results on easy datasets.}}
  \label{tab:exp-main-easy}
  {\color{black}\begin{tabular}{l|l|c|c|c|c|c|c|c|c|c|c}
    \Xhline{0.8pt}
    \multicolumn{1}{c|}{}&\multicolumn{1}{c|}{}&\multicolumn{1}{c|}{\textbf{eBay}}&\multicolumn{1}{c|}{\textbf{Amazon}}&\multicolumn{1}{c|}{\textbf{Google}}&\multicolumn{1}{c|}{\textbf{Schema}}&\multicolumn{1}{c|}{\textbf{ACM-CCS}}&\multicolumn{1}{c|}{\textbf{GeoNames}}&\multicolumn{1}{c|}{\textbf{Glottolog}}&\multicolumn{1}{c|}{\textbf{ICD-10-CM}}&\multicolumn{1}{c|}{\textbf{OAE}}&\multicolumn{1}{c}{\textbf{NCBI}}\\
    \hline
    \multirow{2}{*}{GPT-3.5} & $A$ &  0.921&0.775&0.920&0.593&0.711&0.598&0.563&0.851&0.778&0.529
 \\
    & $M$ &  0.026&0.148&0.045&0.334&0.169&0.057&0.310&0.099&0.167&0.354
  \\
    \multirow{2}{*}{GPT-4} & $A$ &0.946&0.879&0.962&0.773&0.860&0.648&0.710&0.964&0.866&0.789
 \\
    & $M$ &  0.002&0.044&0.008&0.183&0.014&0.002&0.141&0.001&0.032&0.089
\\
\hline
    \multirow{2}{*}{Claude-3} & $A$ &  0.932&0.758&0.910&0.331&0.784&0.679&0.256&0.953&0.869&0.486
\\
    & $M$ &  0.018&0.199&0.068&0.664&0.121&0.138&0.736&0.021&0.095&0.494
\\
    \hline
    \multirow{2}{*}{Llama-2-7B} & $A$ & 0.196&0.053&0.090&0.000&0.032&0.006&0.001&0.115&0.004&0.000
 \\
    & $M$ & 0.804&0.946&0.908&1.000&0.967&0.994&0.999&0.880&0.996&1.000
  \\
    \multirow{2}{*}{Llama-2-13B} & $A$ & 0.926&0.798&0.886&0.727&0.789&0.543&0.149&0.815&0.714&0.411
 \\
    & $M$ & 0.000&0.012&0.005&0.020&0.024&0.006&0.733&0.077&0.145&0.310
 \\
    \multirow{2}{*}{Llama-2-70B} & $A$ & 0.931&0.865&0.945&0.616&0.817&0.553&0.292&0.884&0.761&0.536
 \\
    & $M$ & 0.000&0.001&0.000&0.001&0.004&0.000&0.487&0.040&0.022&0.138
 \\
    \hline
    \multirow{2}{*}{Llama-3-8B} & $A$ & 0.931&0.871&0.940&0.819&0.829&0.663&0.665&0.915&0.893&0.785
 \\
    & $M$ & 0.000&0.000&0.000&0.000&0.000&0.000&0.086&0.020&0.000&0.015
 \\
    \multirow{2}{*}{Llama-3-70B} & $A$ & 0.939&0.797&0.927&0.376&0.787&0.693&0.354&0.899&0.804&0.514
 \\
    & $M$ & 0.005&0.103&0.039&0.602&0.104&0.073&0.591&0.055&0.139&0.399
\\
    \hline
    \multirow{2}{*}{Flan-T5-3B} & $A$ & 0.926&0.826&0.934&0.751&0.732&0.539&0.588&0.811&0.851&0.605
 \\
    & $M$ & 0.000 & 0.000 & 0.000 & 0.000 & 0.000 & 0.000 & 0.000 & 0.000 & 0.000 & 0.000  \\
    \multirow{2}{*}{Flan-T5-11B} & $A$ & 0.944&0.834&0.944&0.804&0.737&0.520&0.595&0.871&0.875&0.643
 \\
    & $M$ & 0.000 & 0.000 & 0.000 & 0.000 & 0.000 & 0.000 & 0.000 & 0.000 & 0.000 & 0.000  \\
    \hline
    \multirow{2}{*}{Falcon-7B} & $A$ & 0.607&0.553&0.574&0.504&0.577&0.533&0.504&0.677&0.495&0.618
 \\
    & $M$ & 0.000 & 0.000 & 0.000 & 0.000 & 0.000 & 0.000 & 0.000 & 0.000 & 0.000 & 0.000  \\
    \multirow{2}{*}{Falcon-40B} & $A$ & 0.434&0.255&0.357&0.013&0.042&0.106&0.021&0.449&0.005&0.012
 \\
    & $M$ & 0.541&0.732&0.636&0.987&0.957&0.860&0.977&0.536&0.994&0.987
 \\
        \hline
    \multirow{2}{*}{Vicuna-7B} & $A$ & 0.919&0.814&0.915&0.723&0.748&0.705&0.671&0.877&0.878&0.676
 \\
    & $M$ & 0.000&0.000&0.000&0.000&0.000&0.000&0.000&0.000&0.000&0.002
 \\
    \multirow{2}{*}{Vicuna-13B} & $A$ & 0.812&0.708&0.741&0.580&0.575&0.492&0.297&0.668&0.409&0.347
 \\
    & $M$ & 0.035&0.115&0.084&0.133&0.059&0.114&0.620&0.234&0.526&0.530
\\
    \multirow{2}{*}{Vicuna-33B} & $A$ & 0.871&0.839&0.857&0.792&0.737&0.728&0.536&0.887&0.887&0.601
 \\
    & $M$ & 0.000&0.000&0.000&0.010&0.000&0.000&0.272&0.016&0.001&0.216
\\
    \hline
    \multirow{2}{*}{Mistral} & $A$ & 0.571&0.460&0.509&0.211&0.428&0.240&0.146&0.467&0.405&0.176
\\
    & $M$ & 0.347&0.480&0.433&0.774&0.471&0.691&0.842&0.491&0.565&0.811
\\
    \multirow{2}{*}{Mixtral} & $A$ & 0.898&0.829&0.894&0.745&0.656&0.604&0.369&0.873&0.809&0.451
\\
    & $M$ & 0.005&0.002&0.002&0.024&0.202&0.041&0.526&0.048&0.066&0.465
\\
    \hline
    \multirow{2}{*}{LLMs4OL} & $A$ & 0.929&0.890&0.908&0.932&0.844&0.677&0.739&0.935&0.933&0.748
\\
    & $M$ & 0.000 & 0.000 & 0.000 & 0.000 & 0.000 & 0.000 & 0.000 & 0.000 & 0.000 & 0.000  \\

    \Xhline{0.8pt}
  \end{tabular}}
\end{table*}

\begin{table*}[t!]
  \caption{\rev{Overall results on MCQ datasets.}}
  \label{tab:exp-mcq}
  {\color{black}\begin{tabular}{l|l|c|c|c|c|c|c|c|c|c|c}
    \Xhline{0.8pt}
    \multicolumn{1}{c|}{}&\multicolumn{1}{c|}{}&\multicolumn{1}{c|}{\textbf{eBay}}&\multicolumn{1}{c|}{\textbf{Amazon}}&\multicolumn{1}{c|}{\textbf{Google}}&\multicolumn{1}{c|}{\textbf{Schema}}&\multicolumn{1}{c|}{\textbf{ACM-CCS}}&\multicolumn{1}{c|}{\textbf{GeoNames}}&\multicolumn{1}{c|}{\textbf{Glottolog}}&\multicolumn{1}{c|}{\textbf{ICD-10-CM}}&\multicolumn{1}{c|}{\textbf{OAE}}&\multicolumn{1}{c}{\textbf{NCBI}}\\
    \hline
    \multirow{2}{*}{GPT-3.5} & $A$ &  0.931&0.782&0.894&0.842&0.709&0.504&0.534&0.915&0.824&0.531
  \\
    & $M$ & 0.013&0.067&0.025&0.040&0.044&0.053&0.254&0.023&0.038&0.209
\\
    \multirow{2}{*}{GPT-4} & $A$ &  0.947&0.849&0.932&0.907&0.790&0.695&0.683&0.964&0.900&0.701
\\
    & $M$ & 0.000&0.027&0.001&0.000&0.001&0.000&0.003&0.000&0.004&0.009
 \\
 \hline
    \multirow{2}{*}{Claude-3} & $A$ &  0.947&0.830&0.922&0.894&0.741&0.638&0.497&0.962&0.885&0.577
\\
    & $M$ & 0.000&0.043&0.003&0.020&0.022&0.004&0.401&0.004&0.032&0.286
\\
    \hline
    \multirow{2}{*}{Llama-2-7B} & $A$ &  0.488&0.399&0.393&0.304&0.259&0.305&0.313&0.406&0.283&0.271
\\
    & $M$ & 0.000&0.005&0.002&0.008&0.003&0.004&0.089&0.004&0.035&0.002
\\
    \multirow{2}{*}{Llama-2-13B} & $A$ & 0.680&0.551&0.547&0.448&0.450&0.313&0.305&0.695&0.429&0.368
\\
    & $M$ & 0.003&0.026&0.037&0.070&0.006&0.016&0.072&0.015&0.088&0.001
 \\
    \multirow{2}{*}{Llama-2-70B} & $A$ &  0.881&0.704&0.794&0.604&0.556&0.350&0.352&0.776&0.613&0.359
\\
    & $M$ & 0.000&0.004&0.000&0.004&0.001&0.004&0.015&0.003&0.002&0.000
\\
    \hline
    \multirow{2}{*}{Llama-3-8B} & $A$ & 0.865&0.725&0.817&0.689&0.610&0.431&0.449&0.896&0.813&0.449
 \\
    & $M$ &  0.000&0.017&0.006&0.014&0.006&0.000&0.101&0.001&0.001&0.003
\\
    \multirow{2}{*}{Llama-3-70B} & $A$ & 0.941&0.790&0.898&0.805&0.729&0.598&0.634&0.956&0.905&0.650
 \\
    & $M$ & 0.000&0.021&0.001&0.010&0.001&0.000&0.011&0.000&0.002&0.003
 \\
    \hline
    \multirow{2}{*}{Flan-T5-3B} & $A$ &  0.898&0.756&0.878&0.799&0.664&0.455&0.506&0.836&0.777&0.524
  \\
    & $M$ & 0.000&0.000&0.000&0.000&0.000&0.000&0.000&0.000&0.000&0.000
\\
    \multirow{2}{*}{Flan-T5-11B} & $A$ & 0.904&0.805&0.902&0.841&0.696&0.634&0.583&0.877&0.817&0.550
\\
    & $M$ & 0.000&0.000&0.000&0.000&0.000&0.000&0.000&0.000&0.000&0.000 \\
    \hline
    \multirow{2}{*}{Falcon-7B} & $A$ & 0.261&0.259&0.275&0.233&0.240&0.256&0.260&0.254&0.262&0.255
\\
    & $M$ & 0.000&0.000&0.000&0.000&0.000&0.000&0.000&0.003&0.000&0.000
\\
    \multirow{2}{*}{Falcon-40B} & $A$ & 0.578&0.465&0.525&0.368&0.389&0.305&0.036&0.711&0.168&0.117
\\
    & $M$ & 0.168&0.332&0.269&0.494&0.198&0.215&0.960&0.059&0.593&0.804
\\
        \hline
    \multirow{2}{*}{Vicuna-7B} & $A$ & 0.617&0.493&0.527&0.425&0.384&0.313&0.409&0.528&0.473&0.392
 \\
    & $M$ & 0.000&0.001&0.000&0.000&0.000&0.008&0.017&0.011&0.166&0.005
\\
    \multirow{2}{*}{Vicuna-13B} & $A$ & 0.838&0.636&0.747&0.523&0.543&0.317&0.205&0.833&0.691&0.362
\\
    & $M$ & 0.013&0.100&0.061&0.137&0.079&0.146&0.664&0.045&0.196&0.375
\\
    \multirow{2}{*}{Vicuna-33B} & $A$ & 0.327&0.307&0.338&0.266&0.261&0.264&0.253&0.401&0.285&0.252
\\
    & $M$ & 0.000&0.000&0.000&0.000&0.000&0.000&0.000&0.001&0.009&0.000
\\
    \hline
    \multirow{2}{*}{Mistral} & $A$ & 0.828&0.666&0.713&0.692&0.582&0.415&0.436&0.765&0.768&0.519
 \\
    & $M$ & 0.000&0.014&0.006&0.006&0.003&0.000&0.179&0.008&0.004&0.044
 \\
    \multirow{2}{*}{Mixtral} & $A$ & 0.924&0.768&0.876&0.775&0.707&0.537&0.611&0.923&0.797&0.634
 \\
    & $M$ & 0.000&0.040&0.008&0.038&0.008&0.020&0.089&0.005&0.003&0.067
\\
    \hline
    \multirow{2}{*}{LLMs4OL} & $A$ & 0.954&0.851&0.924&0.921&0.796&0.683&0.660&0.921&0.941&0.650
 \\
    & $M$ & 0.000&0.000&0.000&0.000&0.000&0.000&0.000&0.000&0.000&0.000
\\

    \Xhline{0.8pt}
  \end{tabular}}
\end{table*}    

\end{document}